\documentclass{article}

\usepackage[english]{babel}

\usepackage[T1]{fontenc}
\usepackage{lmodern}
\usepackage[letterpaper,top=2cm,bottom=2cm,left=3cm,right=3cm,marginparwidth=1.75cm]{geometry}
\usepackage{booktabs}
\usepackage[table]{xcolor}
\usepackage{amsmath}
\usepackage{amssymb}
\usepackage{graphicx}

\usepackage[colorlinks=true, allcolors=black]{hyperref}

\usepackage{tikz}
\usetikzlibrary{shapes.geometric, arrows, arrows.meta, positioning, calc}
\usepackage{subcaption}
\usepackage{array}
\usepackage{caption}
\usepackage{xparse}

\providecommand{\multirow}[3]{\raisebox{-0.5ex}[0pt][0pt]{#3}}

\RenewDocumentCommand{\textcolor}{O{}mm}{#3}
\RenewDocumentCommand{\color}{O{}m}{}

\DeclareMathOperator*{\argmin}{argmin}

\newtheorem{theorem}{Theorem}
\newtheorem{lemma}{Lemma}
\newtheorem{definition}{Definition}
\newtheorem{fact}{Fact}

\newtheorem{corollary}{Corollary}

\newtheorem{proposition}{Proposition}

\newenvironment{proofof}[1]{\par\vspace{0.05in}\noindent\textbf{Proof of #1.}\rmfamily\ }{\hspace*{\fill}$\Box$\par\vspace{0.05in}}
\newenvironment{proof}{\par\vspace{0.05in}\noindent\textbf{Proof.}\rmfamily\ }{\hspace*{\fill}$\Box$\par\vspace{0.05in}}

\title{\textcolor{black}{Towards Provable and Scalable Training of Quantized Neural Networks with Ising Optimization}}
\author{Wenxin Li\textsuperscript{1}\thanks{Email: liwx@boseq.com}, Chuan Wang\textsuperscript{2}\thanks{Corresponding Author. Email: wangchuan@bnu.edu.cn}, Hongdong Zhu\textsuperscript{1}, Qi Gao\textsuperscript{1}, Yin Ma\textsuperscript{1}, Hai Wei\textsuperscript{1}, Kai Wen\textsuperscript{1}\thanks{Corresponding Author. Email: wenk@boseq.com}
\\
\small
\textsuperscript{1}{Beijing QBoson Quantum Technology Co., Ltd., Beijing 100015, China}\\
\small
\textsuperscript{2}{School of Artificial Intelligence, Beijing Normal University, Beijing 100875, China}
}

\begin{document}

\maketitle

\begin{abstract}
In this work, we introduce a Quadratic Constrained Binary Optimization (QCBO) framework for training quantized neural networks with provable guarantees. We first characterize the stratified topology of zero-loss sets for continuous piecewise-polynomial networks: generic interior strata are smooth, yet globally optimal components can be disconnected even under overparameterization. This separation between local regularity and global connectivity exposes a geometric obstruction to local training. We then compile any prescribed finite-depth feedforward architecture with finite parameter codebooks and Forward Interval Propagation (FIP)-bounded states into a bounded QCBO. Continuous piecewise-polynomial activations and losses are represented exactly; more general continuous nonlinearities are handled by controlled spline approximation on compact propagated ranges. The resulting finite model admits an exact completely positive reformulation that preserves its global optimal value and optimal face. Unlike sign-pattern convexifications tailored primarily to shallow, positively homogeneous ReLU networks and semidefinite relaxations that can introduce an optimality gap, our formulation accommodates arbitrary depth and width within this broad activation class. Convexification does not remove combinatorial hardness, but isolates it in completely positive cone geometry and enables a structured Ising--classical division of labor via Quantum Conditional Gradient Descent (QCGD): affine constraints are enforced classically, while the physical Ising device supplies unconstrained QUBO linear-minimization oracles. We connect binary network description length to finite-class generalization and logical-spin requirements, establish QCGD convergence under stochastic hardware noise and finite coefficient precision, and formulate sample-wise Decomposed Lower-Bound Optimization (DLBO) to reduce each Ising call from dataset to single-sample scale. The DLBO moment hierarchy also forms a Hamiltonian-locality hierarchy, with order two giving an auxiliary-free pairwise QUBO oracle and higher orders trading interaction locality for tighter bounds. Spectral--ADMM rounding recovers feasible mixed-binary parameters from lifted moments. Experiments on a coherent Ising machine achieve $94.95\%$ Fashion-MNIST accuracy at 1.1-bit precision, outperforming classical quantization-aware and post-training baselines in accuracy, training speed, memory footprint, and inference latency.
\end{abstract}

\section{Introduction}

The deployment of deep neural networks (DNNs) in resource-constrained edge environments demands dramatic reductions in memory footprint, computational complexity, and energy consumption~\cite{liu2024lightweight,hubara2018quantized}. Neural network quantization addresses these challenges by mapping continuous floating-point weights and activations onto low-bit discrete representations (e.g., binary or ternary values), substantially accelerating inference on specialized hardware~\cite{ignatov2018ai}. 

However, training quantized neural networks is fundamentally a discrete combinatorial optimization problem~\cite{jacob2018quantization}. Conventional training strategies circumvent discreteness through continuous relaxations, dividing broadly into post-training quantization (PTQ) and quantization-aware training (QAT). PTQ applies heuristic rounding to pre-trained continuous models (e.g., AdaRound~\cite{nagel2020up}), which is computationally lightweight but incurs severe accuracy degradation under low-bit quantization ($\le 4$ bits). Conversely, QAT incorporates simulated quantization during forward propagation and relies on the straight-through estimator (STE) to approximate zero-almost-everywhere gradients~\cite{liu2022nonuniform}. While effective in moderate precision regimes, these continuous gradient-based heuristics fail to globally explore the underlying combinatorial landscape and frequently settle into suboptimal discrete configurations (for a comprehensive review and systematic positioning of related literature on convex neural networks, quantization paradigms, Ising optimization, and stochastic decomposition, see Supplementary Section~\ref{sec:supp-related-work}).

Physical Ising machines and dedicated annealers provide a natural paradigm for navigating complex combinatorial landscapes by encoding optimization problems into the ground states of Ising Hamiltonians~\cite{wang2013coherent,inagaki2016coherent,honjo2021100,mohseni2022ising,kleyko2023efficient,yue2024scalable,takesue2025finding}. Leveraging coherent optical and non-equilibrium physical dynamics, Ising architectures have shown compelling empirical performance across statistical learning, computer vision, molecular docking, and combinatorial optimization~\cite{bohm2022noise,yamashita2023low,laydevant2024training,niazi2024training,kim2019leveraging,li2025unified,zha2023encoding,li2020quantum,birdal2021quantum}. A pioneering effort to harness Ising optimization for neural network training is the complete quantum neural network (CQNN) framework~\cite{abel2022completely}, which compiles weights, activations, and loss minimization into a single quadratic unconstrained binary optimization (QUBO) Hamiltonian.

Nevertheless, CQNN exhibits severe structural limitations in its handling of activation functions. To conform to quadratic Hamiltonian interactions, CQNN approximates non-linear activations (e.g., ReLU, sigmoid, or tanh) via low-degree polynomials. According to classical approximation theory, multilayer feedforward networks with non-polynomial activations are dense in $C(\mathbb{R}^d)$ under the compact-open topology, ensuring universal approximation~\cite{leshno1993multilayer,hornik1991approximation}. At fixed depth, polynomial activations instead restrict the network output to polynomials of bounded degree. Furthermore, reducing high-degree polynomial interactions to quadratic forms requires extensive ancillary spins for degree reduction, creating a prohibitive physical spin overhead. Adapting expressive, non-polynomial activations to Ising optimization without sacrificing approximation power or inflating spin overhead therefore remains an essential challenge.

Before addressing this challenge algorithmically, we characterize the geometry that makes global training difficult. For continuous piecewise-polynomial networks, we prove that a generic zero-loss set is a finite stratified union: its interior strata are smooth manifolds of codimension $Nm$, whereas the non-smooth breakpoint strata have strictly higher codimension and zero top-dimensional Hausdorff measure. Local smoothness, however, does not imply a globally navigable landscape. An explicit overparameterized ReLU construction has two disconnected zero-loss components separated by a parameter hypersurface containing no interpolating solution. The obstruction is therefore topological as well as differential: overparameterization can create many global minimizers without providing a connected path between them. This motivates an exact global formulation that does not rely on traversing a single smooth basin.

A second, more profound bottleneck arises in constraint handling. Ising hardware natively minimizes unconstrained quadratic energies, whereas an exact encoding of neural-network training contains a large system of coupled consistency constraints across samples, layers, and activation intervals. The standard interface resolves this mismatch by penalizing constraint violations into a single unconstrained QUBO Hamiltonian~\cite{lucas2014ising}. However, this penalty-scalarization strategy suffers from fundamental physical and algorithmic pathologies that extend far beyond empirical hyperparameter tuning. First, enforcing strict multi-layer and inter-sample consistency requires large penalty multipliers; when mapped onto physical Ising machines with finite dynamic range, Hamiltonian normalization scales down the task loss, compressing the energy differences that distinguish competing feasible networks below the hardware's intrinsic noise threshold~\cite{mirkarimi2024quantum}. Second, squaring affine constraint residuals $(\mathbf{a}^\top \mathbf{u} - b)^2$ introduces dense pairwise couplings among otherwise uncoupled variables, dramatically increasing graph density and inflating minor-embedding overhead on hardware graphs with sparse connectivity~\cite{pelofske2023comparing}. Finally, a single scalarized Hamiltonian entangles mathematical model exactness, physical constraint violation, and stochastic hardware errors, precluding certified convergence guarantees under physical noise.

Our route to such a formulation is to use QCBO as an architecture-independent intermediate representation. For any prescribed finite-depth feedforward architecture with finite quantized parameter codebooks and FIP-certified bounded states, continuous piecewise-polynomial activations and losses admit a finite exact QCBO encoding. More general continuous nonlinearities, including sigmoid and tanh, can be approximated uniformly by piecewise-linear or piecewise-polynomial splines on the propagated compact ranges; in this case, the subsequent optimization formulation is exact relative to the chosen spline model, with the approximation error controlled separately.

Copositive programming (CPP) then provides a principled route from this QCBO representation to global convex optimization~\cite{burer2009copositive,dur2010copositive,bomze2022two}. Rather than distorting the objective landscape with heuristic penalty potentials, Burer's copositive representation theorem~\cite{burer2009copositive} establishes that the bounded mixed-binary QCBO model admits an \emph{exact completely positive lift}: the optimal value and the convex hull of globally optimal discrete networks are rigorously preserved. Crucially, this convexification does not make combinatorial NP-hardness vanish; rather, it cleanly isolates the combinatorial difficulty within the geometry of the completely positive cone $\mathcal{C}^*$, while all architectural consistency, activation interval branching, and bilinear relations become explicit affine moment identities.

This scope distinguishes the formulation from two prominent lines of prior work. Exact sign-pattern convexifications exploit the positive homogeneity of ReLU and are developed primarily for shallow architectures; their extension requires enumerating data-dependent activation patterns~\cite{pilanci2020neural,ergen2021revealing,sahiner2021vector}. Standard semidefinite formulations are more generic but usually obtain tractability by relaxing integrality or rank structure, and therefore need not preserve the global optimum~\cite{bartan2021training,prakhya2025convex}. In contrast, the completely positive lift is exact for the bounded QCBO compiled from arbitrary depth and width within the activation class above. Its significance is thus not a claim of polynomial-time training: exact completely positive cone membership remains intractable in general. The gain is a convex global-optimality representation that separates architectural constraints from the hard conic oracle, supports systematic inner or outer approximations, and provides a mathematically clean interface for hybrid Ising--classical algorithms.

This geometric isolation establishes a natural, modular Ising--classical division of labor. Classical primal--dual and augmented Lagrangian routines deterministically maintain the exact affine moment constraints, completely eliminating penalty-induced distortions, energy compression, and graph fill-in. Simultaneously, the physical Ising machine is deployed exclusively for its native capability: serving as an unconstrained Boolean Linear Minimization Oracle (LMO) over the moment polytope~\cite{yurtsever2022q}. This clean decoupling enables the first rigorous convergence proofs for Quantum Conditional Gradient Descent (QCGD) under stochastic hardware noise, bounded oracle variance, and finite coefficient precision~\cite{locatello2017unified,dunn1978conditional}. Furthermore, the algebraic block structure of the completely positive cone naturally admits sample-wise Decomposed Lower-Bound Optimization (DLBO)~\cite{bomze2022two,gabl2023sparse}, shrinking the logical spin requirement of each Ising subproblem from dataset scale $\mathcal{O}(N)$ to single-sample scale $\mathcal{O}(1)$ and providing the conceptual bridge from an exact global model to scalable, noise-aware Ising optimization algorithms. Finally, to bridge continuous lifted conic solutions back to valid discrete parameters, we introduce a Spectral-ADMM rounding scheme with continuous Moore--Penrose slack compensation, recovering strictly feasible mixed-binary network weights that tightly track the global convex relaxation bound.

In this study, we present an end-to-end framework for training quantized neural networks using coherent Ising machines (CIM) and hybrid algorithms. Our main contributions are summarized as follows:
\begin{itemize}

\item \emph{Topological diagnosis and broad exact conic convexification}:
We first characterize the stratified topology of the zero-loss set for continuous piecewise-polynomial networks. Generic interior fibers are smooth manifolds of codimension $Nm$, but an explicit overparameterized ReLU example proves that globally optimal components can remain disconnected. We then compile finite-depth, arbitrary-width networks with FIP-bounded states into a bounded QCBO and derive an exact completely positive formulation that preserves the global value and the convex hull of optimal solutions. The encoding is exact for continuous piecewise-polynomial activations and losses and accommodates more general continuous nonlinearities through controlled spline approximation. Relative to shallow ReLU sign-pattern convexifications and potentially inexact SDP relaxations, this QCBO-to-conic route substantially broadens architectural and activation scope while making the residual hardness explicit as completely positive cone membership. We further connect the binary description length of the compiled model to both statistical generalization and its logical-spin footprint.

\item \emph{Sample-wise DLBO, Hamiltonian locality hierarchy, and noise-robust QCGD engine}:
To overcome the physical pathologies of penalty-based QUBO scalarization (energy resolution collapse and coupling fill-in) and scale exact conic convexification to large datasets, we formulate sample-wise Decomposed Lower-Bound Optimization (DLBO) with two complementary solvers. Parallel Quantum Progressive Hedging (QPH--QCGD) reduces the per-subproblem logical spin requirement from dataset scale $\mathcal{O}(N)$ to single-sample scale $d_s = n + m_s$ with zero cross-sample second-moment fill-in. We prove that the DLBO moment hierarchy is simultaneously a Hamiltonian-locality hierarchy: order two is the highest level whose QCGD oracle is uniformly an auxiliary-free pairwise QUBO, whereas higher orders ($r > 2$) systematically trade higher-body spin interactions (PUBO) for tighter bounds, reaching exactness at full order. Furthermore, we establish the theoretical robustness and convergence of Quantum Conditional Gradient Descent (QCGD) under physical hardware noise: proving non-asymptotic $\mathcal{O}(T^{-1/2})$ convergence rates and almost-sure convergence under stochastic additive noise, bounded oracle variance, and finite coefficient precision, alongside an upper bound on Time-To-Solution (TTS) and certified copositive column generation with explicit local optimality intervals.

\item \emph{Closed-loop discrete recovery via Spectral--ADMM rounding and physical CIM validation}:
To bridge continuous lifted moment solutions back to valid discrete parameters, we introduce an inexact non-convex Spectral--ADMM rounding scheme paired with a Moore--Penrose continuous slack repair operator ($\mathsf R_{\mathrm{QCBO}}$), efficiently recovering strictly feasible mixed-binary network parameters that tightly track the global convex relaxation bound.  Extensive benchmark evaluations on a physical Coherent Ising Machine (CIM) demonstrate that our framework achieves $94.95\%$ accuracy on Fashion-MNIST with ultra-low $1.1$-bit quantization.  Compared to classical quantization-aware training (QAT), training speed is accelerated by orders of magnitude; compared to post-training quantization (PTQ), it eliminates low-bit representational collapse while substantially reducing inference memory footprint and latency.
\end{itemize}
\section{Results}
\subsection{Framework overview}

The approach first conceptualizes a quantized neural network (QNN) as a parameterized operator $\mathcal{Q}: \mathbb{R}^d \to \mathbb{R}$, which transforms an input space $\mathbb{R}^d$ into the output space $\mathbb{R}$ through a parameter set $\mathcal{W}$ drawn from a discrete parameter space. This operator emerges from a hierarchical cascade of quantized transformations, reflecting the network's layered and constrained architecture.

The global operator $\mathcal{Q}(x, \mathcal{W})$ could be expressed as a nested sequence of $L$ quantized layers:
\begin{align}
\mathcal{Q}(x, \mathcal{W}) = \mathcal{Q}_L (\mathcal{Q}_{L-1} (\cdots \mathcal{Q}_1 (x, \mathcal{W}_1), \mathcal{W}_2) \cdots , \mathcal{W}_L),
\end{align}
where $L$ denotes the depth of the network, and $\mathcal{W}$ encompasses the quantized parameters across all layers. For each layer $l$ ($1\le l\le L$), the parameter set $\mathcal{W}_l = (\mathbf{W}_l, \mathbf{b}_l)$ consists of a quantized weight matrix $\mathbf{W}_l \in \mathbb{R}^{m_l \times m_{l-1}}$ and a quantized bias vector $\mathbf{b}_l \in \mathbb{R}^{m_l}$, with input dimension $m_0 = d$ and output dimension $m_L = 1$.

Each layer's transformation $\mathcal{Q}_l$ operates on the output of the preceding layer:
\begin{align}
\mathcal{Q}_l(x, \mathcal{W}_l) = \sigma (\mathbf{W}_l x + \mathbf{b}_l),
\end{align}
where $\sigma: \mathbb{R} \to \mathbb{R}$ denotes the activation function applied elementwise.

The coordinate-wise pre-activation $z_j^{(l)}$ and post-activation $a_j^{(l)}$ of the $j^{\text{th}}$ neuron in layer $l$ satisfy:
\begin{align}
a_j^{(l)} = \sigma\left(z_j^{(l)}\right), \qquad z_j^{(l)} = \sum_{k=1}^{m_{l-1}} W_{jk}^{(l)} a_k^{(l-1)} + b_j^{(l)} \quad (l \in [L], \; j \in [m_l]),
\end{align}
with initial input $a_k^{(0)} = x_k$.


\begin{figure}[htbp]
    \centering
    \includegraphics[width=\textwidth]{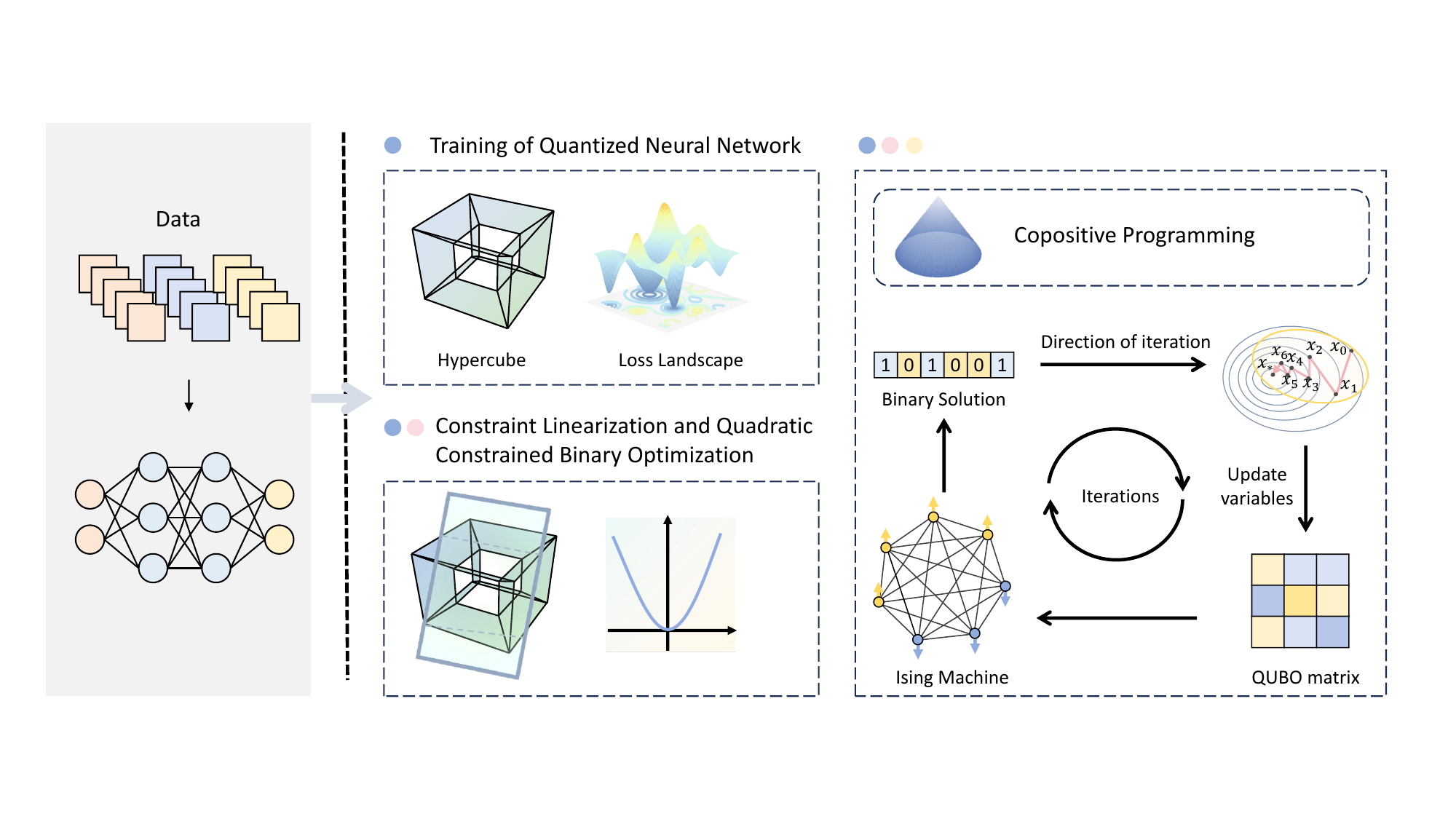}
    \caption{The schematic workflow of the quantized neural network training framework using coherent Ising machines (CIM) and hybrid algorithms.}
    \label{workflow}
\end{figure}

Figure~\ref{workflow} illustrates the overarching architecture of the Ising--classical hybrid training framework. Starting from input training data, the network feedforward dynamics and activation nonlinearities are discretized via Forward Interval Propagation (FIP). The resulting bounded Quadratic Constrained Binary Optimization (QCBO) model is lifted into an exact completely positive program. In each iteration of the hybrid solver (QCGD or parallel QPH--QCGD), classical routines perform primal--dual and consensus updates to enforce affine moment constraints, while coherent Ising machines (CIM) or physical Ising hardware solve unconstrained QUBO subproblems as linear minimization oracles, ensuring certified convergence toward the global discrete optimum.

\subsection{Spline-induced neural-network formulation}

We use piecewise-linear activations because they admit exact finite encodings while retaining the approximation rates of ReLU networks, subject to the qualifications below.
\begin{fact}[Quantized ReLU and PWL transfer~\cite{dinguniversal,yarotsky2017error}]
\label{universal_approximation_theorem}
Let $d,n\in\mathbb N$, $\lambda\geq2$ be an integer, and
\begin{align}
\mathcal F_{d,n}
  :=\left\{f\in\mathcal W^{n,\infty}([0,1]^d):
  \max_{|\boldsymbol\alpha|\leq n}
  \operatorname*{ess\,sup}_{\mathbf x\in[0,1]^d}
  \bigl|D^{\boldsymbol\alpha}f(\mathbf x)\bigr|\leq1\right\}.
\end{align}
For every $f\in\mathcal F_{d,n}$ and $\epsilon\in(0,1)$, there exist a codebook of $\lambda$ real values and a ReLU network $f_{q,\mathrm{ReLU}}$ whose weights belong to that codebook such that
\begin{align}
\bigl\|f-f_{q,\mathrm{ReLU}}\bigr\|_{L^\infty([0,1]^d)}\leq\epsilon.
\end{align}
The construction of Ding et al.~\cite{dinguniversal} has depth and weight count
\begin{align}
L_q
  &=\mathcal O_{d,n}\!\left(
    \lambda\log^{\frac{1}{\lambda-1}}(1/\epsilon)
    +\log(1/\epsilon)\right),\\
W_q
  &=\mathcal O_{d,n}\!\left(
    \lambda\log^{1+\frac{1}{\lambda-1}}(1/\epsilon)
    (1/\epsilon)^{d/n}\right).
\end{align}
Now fix a continuous piecewise-linear activation
$\rho:\mathbb R\to\mathbb R$ with finitely many pieces and at least one
genuine breakpoint.  Proposition~1(b) of Yarotsky~\cite{yarotsky2017error}
yields a generally real-weight $\rho$-network $f_\rho$ of depth $L_q$ and
at most $4W_q$ weights such that
\begin{align}
f_\rho(\mathbf x)=f_{q,\mathrm{ReLU}}(\mathbf x)
\quad\text{for every }\mathbf x\in[0,1]^d.
\end{align}
Consequently, $f_\rho$ attains the same approximation error and preserves
the ReLU weight-count rate up to a constant factor.
\end{fact}
Fact~\ref{universal_approximation_theorem} therefore separates two claims:
Ding et al.~\cite{dinguniversal} establish universal approximation and
complexity bounds for $\lambda$-level quantized \emph{ReLU} networks, whereas
Yarotsky~\cite{yarotsky2017error} transfers ReLU approximation rates to any
fixed continuous, non-affine PWL activation using unrestricted real weights.
For fixed $\lambda$, Ding et al.'s constructive upper bound is within an
$\mathcal O(\log^5(1/\epsilon))$ factor of the compatible unquantized ReLU
lower bound.  Their bound-based calculation further suggests bit-widths
between one and four in the parameter regimes examined, but it is not an
activation-independent optimal-bit theorem.  Our optimization formulation can
encode a chosen PWL activation together with any prescribed finite parameter
codebook; a corresponding fixed-codebook approximation theorem for that
activation would require an additional quantization argument.


While Fact~\ref{universal_approximation_theorem} transfers ReLU approximation
rates to continuous non-affine PWL activations, explicitly bounding the
interpolation error when replacing smooth activations with PWL interpolants
provides rigorous non-asymptotic control; we establish this complete error bound
in Supplementary Section~\ref{sec:log_pwl_supp}.

Given a dataset of $N$ training samples $\{(\mathbf{x}^{(s)}, y^{(s)})\}_{s=1}^N$, empirical risk minimization over the quantized parameter space $\mathcal{W}_{\text{quant}}$ is formulated as:
\begin{align}
\min_{\mathcal{W} \in \mathcal{W}_{\text{quant}}} \sum_{s=1}^N \mathcal{L}\left(a^{(s,L)}, y^{(s)}\right),
\end{align}
subject to the quantized feedforward layer dynamics for each sample $s \in [N]$:
\begin{align}
a_j^{(s,\ell)} &= \sigma \left(z_j^{(s,\ell)}\right), \qquad z_j^{(s,\ell)} = \sum_{k=1}^{m_{\ell-1}} W_{jk}^{(\ell)} a_k^{(s,\ell-1)} + b_j^{(\ell)} \quad (\ell \in [L], \; j \in [m_\ell]),
\end{align}
where $\mathbf{a}^{(s,0)} = \mathbf{x}^{(s)}$ serves as the input, $a^{(s,L)}$ is the network scalar output, $y^{(s)}$ represents the target label, and $\mathcal{W}_{\text{quant}}$ denotes the set of permissible quantized parameter configurations. This formulation captures the fundamental trade-off between task accuracy and discrete computational efficiency.


\subsubsection{Forward interval propagation and exact QCBO encoding}

Forward interval propagation (FIP) bounds every pre-activation and partitions
the resulting compact range into finitely many intervals.  On each interval
we use the straight line joining the two endpoint values of the activation;
the terminal loss is treated in the same way.  Thus the approximation is
piecewise linear and interpolates the function at every breakpoint.  In
particular, no interval is replaced by the average of its endpoint values.
The one-hot interval selector used below keeps all cross-layer consistency
conditions explicit in a single finite mixed-binary optimization problem.

For sample $s$ and layer $\ell$, let
$\mathbf z^{(s,\ell)},\mathbf a^{(s,\ell)}\in\mathbb R^{m_\ell}$ denote the
pre- and post-activations, with
$\mathbf a^{(s,0)}=\mathbf x^{(s)}$. FIP supplies finite bounds
$\underline a_k^{(s,\ell-1)}\le a_k^{(s,\ell-1)}
\le\overline a_k^{(s,\ell-1)}$. Quantized parameters are affine in binary
code variables:
\begin{align}
 \operatorname{vec}(\mathbf W^{(\ell)})
 &=\mathbf C_W^{(\ell)}\boldsymbol\delta_W^{(\ell)}
   +\mathbf d_W^{(\ell)},&
 \mathbf b^{(\ell)}
 &=\mathbf C_b^{(\ell)}\boldsymbol\delta_b^{(\ell)}
   +\mathbf d_b^{(\ell)},                                \label{eq:affine-binary-codebook}
\end{align}
where every coordinate of
$\boldsymbol\delta_W^{(\ell)}$ and
$\boldsymbol\delta_b^{(\ell)}$ is binary. This representation includes the
one-hot codebooks described in
Section~\ref{sec:multilevel-binary-codebook} as well as ordinary positional
binary codes.


\begin{figure}[htbp]
 \begin{subfigure}{0.49\textwidth}
        \centering
\includegraphics[width=\textwidth]{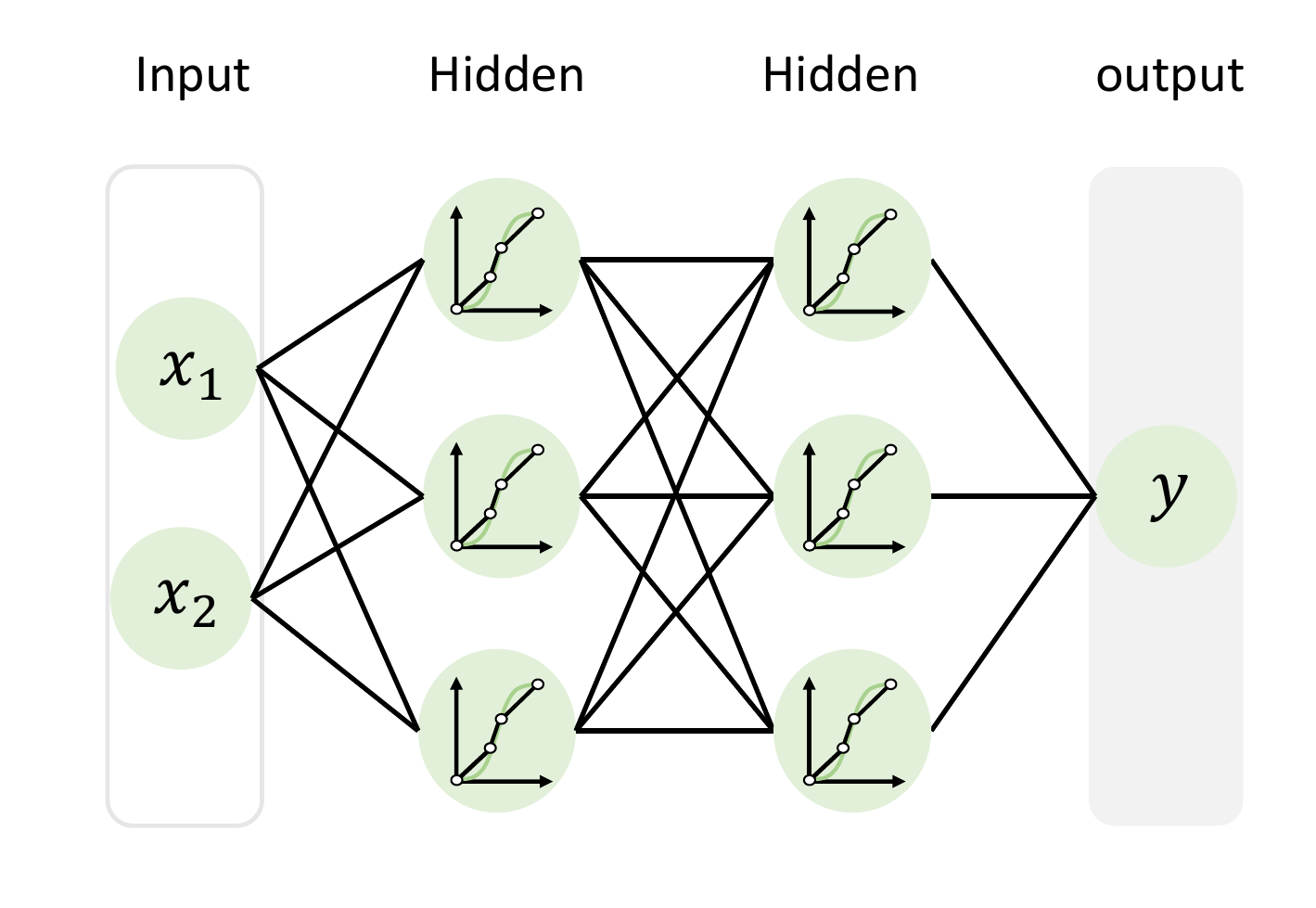}
    \caption{}
    \label{NN}
        \end{subfigure}
            \hfill
\begin{subfigure}{0.49\textwidth}
        \centering
\includegraphics[width=\textwidth]{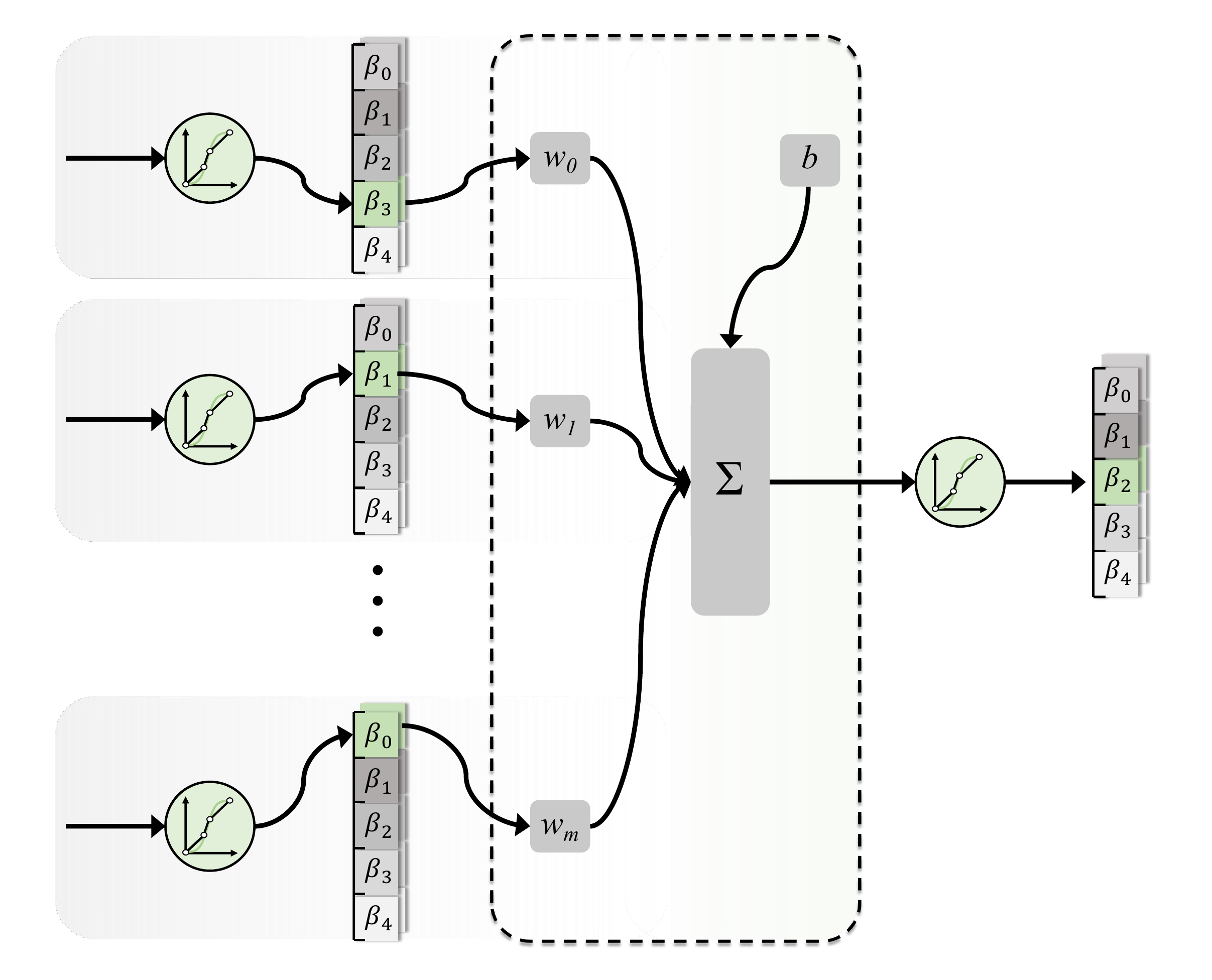}
\caption{}
    \label{NN_pwl}
        \end{subfigure}
\caption{(a) Feedforward network. (b) FIP partitions each bounded
pre-activation range into intervals. A one-hot variable $\beta_i$ selects one
segment, and a local interpolation coordinate evaluates the corresponding
linear activation before propagation through the next affine layer.}
\label{fip}
\end{figure}

For a positive integer $d$, write $[d]:=\{1,\ldots,d\}$.  We partition
$[\underline M,\overline M]$ at strictly increasing breakpoints
$\underline M=M_0<M_1<\cdots<M_n=\overline M$. Define
$\Delta M_i:=M_i-M_{i-1}$ and
$\Delta\sigma_i:=\sigma(M_i)-\sigma(M_{i-1})$.  For each neuron introduce a
one-hot segment selector $\boldsymbol\beta_z$ and local interpolation
coordinates $\boldsymbol\theta_z$ satisfying
$0\le\theta_{z,i}\le\beta_{z,i}$.  When $\beta_{z,i}=1$, the coordinate
$\theta_{z,i}\in[0,1]$ moves continuously from the left to the right endpoint
of segment $i$.

For a scalar network output, let
$P_0<P_1<\cdots<P_q$ be breakpoints spanning its bounded range, put
$\Delta P_i:=P_i-P_{i-1}$, and define
\begin{align}
 L_{i,-}^{(s)}&:=\mathcal L(P_{i-1},y^{(s)}),&
 \Delta L_i^{(s)}&:=\mathcal L(P_i,y^{(s)})-
                         \mathcal L(P_{i-1},y^{(s)}).
 \label{eq:loss-endpoint-data-main}
\end{align}
The variables $\boldsymbol\beta_a^{(s)}$ and
$\boldsymbol\theta_a^{(s)}$ select and interpolate a loss segment.  The
one-hot piecewise-linear training problem is
\begin{align}
 \min\quad
 &\sum_{s=1}^{N}\sum_{i=1}^{q}
   \left(L_{i,-}^{(s)}\beta^{(s)}_{a,i}
   +\Delta L_i^{(s)}\theta^{(s)}_{a,i}\right)           \label{qcbo_obj}\\
 \text{s.t.}\quad
 &z_j^{(s,\ell)}
   =\sum_{k=1}^{m_{\ell-1}}W_{jk}^{(\ell)}
       a_k^{(s,\ell-1)}+b_j^{(\ell)},
 &&s\in[N],\ \ell\in[L],\ j\in[m_\ell],               \label{eq:fip-affine-main}\\
 &z_j^{(s,\ell)}
   =\sum_{i=1}^{n}\left(M_{i-1}\beta^{(s,\ell)}_{z,j,i}
                    +\Delta M_i\theta^{(s,\ell)}_{z,j,i}\right),
 &&s\in[N],\ \ell\in[L],\ j\in[m_\ell],               \label{quadcons}\\
 &a_j^{(s,\ell)}
   =\sum_{i=1}^{n}\left(\sigma(M_{i-1})
                    \beta^{(s,\ell)}_{z,j,i}
                    +\Delta\sigma_i
                    \theta^{(s,\ell)}_{z,j,i}\right),
 &&s\in[N],\ \ell\in[L],\ j\in[m_\ell],               \label{eq:fip-value-main}\\
 &0\le\theta^{(s,\ell)}_{z,j,i}
       \le\beta^{(s,\ell)}_{z,j,i},
 &&s\in[N],\ \ell\in[L],\ j\in[m_\ell],\ i\in[n],   \label{eq:fip-local-main}\\
 &a^{(s,L)}
   =\sum_{i=1}^{q}\left(P_{i-1}\beta^{(s)}_{a,i}
                    +\Delta P_i\theta^{(s)}_{a,i}\right),
 &&s\in[N],                                                \label{eq:fip-output-main}\\
 &0\le\theta^{(s)}_{a,i}\le\beta^{(s)}_{a,i},
 &&s\in[N],\ i\in[q],                                    \label{eq:fip-loss-local-main}\\
 &\sum_{i=1}^{n}\beta^{(s,\ell)}_{z,j,i}=1,
 &&s\in[N],\ \ell\in[L],\ j\in[m_\ell],                 \notag\\
 &\sum_{i=1}^{q}\beta^{(s)}_{a,i}=1,
 &&s\in[N],                                                \notag\\
 &\beta^{(s,\ell)}_{z,j,i}\in\{0,1\},
 &&s\in[N],\ \ell\in[L],\ j\in[m_\ell],\ i\in[n],     \notag\\
 &\beta^{(s)}_{a,i}\in\{0,1\},
 &&s\in[N],\ i\in[q].
 \label{qcbo_last}
\end{align}

If segment $i$ is selected, the two equalities above reduce to
$z=M_{i-1}+\Delta M_i\theta$ and
$a=\sigma(M_{i-1})+\Delta\sigma_i\theta$. Eliminating $\theta$ therefore
gives the linear interpolant between
$(M_{i-1},\sigma(M_{i-1}))$ and $(M_i,\sigma(M_i))$. The same argument shows
that \eqref{qcbo_obj} is the piecewise-linear interpolation of the loss from
its endpoint values. At a shared breakpoint either adjacent selector is
permitted, but both representations return the same activation or loss value.

It remains to express the products in \eqref{eq:fip-affine-main} without
relaxation. Write the coordinate form of \eqref{eq:affine-binary-codebook} as
\begin{align}
 W_{jk}^{(\ell)}
 &=d_{W,jk}^{(\ell)}+
   \sum_{r=1}^{R_\ell}c_{W,jk,r}^{(\ell)}
   \delta_{W,jk,r}^{(\ell)},&
 b_j^{(\ell)}
 &=d_{b,j}^{(\ell)}+
   \sum_{r=1}^{R_{b,\ell}}c_{b,j,r}^{(\ell)}
   \delta_{b,j,r}^{(\ell)}.                              \label{eq:coordinate-binary-codebook}
\end{align}
For every product between a weight bit and a bounded activation, introduce
$v_{jk,r}^{(s,\ell)}$ and impose
\begin{align}
 \underline a_k^{(s,\ell-1)}\delta_{W,jk,r}^{(\ell)}
 &\le v_{jk,r}^{(s,\ell)}
 \le\overline a_k^{(s,\ell-1)}\delta_{W,jk,r}^{(\ell)},\notag\\
 a_k^{(s,\ell-1)}-
 \overline a_k^{(s,\ell-1)}
 \bigl(1-\delta_{W,jk,r}^{(\ell)}\bigr)
 &\le v_{jk,r}^{(s,\ell)}\notag\\
 &\le a_k^{(s,\ell-1)}-
 \underline a_k^{(s,\ell-1)}
 \bigl(1-\delta_{W,jk,r}^{(\ell)}\bigr).                \label{eq:mcc-again}
\end{align}
When the bit is zero, the first line forces $v_{jk,r}^{(s,\ell)}=0$; when it
is one, the second pair forces
$v_{jk,r}^{(s,\ell)}=a_k^{(s,\ell-1)}$. Hence
\eqref{eq:mcc-again} is the exact graph of
$v_{jk,r}^{(s,\ell)}=
\delta_{W,jk,r}^{(\ell)}a_k^{(s,\ell-1)}$, rather than a continuous
relaxation. The affine layer equation is therefore equivalently
\begin{align}
 z_j^{(s,\ell)}
 ={}&\sum_{k=1}^{m_{\ell-1}}
 \left(d_{W,jk}^{(\ell)}a_k^{(s,\ell-1)}
 +\sum_{r=1}^{R_\ell}c_{W,jk,r}^{(\ell)}
 v_{jk,r}^{(s,\ell)}\right)\notag\\
 &+d_{b,j}^{(\ell)}
 +\sum_{r=1}^{R_{b,\ell}}c_{b,j,r}^{(\ell)}
 \delta_{b,j,r}^{(\ell)}.                               \label{eq:fip-affine-linearized-main}
\end{align}

Collect the code bits, one-hot segment selectors, local interpolation
coordinates, bounded activations and pre-activations, and product variables
into a finite vector. Shift every
bounded continuous coordinate to the nonnegative orthant, introduce an
upper-bound complement for every coordinate, and equalize each linear
inequality with a nonnegative slack. The complete FIP model then has the
bounded mixed-binary quadratic normal form
\begin{align}
 \min_{\mathbf u\ge\mathbf{0}}\quad
 &\mathbf u^\top\mathbf Q\mathbf u+2\mathbf c^\top\mathbf u\notag\\
 \text{s.t.}\quad
 &\mathbf A\mathbf u=\mathbf b,
 \qquad u_i\in\{0,1\}\quad(i\in\mathcal B).             \label{eq:main-qcbo-normal-form}
\end{align}
For the piecewise-linear loss in \eqref{qcbo_obj}, $\mathbf Q=\mathbf{0}$;
quadratic regularizers or an exactly retained quadratic loss populate
$\mathbf Q$ without changing the form. The complement equations make the
continuous relaxation
$\{\mathbf u\ge\mathbf{0}:\mathbf A\mathbf u=\mathbf b\}$ compact. Thus
\eqref{qcbo_obj}--\eqref{qcbo_last},
\eqref{eq:affine-binary-codebook}, and \eqref{eq:mcc-again} explicitly give
the QCBO required by the completely positive lift. No penalty reduction is used; the primitive-vector ordering, expanded block matrices, and dimensional bookkeeping for the completely positive lift are formalized below in Section~\ref{subsubsec:global-convexification} and Methods.
Here and below, ``exact'' means exact relative to the finite
FIP-discretized problem; the activation and loss discretization errors are
controlled separately.

For clarity, Section~2.2.1 uses the unary one-hot encoding, requiring $n$
binary selector variables per activation. This is not the only exact
encoding. Supplementary Section~\ref{sec:log_pwl_supp} gives a self-contained
logarithmic disaggregated formulation that represents the same piecewise-linear
graph with $\lceil\log_2 n\rceil$ selector bits, together with its linking
constraints, equivalence proof, loss construction, and interaction with the
weight--activation linearization.


\subsubsection{From multilevel quantization to binary variables}
\label{sec:multilevel-binary-codebook}

The affine Boolean representation in \eqref{eq:affine-binary-codebook} maps
multilevel discrete parameters into binary variables for arbitrary finite
codebooks.  For layer $\ell \in [L]$, let each weight coordinate $W_{jk}^{(\ell)}$
($(j,k) \in [m_\ell] \times [m_{\ell-1}]$) and bias coordinate $b_j^{(\ell)}$
($j \in [m_\ell]$) be restricted to finite discrete alphabets
$\mathcal{S}_{W,jk}^{(\ell)} \subset \mathbb{R}$ and
$\mathcal{S}_{b,j}^{(\ell)} \subset \mathbb{R}$ of cardinality $|\mathcal{S}_{W,jk}^{(\ell)}| = K_{W}$
and $|\mathcal{S}_{b,j}^{(\ell)}| = K_{b}$, respectively.  We formalize two
canonical exact linear embeddings into Boolean domains $\{0,1\}^B$:

\paragraph{Unary simplex encoding (arbitrary discrete alphabets).}
For arbitrary, non-uniform, or asymmetric codebooks
$\mathcal{S}_{W,jk}^{(\ell)} = \{s_1, \dots, s_K\} \subset \mathbb{R}$ (such as
$k$-means centroid quantization \cite{bartan2021training}), introduce a one-hot
indicator block
$\mathbf{y}_{jk}^{(\ell)} = (y_{jk,1}^{(\ell)}, \dots, y_{jk,K}^{(\ell)})^\top \in \{0,1\}^K$.
The scalar parameter is then the exact linear form
\begin{align}
 W_{jk}^{(\ell)} = \sum_{r=1}^K s_r y_{jk,r}^{(\ell)},
 \qquad
 \text{subject to } \sum_{r=1}^K y_{jk,r}^{(\ell)} = 1.
 \label{eq:unary-codebook-encoding}
\end{align}
This embedding accommodates arbitrary code alphabets at the cost of $K$ binary
variables and one affine cardinality constraint per scalar parameter.

\paragraph{Positional radix encoding (uniform and structured grids).}
When $\mathcal{S}_{W,jk}^{(\ell)}$ is a uniform $B$-bit fixed-point grid of
$K = 2^B$ levels with step size $\Delta > 0$ and origin shift $\gamma \in \mathbb{R}$,
namely
$\mathcal{S}_{W,jk}^{(\ell)} = \{ \gamma + \Delta \sum_{b=0}^{B-1} 2^b u_b : u_b \in \{0,1\} \}$,
the coordinate admits an unconstrained binary radix expansion:
\begin{align}
 W_{jk}^{(\ell)}
 = \gamma + \Delta \sum_{b=0}^{B-1} 2^b \delta_{W,jk,b}^{(\ell)}
 = d_{W,jk}^{(\ell)} + \sum_{b=0}^{B-1} c_{W,jk,b}^{(\ell)} \delta_{W,jk,b}^{(\ell)},
 \label{eq:radix-codebook-encoding}
\end{align}
where $\boldsymbol\delta_{W,jk}^{(\ell)} \in \{0,1\}^B$,
$c_{W,jk,b}^{(\ell)} = \Delta 2^b$, and $d_{W,jk}^{(\ell)} = \gamma$.
Crucially, this logarithmic formulation uses exactly $B = \lceil \log_2 K \rceil$
binary variables per parameter and imposes \emph{zero} auxiliary equality
constraints on the parameter code.  Standard signed symmetric quantization
$W_{jk}^{(\ell)} \in \{-\Delta(2^{B-1}-1), \dots, \Delta(2^{B-1}-1)\}$ is recovered
by setting $\gamma = -\Delta(2^{B-1}-1)$, while extreme-low-bit regimes---including
1-bit binary ($\{-1, +1\}$ with $B=1, \gamma=-1, \Delta=2$) and ternary
($\{-1, 0, +1\}$ with $B=2$)---are direct instantiations.

\paragraph{Canonical matrix vectorization.}
Stacking the scalar codes across all $m_\ell m_{\ell-1}$ weights and $m_\ell$
biases of layer $\ell$ yields the global affine block representation of
\eqref{eq:affine-binary-codebook}:
\begin{align}
 \operatorname{vec}(\mathbf W^{(\ell)})
 = \mathbf C_W^{(\ell)} \boldsymbol\delta_W^{(\ell)} + \mathbf d_W^{(\ell)},
 \qquad
 \mathbf b^{(\ell)}
 = \mathbf C_b^{(\ell)} \boldsymbol\delta_b^{(\ell)} + \mathbf d_b^{(\ell)},
 \label{eq:block-matrix-vectorization}
\end{align}
where $\boldsymbol\delta_W^{(\ell)} \in \{0,1\}^{B m_\ell m_{\ell-1}}$ and
$\mathbf C_W^{(\ell)} = \mathbf{I}_{m_\ell m_{\ell-1}} \otimes (c_{W,1}^{(\ell)}, \dots, c_{W,B}^{(\ell)}) \in \mathbb{R}^{m_\ell m_{\ell-1} \times B m_\ell m_{\ell-1}}$
is a block-sparse decoding matrix.  This establishes the parameter bit dimension
$B$ that governs the logical-spin complexity $O(BP)$ in Table~\ref{tab:logical-spin-count}
and the sample complexity bound in Theorem~\ref{thm:finite-class-sample-complexity}.

\subsubsection{Logical-spin and statistical sample complexity}
\label{sec:spin-sample-complexity}

The size of a compiled training instance and the number of observations needed
for generalization are distinct complexity measures.  We quantify them
separately before relating them.  Let
\begin{align}
 P&:=\sum_{\ell=1}^{L}m_\ell(m_{\ell-1}+1),&
 U&:=\sum_{\ell=1}^{L}m_\ell,
 \label{eq:parameter-unit-counts}
\end{align}
denote the number of trainable scalar parameters and the number of non-input
units, respectively.  Let $B$ upper-bound the number of code bits per
parameter, and let $n$ be the number of FIP intervals.  Before hardware
embedding, assume each remaining bounded coordinate is represented by a
fixed-size finite code, as required by the Boolean LMO.  The monolithic
$N$-sample formulation then has logical-variable footprint
\begin{align}
 S_{\mathrm{logical}}(N)
 =O\!\left(BP+nNU+NBP\right).                            \label{eq:logical-spin-bound}
\end{align}
The three terms account for parameter codes, sample-wise one-hot
selector/interpolation blocks, and sample--parameter product/linearization
blocks.  Bound complements and
slacks change only the constant factor under a fixed finite encoding.  For a
fully connected network with $m_\ell\le W$, this becomes
\begin{align}
 S_{\mathrm{logical}}(N)
 =O\!\left(BLW^2+nNLW+NBLW^2\right).                    \label{eq:width-spin-bound}
\end{align}
Thus the monolithic footprint is linear in $N$ but generally quadratic, not
linear, in the maximum width.  These are logical spins; a hardware minor
embedding can require additional physical spins depending on the device
topology.  If a nonbinary coordinate instead uses a growing $b_c$-bit
encoding, its corresponding block must be multiplied by $b_c$.

\begin{table}[h!]
\centering
\begin{tabular}{@{}lll@{}}
\toprule
Block & General count & Uniform-width bound \\ \midrule
Parameter code bits & $O(BP)$ & $O(BLW^2)$ \\
FIP selector/interpolation blocks & $O(nNU)$ & $O(nNLW)$ \\
Product/linearization variables & $O(NBP)$ & $O(NBLW^2)$ \\
Complements and slacks & same asymptotic order & same asymptotic order \\ \bottomrule
\end{tabular}
\caption{Logical-variable footprint of the monolithic finite encoding.  The
count excludes topology-dependent overhead from embedding the logical QUBO on
physical hardware.}
\label{tab:logical-spin-count}
\end{table}

Quantization precision has two opposing effects.  Increasing $B$ can reduce
representation error, but it enlarges both the compiled QCBO and the finite
hypothesis class.  We therefore ask how the binary description length that
determines the logical-spin footprint translates into a sufficient sample
size.  For a fixed FIP table, let $\Theta_B$ be the finite set of quantized
parameter configurations, let
$\mathcal H_B:=\{f_\theta:\theta\in\Theta_B\}$ be the induced predictor
class, and define
\begin{align}
 \mathcal R(\theta)
 &:=\mathbb E_{(x,y)\sim\mathcal P}[C(f_\theta(x),y)],&
 \widehat{\mathcal R}_N(\theta)
 &:=\frac1N\sum_{i=1}^{N}C(f_\theta(x_i),y_i).
 \label{eq:population-empirical-risk}
\end{align}

\begin{theorem}[Description-length excess-risk bound for quantized networks]
\label{thm:finite-class-sample-complexity}
Assume that the loss is bounded as
\begin{align}
 0\le C(f_\theta(x),y)\le\mathcal L_{\max}
 \qquad\text{for every }\theta\in\Theta_B.
\end{align}
Let
$\theta_B^\star\in\arg\min_{\theta\in\Theta_B}\mathcal R(\theta)$.  Suppose
$\widehat\theta_N$ has empirical suboptimality at most
$\varepsilon_{\mathrm{opt}}$, that is,
\begin{align}
 \widehat{\mathcal R}_N(\widehat\theta_N)
 \le\min_{\theta\in\Theta_B}\widehat{\mathcal R}_N(\theta)
 +\varepsilon_{\mathrm{opt}}.
\end{align}
For $\varepsilon_{\mathrm{stat}},\delta\in(0,1)$, if
\begin{align}
 N
 \ge
 \frac{2\mathcal L_{\max}^2}{\varepsilon_{\mathrm{stat}}^2}
 \left(BP\log2+\log\frac{2}{\delta}\right),             \label{eq:finite-class-sample-bound}
\end{align}
then, with probability at least $1-\delta$,
\begin{align}
 \mathcal R(\widehat\theta_N)-\mathcal R(\theta_B^\star)
 \le\varepsilon_{\mathrm{stat}}+\varepsilon_{\mathrm{opt}}.
 \label{eq:stat-opt-decomposition}
\end{align}
\end{theorem}

The concentration argument is standard: it uses
$|\Theta_B|\le2^{BP}$, Hoeffding's inequality, and a union bound;
Supplementary Section~\ref{sec:supp-sample-complexity} gives the proof.  Its
role here is operational rather than a claim of a new concentration
inequality: the same description length $BP$ controls both generalization and
the compiled optimization instance.

\begin{corollary}[End-to-end statistical and logical-spin complexity]
\label{cor:statistical-spin-complexity}
Let $\mathcal F_{\mathrm{ref}}\supseteq\mathcal H_B$ be a reference predictor
class and define its optimal risk and the representation gap by
\begin{align}
 \mathcal R_{\mathrm{ref}}^\star
 &:=\inf_{f\in\mathcal F_{\mathrm{ref}}}
 \mathbb E_{(x,y)\sim\mathcal P}[C(f(x),y)],&
 \varepsilon_{\mathrm{repr}}(B,n)
 &:=\mathcal R(\theta_B^\star)-\mathcal R_{\mathrm{ref}}^\star.
 \label{eq:representation-gap}
\end{align}
Choosing $N$ at the sufficient order in
\eqref{eq:finite-class-sample-bound} gives, with probability at least
$1-\delta$,
\begin{align}
 \mathcal R(\widehat\theta_N)-\mathcal R_{\mathrm{ref}}^\star
 \le
 \varepsilon_{\mathrm{repr}}(B,n)
 +\varepsilon_{\mathrm{stat}}
 +\varepsilon_{\mathrm{opt}},                            \label{eq:end-to-end-risk-decomposition}
\end{align}
while the monolithic logical-spin requirement obeys
\begin{align}
 S_{\mathrm{logical}}
 =O\!\left(
 BP+
 (nU+BP)\left[
 1+\frac{\mathcal L_{\max}^2}{\varepsilon_{\mathrm{stat}}^2}
 \left(BP+\log\frac1\delta\right)\right]
 \right).                                                 \label{eq:statistical-spin-composition}
\end{align}
\end{corollary}

Corollary~\ref{cor:statistical-spin-complexity} exposes the joint trade-off:
additional precision may reduce
$\varepsilon_{\mathrm{repr}}(B,n)$, but it increases the hypothesis-class
description length, the statistically sufficient sample size, and the
compiled spin footprint.  Equation~\eqref{eq:statistical-spin-composition}
follows by substituting \eqref{eq:finite-class-sample-bound} into \eqref{eq:logical-spin-bound},
quantifying the intrinsic logical spin requirement prior to hardware graph embedding.
Supplementary Section~\ref{sec:supp-sample-complexity} gives the derivation.

For binary classification, the induced class satisfies
\begin{align}
 d_{\mathrm{VC}}(\mathcal H_B)
 \le\log_2|\mathcal H_B|
 \le\log_2|\Theta_B|
 \le BP.                                                  \label{eq:vc-description-length-comparison}
\end{align}
An architecture-specific VC bound can be tighter because parameter symmetries
and infeasible encodings may map multiple bit strings to the same predictor.
For real-valued prediction, the analogous comparison uses pseudo-dimension.
The advantage of the description-length bound is therefore not universal
tightness, but its direct composability with the QCBO resource count.

\subsection{Convex formulation for neural-network training}
\label{subsec:results-convex-formulation}

\subsubsection{Stratified manifold geometry of the zero-loss level set}
\label{subsubsec:geometric-landscape}

To rigorously characterize the non-convex optimization landscape prior to convexification, we examine the geometric topology of the global zero-loss solution set.  For a training dataset $\{(\mathbf{x}^{(s)}, \mathbf{y}^{(s)})\}_{s=1}^N \subset \mathbb{R}^{m_0} \times \mathbb{R}^m$, let $L \ge 2$ denote the network depth, $P = \sum_{\ell=1}^L m_\ell(m_{\ell-1}+1)$ the total parameter dimension, and $q := Nm$ the aggregate number of scalar fitting constraints.  Identifying the dataset target space with $\mathbf{Y} \in \mathbb{R}^q$, we define the evaluation map $\Phi:\mathbb{R}^P \to \mathbb{R}^q$ by $\Phi(\boldsymbol\theta) := (f_{\boldsymbol\theta}(\mathbf{x}^{(1)}), \dots, f_{\boldsymbol\theta}(\mathbf{x}^{(N)}))^\top$.  The zero-loss level set is the fiber $\mathcal{M}_{\mathrm{total}}(\mathbf{Y}) := \Phi^{-1}(\mathbf{Y})$.

For globally smooth activations, Cooper established via Sard's theorem that generic non-empty zero-loss fibers are smooth submanifolds of codimension $q$ \cite{cooper2021loss}, a result paired with non-affine interpolation by Constantinescu and Popescu \cite{constantinescu2024approximation}.  However, this regular-value principle does not apply globally to piecewise-smooth architectures (e.g., multilinear ReLU networks \cite{laurent2018multilinear}).  Below, we establish a stratified extension for continuous piecewise-polynomial networks, proving that non-smooth boundary fibers are strictly lower-dimensional and have measure zero.

Let each hidden activation $\sigma:\mathbb{R}\to\mathbb{R}$ be continuous and piecewise polynomial with finite breakpoint set $\mathcal{B}_\sigma \subset \mathbb{R}$.  Assume each hidden neuron possesses an independent bias coordinate.  A parameter vector $\boldsymbol\theta \in \mathbb{R}^P$ is \emph{strictly interior} if $z_j^{(s,\ell)}(\boldsymbol\theta) \notin \mathcal{B}_\sigma$ for all $(s,\ell,j) \in [N] \times [L-1] \times [m_\ell]$; otherwise, it is a \emph{boundary} parameter.  This partitions $\mathbb{R}^P$ into a finite set of open semialgebraic activation cells $\{\mathcal{C}_\alpha\}_{\alpha \in \mathcal{A}}$ and a boundary variety $\Sigma_{\mathrm{bnd}} := \mathbb{R}^P \setminus \bigcup_{\alpha \in \mathcal{A}} \mathcal{C}_\alpha$.

\begin{theorem}[Generic Stratified Structure of Global Optima]
\label{thm:global_manifold}
There exists a Lebesgue-null set $\mathcal{E} \subset \mathbb{R}^q$ such that, for every target label vector $\mathbf{Y} \notin \mathcal{E}$:
\begin{enumerate}
    \item The interior fiber $\mathcal{M}_0(\mathbf{Y}) := \mathcal{M}_{\mathrm{total}}(\mathbf{Y}) \setminus \Sigma_{\mathrm{bnd}}$ is a finite disjoint union of strata $\mathcal{M}_\alpha(\mathbf{Y}) := \mathcal{M}_{\mathrm{total}}(\mathbf{Y}) \cap \mathcal{C}_\alpha$.  Every non-empty stratum $\mathcal{M}_\alpha(\mathbf{Y})$ is a smooth embedded submanifold of $\mathbb{R}^P$ of codimension $q$, hence of dimension $P - q$;
    \item The boundary fiber $\mathcal{M}_{\mathrm{bnd}}(\mathbf{Y}) := \mathcal{M}_{\mathrm{total}}(\mathbf{Y}) \cap \Sigma_{\mathrm{bnd}}$ is contained in a finite union of smooth submanifolds of dimension at most $P - q - 1$.  Consequently, its Hausdorff dimension and measure satisfy:
    \begin{align}
       \dim_{\mathrm H}\mathcal{M}_{\mathrm{bnd}}(\mathbf{Y}) \le P - q - 1,
       \qquad
       \mathcal{H}^{P-q}\!\left(\mathcal{M}_{\mathrm{bnd}}(\mathbf{Y})\right) = 0.
    \end{align}
\end{enumerate}
\end{theorem}

Theorem~\ref{thm:global_manifold} proves that the top-dimensional component of any generic zero-loss fiber is an interior $(P-q)$-dimensional smooth manifold, while non-smooth boundary fibers are negligible.  To ensure non-emptiness ($\mathcal{M}_0(\mathbf{Y}) \neq \varnothing$), we establish a strict-interior interpolation guarantee under mild overparameterization:

\begin{proposition}[Strict-Interior Exact Interpolation]
\label{prop:strict_interpolation}
Let training inputs $\mathbf{x}^{(1)}, \dots, \mathbf{x}^{(N)}$ be pairwise distinct.  Suppose hidden activation $\sigma:\mathbb{R}\to\mathbb{R}$ is continuous, piecewise polynomial with finite breakpoint set $\mathcal{B}_\sigma$, and does not agree almost everywhere with any single polynomial.  If the penultimate width satisfies $m_{L-1} \ge N - 1$ and all preceding hidden widths satisfy $m_\ell \ge 1$ ($1 \le \ell \le L-2$), then for every label matrix $\mathbf{Y} \in \mathbb{R}^{N \times m}$, there exists $\boldsymbol\theta \in \mathbb{R}^P$ such that
\begin{align}
 \Phi(\boldsymbol\theta) = \mathbf{Y},
 \qquad
 z_j^{(s,\ell)}(\boldsymbol\theta) \notin \mathcal{B}_\sigma
 \quad
 (\forall s \in [N], \, \ell \in [L-1], \, j \in [m_\ell]).
\end{align}
Consequently, $\mathcal{M}_0(\mathbf{Y}) \neq \varnothing$ (proof in Supplementary Section~\ref{app:strict_interpolation}).
\end{proposition}

While finite-sample interpolation by nonpolynomial activations builds on the density framework of Leshno et al.~\cite{leshno1993multilayer} and Constantinescu and Popescu \cite{constantinescu2024approximation}, Proposition~\ref{prop:strict_interpolation} provides the \emph{strict-interior} construction required by Theorem~\ref{thm:global_manifold}: the affine output bias reduces the penultimate width requirement to $N-1$, preceding layers require only width one, and all pre-activations strictly avoid $\mathcal{B}_\sigma$.

\subsubsection{Global exact convexification via copositive lifting}
\label{subsubsec:global-convexification}

Theorem~\ref{thm:global_manifold} shows that generic zero-loss fibers are locally smooth within each activation stratum, but this local regularity does not simplify the global problem. Stratum closures meet nonlinearly along activation boundaries, and the explicit ReLU construction in Supplementary Section~\ref{app:proof} shows that distinct global-optimum components can be separated by a hypersurface containing no zero-loss solution. Thus neither overparameterization nor stratum-wise smoothness alone provides a connected route between global solutions. This topological obstruction motivates a global formulation that does not depend on local basin traversal.

To bypass non-convex local search, we construct an exact convex conic reformulation.  The FIP, piecewise-polynomial reduction, and McCormick linearizations established in Section~\ref{sec:multilevel-binary-codebook} cast the empirical training problem into the bounded mixed-binary quadratic program:
\begin{align}
 \min_{\mathbf u\in\mathbb R_+^p}\quad
 &\mathbf u^\top\mathbf Q\mathbf u+2\mathbf c^\top\mathbf u
 \label{eq:qcbo-canonical}\\
 \mathrm{s.t.}\quad
 &\mathbf a_h^\top\mathbf u=b_h\quad(h\in[r]),
 \qquad
 u_i\in\{0,1\}\quad(i\in\mathcal B),
 \notag
\end{align}
where FIP bounds guarantee that the affine feasible domain $\mathcal F_{\mathrm{QCBO}} \subset \mathbb R_+^p$ is compact and non-empty.  Piecewise-polynomial monomials of degree $d \ge 3$ are converted to bilinear terms $y_j = y_{j-1}z$ on FIP-certified intervals, yielding the canonical quadratic form without heuristic relaxation.

\paragraph{Completely positive matrix lifting.}
For discrete quantized parameters, we lift the decision vector $\mathbf u \in \mathbb R_+^p$ into the symmetric matrix variable $\mathbf X \in \mathbb R^{(p+1)\times(p+1)}$:
\begin{align}
 \mathbf X =
 \begin{bmatrix}\boldsymbol\Lambda&\mathbf u\\ \mathbf u^\top&1\end{bmatrix}
 \in\mathcal C_{p+1}^*,
 \qquad
 \mathcal C_{p+1}^* := \operatorname{cone}\left\{\mathbf v\mathbf v^\top : \mathbf v\in\mathbb R_+^{p+1}\right\},
 \label{eq:cp-cone-def}
\end{align}
where $\mathcal C_{p+1}^*$ is the completely positive cone.  Enforcing the paired squared equalities $\mathbf a_h^\top\boldsymbol\Lambda\mathbf a_h = b_h^2$ ($h\in[r]$) forces each rank-one atom in a completely positive decomposition to satisfy the affine constraints, while the diagonal conditions $\Lambda_{ii} = u_i$ ($i\in\mathcal B$) enforce exact Booleanity.  Defining the lifted cost matrix $\widehat{\mathbf Q} := [\mathbf Q, \mathbf c; \mathbf c^\top, 0]$ yields the completely positive program:
\begin{align}
 \mathcal P_{\mathrm{CPP}}:\quad
 \min_{\mathbf X\in\mathcal C_{p+1}^*}\quad
 \langle\widehat{\mathbf Q},\mathbf X\rangle
 \quad
 \mathrm{s.t.}\quad
 \mathcal A(\mathbf X)=\mathbf b.
 \label{eq:cpp-formulation-results}
\end{align}

\paragraph{Generalized copositive lifting for continuous architectures.}
For continuous parameters $\mathbf W^{(\ell)} \in \mathbb R^{m_\ell \times m_{\ell-1}}$ and activations residing in a compact polyhedral domain $\mathcal K \subset \mathbb R^p$, the network dynamics form a bounded quadratically constrained quadratic program (QCQP).  Under the generalized copositive framework of Burer and Dong~\cite{burer2012representing}, this problem embeds over the domain cone $\mathcal K_+ := \{\lambda [\mathbf y; 1] : \mathbf y \in \mathcal K, \lambda \ge 0\}$ into the generalized completely positive cone:
\begin{align}
 \mathcal C^*(\mathcal K) := \operatorname{cl}\operatorname{cone}\left\{
 \begin{bmatrix}\mathbf y\\1\end{bmatrix}
 \begin{bmatrix}\mathbf y\\1\end{bmatrix}^{\!\top} :
 \mathbf y\in\mathcal K
 \right\},
 \label{eq:generalized-copositive-cone}
\end{align}
yielding the convex conic program $\mathcal P_{\mathrm{GCPP}}: \min_{\mathbf X\in\mathcal C^*(\mathcal K)} \{\langle\widehat{\mathbf Q},\mathbf X\rangle : \mathcal A(\mathbf X)=\mathbf b\}$.

\begin{theorem}[Unified Exact Conic Convexification of Neural Network Training]
\label{thm:exact_cpp_equivalence}
Let training data $\{(\mathbf{x}^{(s)}, \mathbf{y}^{(s)})\}_{s=1}^N$ be given, and let hidden activations $\sigma$ be continuous and piecewise polynomial with FIP-certified compact pre-activation bounds.  Then:
\begin{enumerate}
    \item \emph{Discrete-Quantized Networks (Completely Positive Lift)}: On the compact feasible set $\mathcal F_{\mathrm{QCBO}} \neq \varnothing$:
    \begin{enumerate}
        \item \emph{Global Value Equivalence}: $p^*_{\mathrm{QCBO}} = p^*_{\mathrm{CPP}}$;
        \item \emph{Optimal-Face Preservation}: $\pi_{\mathbf u}\!\left(\operatorname{Argmin}(\mathcal P_{\mathrm{CPP}})\right) = \operatorname{conv}\!\left(\operatorname{Argmin}(\mathcal P_{\mathrm{QCBO}})\right)$;
        \item \emph{Rank-One Extreme Points}: Every extreme point of $\operatorname{Argmin}(\mathcal P_{\mathrm{CPP}})$ is of the form $[\mathbf u^\star; 1][\mathbf u^\star; 1]^\top$ for a globally optimal discrete network parameter $\mathbf u^\star \in \operatorname{Argmin}(\mathcal P_{\mathrm{QCBO}})$.
    \end{enumerate}
    \item \emph{Continuous-Weight Networks (Generalized Copositive Lift)}: For continuous parameters on a compact domain $\mathcal K \subset \mathbb R^p$ with non-empty feasible set:
    \begin{enumerate}
        \item \emph{Global Value Equivalence}: $p^*_{\mathrm{QCQP}} = p^*_{\mathrm{GCPP}}$;
        \item \emph{Convex Hull Characterization}: $\operatorname{Argmin}(\mathcal P_{\mathrm{GCPP}}) = \operatorname{conv}\!\left(\left\{[\mathbf y;1][\mathbf y;1]^\top : \mathbf y\in\operatorname{Argmin}(\mathcal P_{\mathrm{QCQP}})\right\}\right)$.
    \end{enumerate}
\end{enumerate}
\end{theorem}

The significance of Theorem~\ref{thm:exact_cpp_equivalence} is representational rather than a claim that neural-network training becomes polynomial-time solvable. Exact completely positive cone membership remains computationally intractable in general; the theorem relocates, rather than removes, the hardness. This distinction also clarifies the relation to existing convex training formulations. Sign-pattern methods derive exact finite convex programs primarily for shallow, positively homogeneous ReLU networks and scale with the number of data-dependent activation patterns~\cite{pilanci2020neural,ergen2021revealing,sahiner2021vector}. Standard SDP lifts cover broader quadratic models but generally relax the completely positive or rank-one structure and can therefore introduce an optimality gap~\cite{bartan2021training,prakhya2025convex}. By contrast, our QCBO-to-conic construction is exact for any prescribed finite depth and width with continuous piecewise-polynomial activations and FIP-certified compact bounds. More general continuous activations are first approximated on those compact ranges, after which exactness holds for the resulting finite spline model. The practical value of the convex formulation is that global optimality, affine architectural consistency, and combinatorial oracle error become separate mathematical objects, enabling certified bounds and a modular Ising--classical solver.

\subsection{Noise-robust hybrid dynamics}
\label{sec:results-noise-robust}

The exact completely positive reformulation reduces network training to linear conic optimization over the compact Boolean moment polytope $\mathcal D := \operatorname{conv}\{[\mathbf w; 1][\mathbf w; 1]^\top : \mathbf w \in \{0,1\}^p\}$.  The Quantum Conditional Gradient Descent (QCGD) framework \cite{yurtsever2022q} resolves the classical-quantum division of labor: deterministic affine constraints $\mathcal A(\mathbf V) = \mathbf b$ and dual multipliers $\mathbf z_t$ are updated classically via augmented Lagrangian ascent, while the non-convex linear minimization oracle (LMO) over $\mathcal D$ is delegated to an unconstrained $p$-bit Ising Hamiltonian:
\begin{align}
 \min_{\mathbf D \in \mathcal D} \langle \mathbf G_t, \mathbf D \rangle
 \;\Longleftrightarrow\;
 \min_{\mathbf w \in \{0,1\}^p}
 \begin{bmatrix}\mathbf w\\1\end{bmatrix}^{\!\top}
 \mathbf G_t
 \begin{bmatrix}\mathbf w\\1\end{bmatrix},
 \qquad
 \mathbf G_t := \widehat{\mathbf Q} + \mathcal A^*(\mathbf z_t + \alpha_t \mathbf r_t),
 \label{eq:lmo-qubo-isomorphism}
\end{align}
where $\mathbf r_t := \mathcal A(\mathbf V_t) - \mathbf b$ is the primal constraint residual.  The complete update order and parameter schedules are given in Supplementary Section~\ref{subsec:supp-standard-qcgd-algo}.

Physical Ising annealers and coherent Ising machines are stochastic and have
finite coefficient precision, so their returned atoms need not solve the QUBO
exactly.  Proposition~\ref{convergence_theorem} shows that QCGD retains its
$\mathcal O(T^{-1/2})$ expected objective and feasibility rates under a
filtration-adapted additive-error oracle and converges almost surely to the
primal solution set.  This robustness requires only logarithmically many
independent hardware repetitions per iteration and logarithmic coefficient
precision, for a cumulative oracle complexity of $\mathcal O(T\log T)$.  The
oracle model, finite-time and almost-sure bounds, amplification argument, and
DAC precision schedule are given in Supplementary
Section~\ref{theoremproof}.

\subsection{Parallel sample-wise QPH--QCGD}
\label{sec:results-samplewise-qph}
\label{sec:samplewise-qph-qcgd}

To scale global conic optimization to large datasets under physical hardware qubit limitations, we introduce \emph{QPH--QCGD}, a two-level decomposition algorithm for the Decomposed Lower-Bound Optimization (DLBO) framework.  Its outer level adapts the scenario-decomposition principle of progressive hedging (PH)~\cite{rockafellar1991scenarios}: identifying each training sample $s \in [N]$ with an independent realization endowed with empirical probability $p_s = 1/N$, it maintains local copies of the network parameters and enforces cross-sample non-anticipativity via probability-weighted augmented Lagrangian multipliers.  Its inner level solves each sample subproblem in parallel using QCGD, delegating only small, unconstrained $d_s$-bit QUBO instances to the physical Ising solver.

\subsubsection{Local QCBO sets and exact completely positive lifts}
Let $\mathbf{x} \in \{0,1\}^n$ denote the shared parameter code bits (with $n = BP$), and let $\mathbf{y}_s \in \{0,1\}^{m_s}$ collect the sample-specific activation selectors, interpolation coordinates, and McCormick linearization variables for sample $s \in [N]$.  After encoding slacks and bounds into binary forms, the discrete feasible configuration set for sample $s$ is:
\begin{align}
 \mathcal{X}_s := \left\{ \mathbf{v}_s = \begin{bmatrix} \mathbf{x}_s \\ \mathbf{y}_s \end{bmatrix} \in \{0,1\}^{d_s} : \mathbf{H}_s \mathbf{v}_s = \mathbf{r}_s \right\},
 \qquad d_s = n + m_s.
 \label{eq:qph-local-assignment-set}
\end{align}
For the lifted variable $\mathbf{w}_s(\mathbf{v}) = [1; \mathbf{v}]$, we distinguish the unconstrained Boolean atomic domain $\mathcal{D}_s$ from its affine feasible slice $\mathcal{K}_s$:
\begin{align}
 \mathcal{D}_s &:= \operatorname{conv}\left\{ \mathbf{w}_s(\mathbf{v})\mathbf{w}_s(\mathbf{v})^\top : \mathbf{v} \in \{0,1\}^{d_s} \right\},
 \label{eq:qph-atomic-domain}\\
 \mathcal{K}_s &:= \left\{ \mathbf{P} \in \mathcal{D}_s : \mathcal{A}_s(\mathbf{P}) = \mathbf{b}_s \right\} = \operatorname{conv}\left\{ \mathbf{w}_s(\mathbf{v})\mathbf{w}_s(\mathbf{v})^\top : \mathbf{v} \in \mathcal{X}_s \right\},
 \label{eq:qph-local-hull}
\end{align}
where $\mathcal{A}_s(\mathbf{P}) = \mathbf{b}_s$ enforces both linear equalities $\mathbf{H}_s \mathbf{v} = \mathbf{r}_s$ and paired squared equalities $\mathbf{h}_{s\ell}^\top \mathbf{V} \mathbf{h}_{s\ell} = r_{s\ell}^2$ ($\ell \in [q_s]$):
\begin{align}
 \mathcal{A}_s(\mathbf{P}) = \mathbf{b}_s \;\Longleftrightarrow\;
 \begin{cases}
  \mathbf{H}_s \mathbf{v} = \mathbf{r}_s, \\
  \mathbf{h}_{s\ell}^\top \mathbf{V} \mathbf{h}_{s\ell} = r_{s\ell}^2, & \ell = 1, \dots, q_s.
 \end{cases}
 \label{eq:qph-affine-map}
\end{align}
For any atomic mixture $\mathbf{P} \in \mathcal{D}_s$, the paired constraints enforce $\operatorname{Var}(\mathbf{h}_{s\ell}^\top \mathbf{v}) = 0$, guaranteeing that every atom in the completely positive decomposition strictly satisfies $\mathcal{X}_s$~\cite{burer2009copositive}.

\subsubsection{Consensus variables, exactness boundary, and the DLBO moment hierarchy}
\label{sec:dlbo-moment-hierarchy}
Let $\mathcal{R}_s$ be a linear operator extracting consensus statistics of the shared parameters from the local lifted matrix $\mathbf{P}_s$, and let $\boldsymbol{\zeta}$ denote the global consensus variable.  The second-order DLBO shares all first- and second-order separator moments:
\begin{align}
 \mathcal{R}_s(\mathbf{P}_s) := \mathcal{R}_s^{(2)}(\mathbf{P}_s) := \bigl(\mathbf{x}_s, \operatorname{svec}(\mathbf{X}_s)\bigr),
 \label{eq:qph-second-order-consensus-map}
\end{align}
yielding the decomposed convex consensus program:
\begin{align}
 \min_{\{\mathbf{P}_s\}, \boldsymbol{\zeta}}\quad
 &\sum_{s=1}^N p_s \langle\mathbf{C}_s, \mathbf{P}_s\rangle
 \label{eq:qph-convex-consensus}\\
 \mathrm{s.t.}\quad
 &\mathbf{P}_s \in \mathcal{K}_s, \qquad \mathcal{R}_s(\mathbf{P}_s) = \boldsymbol{\zeta} \quad (s = 1, \dots, N).
 \notag
\end{align}
While second-order consensus \eqref{eq:qph-second-order-consensus-map} is computationally scalable, full equivalence to the monolithic QCBO requires sharing the complete separator probability mass function $\boldsymbol{\mu}_s = (\mu_s(\mathbf{a}))_{\mathbf{a} \in \{0,1\}^n}$ over the extended domain:
\begin{align}
 \widehat{\mathcal{D}}_s := \operatorname{conv}\left\{ \left(\mathbf{w}_s(\mathbf{v})\mathbf{w}_s(\mathbf{v})^\top, \mathbf{e}_{\mathbf{x}_s}\right) : \mathbf{v} \in \{0,1\}^{d_s} \right\},
 \qquad
 \widehat{\mathcal{K}}_s := \left\{ (\mathbf{P}, \boldsymbol{\mu}) \in \widehat{\mathcal{D}}_s : \mathcal{A}_s(\mathbf{P}) = \mathbf{b}_s \right\},
 \label{eq:qph-extended-local-hull}
\end{align}
under which consensus $\boldsymbol{\mu}_s = \boldsymbol{\mu}$ ($\forall s$) guarantees exact factorization via conditional independence \cite{bomze2022two,gabl2023sparse}:
\begin{align}
 \boldsymbol{\mu}_s = \big(\mu_s(\mathbf{a})\big)_{\mathbf{a}\in\{0,1\}^n}, \qquad \mu_s(\mathbf{a}) = \Pr(\mathbf{x}_s = \mathbf{a}), \qquad \boldsymbol{\mu}_s = \boldsymbol{\mu} \quad (s=1,\dots,N).
 \label{eq:qph-full-marginal-consensus}
\end{align}
For $r \in [n]$, let $\mathfrak S_r$ index the nonconstant Boolean monomials of degree at most $r$, and set $q_r:=|\mathfrak S_r|=\sum_{\ell=1}^r\binom{n}{\ell}$:
\begin{align}
 \mathfrak S_r:=\{S\subseteq[n]:1\le |S|\le r\},\qquad
 \boldsymbol\psi_r(\mathbf x)
 :=\bigl(\chi_S(\mathbf x)\bigr)_{S\in\mathfrak S_r},
 \qquad \chi_S(\mathbf x):=\prod_{j\in S}x_j,
 \label{eq:dlbo-order-r-feature-map}
\end{align}
We augment both the unconstrained atomic domain and its feasible slice:
\begin{align}
 \mathcal D_s^{(r)}
 &:= \operatorname{conv}\left\{
 \left(\mathbf w_s(\mathbf v)\mathbf w_s(\mathbf v)^\top,
 \boldsymbol\psi_r(\mathbf x)\right):
 \mathbf v=(\mathbf x,\mathbf y_s)\in\{0,1\}^{d_s}\right\},
 \label{eq:dlbo-order-r-atomic-domain}\\
 \mathcal K_s^{(r)}
 &:=\left\{(\mathbf P,\mathbf m)\in\mathcal D_s^{(r)}:
 \mathcal A_s(\mathbf P)=\mathbf b_s\right\}
 = \operatorname{conv}\left\{ \left(\mathbf w_s(\mathbf v)\mathbf w_s(\mathbf v)^\top,
 \boldsymbol\psi_r(\mathbf x)\right) : \mathbf v \in \mathcal X_s \right\},
 \label{eq:dlbo-order-r-local-hull}
\end{align}
and define the order-$r$ DLBO value by
\begin{align}
 p_r^\star := \min_{\{(\mathbf P_s, \mathbf m_s^{(r)})\}, \boldsymbol\zeta^{(r)}} \left\{ \sum_{s=1}^N p_s \langle\mathbf C_s, \mathbf P_s\rangle : (\mathbf P_s, \mathbf m_s^{(r)}) \in \mathcal K_s^{(r)}, \; \mathbf m_s^{(r)} = \boldsymbol\zeta^{(r)} \; (s\in[N]) \right\}.
 \label{eq:dlbo-order-r-program}
\end{align}

\begin{fact}[DLBO Hierarchy and Dominance over Standard SDP \cite{bomze2022two,gabl2023sparse}]
\label{fact:dlbo-advantage}
Compared with classical monolithic semidefinite programming (SDP) relaxations (e.g., standard Shor or Shor--McCormick SDP lifts), the Decomposed Lower-Bound Optimization (DLBO) moment hierarchy provides two fundamental theoretical and structural advantages:
\begin{enumerate}
 \item \emph{Bound tightness and asymptotic exactness}: The sequence of order-$r$ DLBO lower bounds forms a monotone tightening hierarchy that strictly dominates the monolithic standard SDP bound:
 \begin{align}
  p_{\mathrm{mono\text{-}SDP}}^\star \;\le\; p_2^\star \;\le\; p_3^\star \;\le\; \cdots \;\le\; p_n^\star \;=\; p_{\mathrm{QCBO}}^\star.
  \label{eq:dlbo-lower-bound-hierarchy}
 \end{align}
 At the base order $r=2$, DLBO already satisfies $p_2^\star \ge p_{\mathrm{mono\text{-}SDP}}^\star$ because local completely positive hulls $\mathcal K_s^{(2)}$ rigorously enforce Boolean integrality on local sample variables.  At full order $r = n$, M\"obius inversion on Boolean marginal moments:
 \begin{align}
  \mu_s(\mathbf a) = \sum_{T \subseteq [n] \setminus A(\mathbf a)} (-1)^{|T|} m_{s, A(\mathbf a) \cup T}, \qquad A(\mathbf a) := \{j : a_j = 1\},
  \label{eq:dlbo-mobius-marginal}
 \end{align}
 guarantees exact equivalence to full discrete distribution consensus, closing the relaxation gap ($p_n^\star = p_{\mathrm{QCBO}}^\star$).
 
 \item \emph{Matrix dimensional scaling and memory reduction}: Monolithic SDP lifts all variables across all $N$ training samples into a single global matrix of dimension $(N m_s + n + 1) \times (N m_s + n + 1)$, requiring $\mathcal O(N^2 m_s^2)$ storage.  In contrast, DLBO decomposes monolithic training into $N$ parallel sample-wise blocks of compact dimension $(d_s + 1) \times (d_s + 1)$ with $d_s = n + m_s \ll N m_s$, reducing the per-subproblem matrix storage from dataset scale $\mathcal O(N^2)$ to single-sample scale $\mathcal O(1)$ with respect to $N$.
\end{enumerate}
\end{fact}

The moment order also determines the interaction locality of the Boolean oracle.

\begin{proposition}[Moment order controls Ising locality]
\label{prop:dlbo-ising-locality}
Fix $r\in\{2,\ldots,n\}$ and a sample $s$.  For any
$\mathbf G\in\mathbb S^{d_s+1}$ and $\mathbf h\in\mathbb R^{q_r}$, write
\begin{align}
 \mathbf G=
 \begin{bmatrix}
  g_{00} & \mathbf g_{0v}^\top\\
  \mathbf g_{0v} & \mathbf G_{vv}
 \end{bmatrix},
 \qquad
 \mathbf Q(\mathbf G):=\mathbf G_{vv}+2\operatorname{Diag}(\mathbf g_{0v}).
 \label{eq:dlbo-block-gradient}
\end{align}
Then the order-$r$ linear minimization oracle (LMO) satisfies
\begin{align}
 &\min_{(\mathbf S,\boldsymbol\eta)\in\mathcal D_s^{(r)}}
 \bigl\{\langle\mathbf G,\mathbf S\rangle
       +\langle\mathbf h,\boldsymbol\eta\rangle\bigr\}
 \notag\\
 &\qquad =g_{00}+\min_{\mathbf v=(\mathbf x,\mathbf y_s)\in\{0,1\}^{d_s}}
 \left\{\mathbf v^\top\mathbf Q(\mathbf G)\mathbf v
 +\sum_{T\in\mathfrak S_r}h_T\prod_{j\in T}x_j\right\}.
 \label{eq:dlbo-order-r-lmo}
\end{align}
Consequently, the oracle has polynomial degree at most $r$.  At $r=2$, it is
an exact $d_s$-variable QUBO, or equivalently a two-local logical Ising
Hamiltonian, without auxiliary variables.  For $r>2$, higher-body terms occur
in general and require either native higher-order interactions or
quadratization before execution on two-local hardware.
\end{proposition}

For the Euclidean QPH penalty, the moment-gradient coefficients at outer
iteration $k$ and inner iteration $t$ are
\begin{align}
 h_{s,T}^{k,t}
 =\omega_{s,T}^k
 +\rho\bigl(m_{s,T}^{(r),k,t}-\bar\zeta_T^{(r),k}\bigr),
 \qquad T\in\mathfrak S_r.
 \label{eq:dlbo-order-r-moment-gradient}
\end{align}
Thus singleton, pair, and higher-order consensus multipliers become,
respectively, local fields, pairwise couplers, and higher-body interactions.
Proposition~\ref{prop:dlbo-ising-locality} identifies $r=2$ as the highest
hierarchy order for which every QCGD oracle is uniformly guaranteed to remain
an auxiliary-free QUBO.  It simultaneously satisfies
$p_2^\star\ge p_{\mathrm{mono\text{-}SDP}}^\star$ and limits each sample oracle
to $d_s=n+m_s$ logical variables, compared with the monolithic unique-variable
count $d_{\mathrm{mono}}=n+\sum_{s=1}^N m_s$; physical embedding overhead is
hardware-topology dependent.  Higher orders trade locality and
$q_r$ consensus coordinates for tighter bounds, culminating in exactness at
$r=n$, where $q_n=2^n-1$.  Supplementary
Section~\ref{sec:supp-dlbo-ising-locality} gives the proof and explicit
two-local reductions.

\subsubsection{Two-level progressive hedging and inner parallel QCGD dynamics}
At outer iteration $k$, with hedging multipliers $\boldsymbol{\omega}_s^k$ (satisfying $\sum_{s=1}^N p_s \boldsymbol{\omega}_s^k = \mathbf{0}$) and consensus reference $\bar{\boldsymbol{\zeta}}^k = \sum_{s=1}^N p_s \mathcal{R}_s(\mathbf{P}_s^k)$, QPH minimizes the augmented Lagrangian objective:
\begin{align}
 \mathcal{L}_{\rho}^{\rm PH} = \sum_{s=1}^N p_s \left[ \langle\mathbf{C}_s, \mathbf{P}_s\rangle + \langle\boldsymbol{\omega}_s^k, \mathcal{R}_s(\mathbf{P}_s) - \boldsymbol{\zeta}\rangle + \frac{\rho}{2}\|\mathcal{R}_s(\mathbf{P}_s) - \boldsymbol{\zeta}\|_2^2 \right].
 \label{eq:qph-augmented-objective}
\end{align}
This decouples into $N$ fully parallel sample subproblems:
\begin{align}
 \mathbf{P}_s^{k+1} \approx \argmin_{\mathbf{P} \in \mathcal{K}_s} \Phi_{s,k}(\mathbf{P}),
 \qquad
 \Phi_{s,k}(\mathbf{P}) := \langle\mathbf{C}_s, \mathbf{P}\rangle + \langle\boldsymbol{\omega}_s^k, \mathcal{R}_s(\mathbf{P})\rangle + \frac{\rho}{2}\|\mathcal{R}_s(\mathbf{P}) - \bar{\boldsymbol{\zeta}}^k\|_2^2.
 \label{eq:qph-sample-subproblem}
\end{align}
Rather than invoking a constrained solver for $\mathcal{K}_s$, each subproblem is solved via an inner QCGD loop over the unconstrained hypercube domain $\mathcal{D}_s$.  At inner step $t$, with local residual $\mathbf{a}_s^{k,t} := \mathcal{A}_s(\mathbf{P}_s^{k,t}) - \mathbf{b}_s$ and inner multiplier $\mathbf{u}_s^{k,t}$, the augmented proxy is:
\begin{align}
 \mathbf{a}_s^{k,t} &:= \mathcal{A}_s(\mathbf{P}_s^{k,t}) - \mathbf{b}_s,
 \label{eq:qph-local-affine-residual}\\
 \mathcal{Q}_{s,k,t}(\mathbf{P}; \mathbf{u}_s^{k,t}) &:= \Phi_{s,k}(\mathbf{P}) + \langle\mathbf{u}_s^{k,t}, \mathcal{A}_s(\mathbf{P}) - \mathbf{b}_s\rangle + \frac{\beta_{s,k,t}}{2}\|\mathcal{A}_s(\mathbf{P}) - \mathbf{b}_s\|_2^2,
 \label{eq:qph-inner-augmented-objective}
\end{align}
yielding the inner gradient and Boolean linear minimization oracle (LMO):
\begin{align}
 \mathbf{G}_s^{k,t} &:= \nabla_{\mathbf{P}}\mathcal{Q}_{s,k,t}(\mathbf{P}_s^{k,t}; \mathbf{u}_s^{k,t})
 = \mathbf{C}_s + \mathcal{R}_s^*(\boldsymbol{\omega}_s^k) + \rho \mathcal{R}_s^*(\mathcal{R}_s(\mathbf{P}_s^{k,t}) - \bar{\boldsymbol{\zeta}}^k) + \mathcal{A}_s^*(\mathbf{u}_s^{k,t} + \beta_{s,k,t}\mathbf{a}_s^{k,t}),
 \label{eq:qph-local-gradient}\\
 \mathbf{S}_s^{k,t} &\in \argmin_{\mathbf{S} \in \mathcal{D}_s} \langle\mathbf{G}_s^{k,t}, \mathbf{S}\rangle,
 \label{eq:qph-lmo}\\
 \mathbf{P}_s^{k,t+1} &:= (1-\gamma_{s,k,t})\mathbf{P}_s^{k,t} + \gamma_{s,k,t}\mathbf{S}_s^{k,t},
 \label{eq:qph-qcgd-matrix-update}\\
 \mathbf{u}_s^{k,t+1} &:= \mathbf{u}_s^{k,t} + \eta_{s,k,t}\mathbf{a}_s^{k,t+1}.
 \label{eq:qph-inner-dual-update}
\end{align}
The gradient $\mathbf{G}_s^{k,t}$ is Lipschitz on $\mathcal{D}_s$ with:
\begin{align}
 L_{s,k,t} \le \rho\|\mathcal{R}_s\|_{\mathrm{op}}^2 + \beta_{s,k,t}\|\mathcal{A}_s\|_{\mathrm{op}}^2.
 \label{eq:qph-inner-smoothness}
\end{align}
The inner Frank--Wolfe gap $g_s^{k,t} := \langle\mathbf{G}_s^{k,t}, \mathbf{P}_s^{k,t} - \mathbf{S}_s^{k,t}\rangle$:
\begin{align}
 g_s^{k,t} := \langle\mathbf{G}_s^{k,t}, \mathbf{P}_s^{k,t} - \mathbf{S}_s^{k,t}\rangle
 \label{eq:qph-fw-gap}
\end{align}
monitors proxy suboptimality.  Crucially, because $\mathcal{D}_s$ is the Boolean atomic hull, the LMO \eqref{eq:qph-lmo} is \emph{strictly isomorphic to an unconstrained $d_s$-bit QUBO}:
\begin{align}
 \mathbf{v}_s^{k,t} \in \argmin_{\mathbf{v} \in \{0,1\}^{d_s}} \left\{ \mathbf{v}^\top \mathbf{G}_{vv,s}^{k,t} \mathbf{v} + 2 (\mathbf{g}_{0v,s}^{k,t})^\top \mathbf{v} \right\},
 \qquad
 \mathbf{S}_s^{k,t} = \mathbf{w}_s(\mathbf{v}_s^{k,t})\mathbf{w}_s(\mathbf{v}_s^{k,t})^\top,
 \label{eq:qph-sample-qubo}
\end{align}
where $\mathbf{G}_{vv,s}^{k,t}$ and $\mathbf{g}_{0v,s}^{k,t}$ are subblocks of $\mathbf{G}_s^{k,t}$.  Inexact physical LMO queries with additive error $\delta_{s,k,t}$:
\begin{align}
 \langle\mathbf{G}_s^{k,t}, \mathbf{S}_s^{k,t}\rangle \le \min_{\mathbf{S} \in \mathcal{D}_s} \langle\mathbf{G}_s^{k,t}, \mathbf{S}\rangle + \delta_{s,k,t}
 \label{eq:qph-inexact-lmo}
\end{align}
are absorbed by the convergence analysis under diminishing error schedules.

\paragraph{Consensus projection and hedging updates.}
After inner subsolvers return, QPH updates the outer consensus and multipliers:
\begin{align}
 \bar{\boldsymbol{\zeta}}^{k+1} &:= \sum_{s=1}^N p_s \mathcal{R}_s(\mathbf{P}_s^{k+1}),
 \label{eq:qph-consensus-average}\\
 \boldsymbol{\omega}_s^{k+1} &:= \boldsymbol{\omega}_s^k + \rho\left(\mathcal{R}_s(\mathbf{P}_s^{k+1}) - \bar{\boldsymbol{\zeta}}^{k+1}\right),
 \label{eq:qph-hedging-update}
\end{align}
tracked by the non-anticipativity residual and primal step displacement:
\begin{align}
 r_{\rm NA}^{k+1} := \left(\sum_{s=1}^N p_s \|\mathcal{R}_s(\mathbf{P}_s^{k+1}) - \bar{\boldsymbol{\zeta}}^{k+1}\|_2^2\right)^{1/2},
 \qquad
 d_{\rm PH}^{k+1} := \rho \|\bar{\boldsymbol{\zeta}}^{k+1} - \bar{\boldsymbol{\zeta}}^k\|_2.
 \label{eq:qph-residuals}
\end{align}

\begin{theorem}[Global Convergence of Sample-Wise QPH--QCGD]
\label{thm:samplewise-qph-qcgd}
Let each local set $\mathcal{K}_s$ be a non-empty, compact convex set, and assume strong duality holds for the consensus problem \eqref{eq:qph-convex-consensus}.  Initialize $\sum_{s=1}^N p_s \boldsymbol{\omega}_s^0 = \mathbf{0}$.  Let $\widetilde{\mathbf{P}}_s^{k+1} \in \argmin_{\mathbf{P} \in \mathcal{K}_s} \Phi_{s,k}(\mathbf{P})$ be the exact sample subproblem minimizer.  If the inner QCGD solvers produce iterates $\mathbf{P}_s^{k+1} \in \mathcal{D}_s$ satisfying the summable inexactness condition:
\begin{align}
 \sum_{k=0}^\infty \|\mathbf{P}_s^{k+1} - \widetilde{\mathbf{P}}_s^{k+1}\|_F < \infty \qquad (\forall s \in [N]),
 \label{eq:qph-summable-inner-error}
\end{align}
then the non-anticipativity residual vanishes ($\lim_{k\to\infty} r_{\rm NA}^k = 0$), the objective values converge to the global optimum of \eqref{eq:qph-convex-consensus}, and every cluster point of $\{\mathbf{P}_s^k\}_{s=1}^N$ is a global primal-dual solution (proof in Supplementary Section~\ref{sec:supp-samplewise-algorithm-proofs}).
\end{theorem}

Theorem~\ref{thm:samplewise-qph-qcgd} decouples the outer scenario consensus from inner QCGD subsolvers.  Because $\Phi_{s,k}$ evolves with $k$, QCGD is executed at each outer step until meeting an inner tolerance $\epsilon_{s,k}$ satisfying $\Phi_{s,k}(\widehat{\mathbf{P}}_s^{k,t}) - \Phi_{s,k}(\widetilde{\mathbf{P}}_s^{k+1}) \le \epsilon_{s,k}$.  Under uniform $\mu_s$-strong convexity of $\Phi_{s,k}$ on $\mathcal{K}_s$, $\|\widehat{\mathbf{P}}_s^{k,t} - \widetilde{\mathbf{P}}_s^{k+1}\|_F \le \sqrt{2\epsilon_{s,k}/\mu_s}$, so that choosing $\sum_{k=0}^\infty \sqrt{\epsilon_{s,k}} < \infty$ unconditionally certifies \eqref{eq:qph-summable-inner-error}.

\subsection{Certified QUBO separation and column generation}
\label{sec:results-certified-separation}
\label{sec:samplewise-copositive-column-generation}

As an exact alternative to conditional gradient dynamics for solving the proximal subproblem \eqref{eq:qph-sample-subproblem}, we formulate a certified copositive cutting-plane and column-generation algorithm.  Building upon the hybrid copositive framework of Brown et al.~\cite{brown2024copositive}, we integrate physical Ising machines into the restricted-master localization loop as weak-separation oracles, provide explicit finite-precision error certificates, and establish an ambiguity-free three-valued oracle model that preserves global QPH convergence.

\subsubsection{Primal completely positive subproblem and its copositive dual}
At outer iteration $k$ for training sample $s \in [N]$, let $s_s := 1 + d_s$, $\mathbf{D}_{s,k} := \mathbf{C}_s + \mathcal{R}_s^*(\boldsymbol{\omega}_s^k)$, and $\mathbf{z}_k := \bar{\boldsymbol{\zeta}}^k$.  The local proximal subproblem over the completely positive cone $\mathcal{C}_{s_s}^*$ is:
\begin{align}
 p_{s,k}^* := \min_{\mathbf{P} \in \mathcal{C}_{s_s}^*}\quad
 &\langle\mathbf{D}_{s,k}, \mathbf{P}\rangle + \frac{\rho}{2}\|\mathcal{R}_s(\mathbf{P}) - \mathbf{z}_k\|_2^2
 \label{eq:cpsep-primal}\\
 \mathrm{s.t.}\quad
 &\mathcal{A}_s(\mathbf{P}) = \mathbf{b}_s.
 \notag
\end{align}
Assume the feasible slice $\mathcal{K}_s = \{\mathbf{P} \in \mathcal{C}_{s_s}^* : \mathcal{A}_s(\mathbf{P}) = \mathbf{b}_s\}$ is nonempty and compact, with trace diameter bound:
\begin{align}
 T_s := \sup\left\{ \operatorname{Tr}(\mathbf{P}) : \mathbf{P} \in \mathcal{C}_{s_s}^*, \; \mathcal{A}_s(\mathbf{P}) = \mathbf{b}_s \right\} \le s_s < \infty.
 \label{eq:cpsep-trace-bound}
\end{align}
Introducing dual multipliers $\mathbf{u}$ for $\mathcal{A}_s(\mathbf{P}) = \mathbf{b}_s$ and $\mathbf{h}$ for the consensus displacement $\mathbf{r} = \mathcal{R}_s(\mathbf{P}) - \mathbf{z}_k$, the conic Lagrangian dual is the concave maximization over the copositive cone $\mathcal{C}_{s_s}$:
\begin{align}
 d_{s,k}^* := \max_{\mathbf{u}, \mathbf{h}}\quad
 &-\langle\mathbf{u}, \mathbf{b}_s\rangle - \langle\mathbf{h}, \mathbf{z}_k\rangle - \frac{1}{2\rho}\|\mathbf{h}\|_2^2
 \label{eq:cpsep-dual}\\
 \mathrm{s.t.}\quad
 &\mathbf{Z}_{s,k}(\mathbf{u}, \mathbf{h}) := \mathbf{D}_{s,k} + \mathcal{A}_s^*(\mathbf{u}) + \mathcal{R}_s^*(\mathbf{h}) \in \mathcal{C}_{s_s}.
 \notag
\end{align}
Under strong conic duality ($p_{s,k}^* = d_{s,k}^*$), for any candidate dual pair $(\mathbf{u}, \mathbf{h})$, testing whether $\mathbf{Z}_{s,k}(\mathbf{u}, \mathbf{h}) \in \mathcal{C}_{s_s}$ is co-NP-complete.  Any nonnegative vector $\mathbf{y} \ge \mathbf{0}$ satisfying:
\begin{align}
 \mathbf{y}^\top \mathbf{Z}_{s,k}(\mathbf{u}, \mathbf{h})\mathbf{y} < 0
 \label{eq:cpsep-negative-ray}
\end{align}
is an exact certificate of dual infeasibility, generating the valid affine cut:
\begin{align}
 \langle\mathbf{D}_{s,k}, \mathbf{y}\mathbf{y}^\top\rangle + \langle\mathbf{u}, \mathcal{A}_s(\mathbf{y}\mathbf{y}^\top)\rangle + \langle\mathbf{h}, \mathcal{R}_s(\mathbf{y}\mathbf{y}^\top)\rangle \ge 0.
 \label{eq:cpsep-dual-cut}
\end{align}
Simultaneously, the rank-one dyad $\mathbf{y}\mathbf{y}^\top \in \mathcal{C}_{s_s}^*$ acts as an extreme primal ray generating a new column for the primal restricted master problem, revealing that \emph{dual weak separation and primal column generation are isomorphic operations across conic duality}.

\subsubsection{Restricted master problem and finite-precision QUBO pricing}
Let $\mathcal{Y}_j \subset [0,1]^{s_s}$ be the accumulated finite set of normalized rays at inner cutting-plane step $j$.  The algorithm proceeds as follows:
\begin{enumerate}
 \item \emph{Restricted Primal Master Solve}: Solve the master linear/quadratic program over the convex hull of active rays to primal-dual gap $\xi_j$:
 \begin{align}
  \mathbf{P}_j = \sum_{\mathbf{y} \in \mathcal{Y}_j} \lambda_{\mathbf{y}}\mathbf{y}\mathbf{y}^\top, \qquad \lambda_{\mathbf{y}} \ge 0, \qquad \mathcal{A}_s(\mathbf{P}_j) = \mathbf{b}_s,
  \label{eq:cpsep-restricted-master}
 \end{align}
 yielding primal objective $U_j$ and dual multiplier pair $(\mathbf{u}_j, \mathbf{h}_j)$ with dual value $d_j$ satisfying $0 \le U_j - d_j \le \xi_j$.
 \item \emph{QUBO Separation Pricing}: Form $\mathbf{Z}_j := \mathbf{Z}_{s,k}(\mathbf{u}_j, \mathbf{h}_j)$ and query an additive-$\delta_j$ Ising oracle on the $b_j$-bit grid encoding:
 \begin{align}
  \min_{\mathbf{z} \in \{0,1\}^{s_s b_j}} \mathbf{z}^\top \mathbf{B}_{b_j}^\top \mathbf{Z}_j \mathbf{B}_{b_j} \mathbf{z},
  \label{eq:cpsep-pricing-qubo}
 \end{align}
 where $\mathbf{B}_{b_j} = \mathbf{I}_{s_s} \otimes [2^{-1}, \dots, 2^{-b_j}]$.
 \item \emph{Ray Augmentation vs. Certificate Termination}: If the returned vector $\widetilde{\mathbf{y}}_j$ satisfies $\widetilde{\mathbf{y}}_j^\top \mathbf{Z}_j \widetilde{\mathbf{y}}_j < 0$, append $\widetilde{\mathbf{y}}_j$ to $\mathcal{Y}_j$ and iterate.  Otherwise, terminate and certify the local lower bound:
 \begin{align}
  \tau_j := \frac{s_s \|\mathbf{Z}_j\|_{\rm op}}{2^{b_j} - 1} + \delta_j,
  \qquad
  L_j := d_j - T_s \tau_j.
  \label{eq:cpsep-local-lower-bound}
 \end{align}
\end{enumerate}

\begin{proposition}[Certified Sample-Solver Accuracy]
\label{prop:cpsep-sample-certificate}
Assume the proximal subproblem \eqref{eq:cpsep-primal} is feasible, satisfies strong duality, and obeys trace bound $T_s$.  Upon termination of the pricing loop at step $j$, the true optimal subproblem value $p_{s,k}^*$ is bounded within the certified interval:
\begin{align}
 L_j \le p_{s,k}^* \le U_j, \qquad 0 \le U_j - p_{s,k}^* \le \xi_j + T_s \tau_j.
 \label{eq:cpsep-certified-gap}
\end{align}
Under fixed grid bit-depth $b_j$ and hardware error $\delta_j$, the residual tolerance band $T_s \tau_j$ constitutes a rigorous certificate of suboptimality.
\end{proposition}

\subsubsection{Three-valued oracle and cutting-plane localization}
To prevent finite-precision discretization from generating spurious feasibility declarations, we formulate an ambiguity-free \emph{three-valued oracle}:
\begin{enumerate}
 \item \emph{CUT}: If $\widetilde{\mathbf{y}}_j^\top \mathbf{Z}_j \widetilde{\mathbf{y}}_j < 0$, add the exact separating hyperplanes \eqref{eq:cpsep-dual-cut} to the localization pool;
 \item \emph{WEAK-FEASIBLE}: If minimum energy is $\ge -\delta_j$, certify the lower bound $\underline{p}_j := d_{s,k}(\boldsymbol{\theta}_j) - T_s \tau_j \le p_{s,k}^*$;
 \item \emph{REFINE}: For target level $\ell$, if $\underline{p}_j < \ell$ while no separating direction is found, dynamically increase $b_j$ and decrease $\delta_j$.
\end{enumerate}
Let $\boldsymbol{\theta} = (\mathbf{u}, \mathbf{h}) \in \mathbb{R}^{q_s}$ (with $q_s = \dim(\mathbf{u}, \mathbf{h})$) and define the exact convex level set:
\begin{align}
 \mathcal{S}_{s,k}(\ell) := \left\{ \boldsymbol{\theta} \in \Theta_{s,k}^0 : d_{s,k}(\boldsymbol{\theta}) \ge \ell, \; \mathbf{Z}_{s,k}(\boldsymbol{\theta}) \in \mathcal{C}_{s_s} \right\}.
 \label{eq:cpsep-level-set}
\end{align}
At test point $\boldsymbol{\theta}_j$, objective violations $d_{s,k}(\boldsymbol{\theta}_j) < \ell$ induce objective cuts:
\begin{align}
 \nabla d_{s,k}(\boldsymbol{\theta}_j)^\top (\boldsymbol{\theta} - \boldsymbol{\theta}_j) \ge \ell - d_{s,k}(\boldsymbol{\theta}_j) > 0,
 \qquad
 \nabla d_{s,k}(\boldsymbol{\theta}_j) = \begin{bmatrix} -\mathbf{b}_s \\ -\mathbf{z}_k - \mathbf{h}_j/\rho \end{bmatrix},
 \label{eq:cpsep-objective-cut}
\end{align}
while negative rays $\mathbf{y}_j$ induce copositivity cuts with normal $\mathbf{a}(\mathbf{y}_j) := [\mathcal{A}_s(\mathbf{y}_j\mathbf{y}_j^\top); \mathcal{R}_s(\mathbf{y}_j\mathbf{y}_j^\top)]$ and scalar $c(\mathbf{y}_j) := \langle\mathbf{D}_{s,k}, \mathbf{y}_j\mathbf{y}_j^\top\rangle$:
\begin{align}
 \mathbf{a}(\mathbf{y}_j) := \begin{bmatrix} \mathcal{A}_s(\mathbf{y}_j\mathbf{y}_j^\top) \\ \mathcal{R}_s(\mathbf{y}_j\mathbf{y}_j^\top) \end{bmatrix},
 \qquad
 c(\mathbf{y}_j) := \langle\mathbf{D}_{s,k}, \mathbf{y}_j\mathbf{y}_j^\top\rangle.
 \label{eq:cpsep-cut-normal}
\end{align}
For ellipsoid localization $\Theta_{s,k}^j = \mathcal{E}(\boldsymbol{\theta}_j, \mathbf{H}_j) := \{\boldsymbol{\theta} : (\boldsymbol{\theta} - \boldsymbol{\theta}_j)^\top \mathbf{H}_j^{-1}(\boldsymbol{\theta} - \boldsymbol{\theta}_j) \le 1\}$ with cut normal $\mathbf{g}_j$:
\begin{align}
 \Theta_{s,k}^j = \mathcal{E}(\boldsymbol{\theta}_j, \mathbf{H}_j) := \left\{ \boldsymbol{\theta} : (\boldsymbol{\theta} - \boldsymbol{\theta}_j)^\top \mathbf{H}_j^{-1}(\boldsymbol{\theta} - \boldsymbol{\theta}_j) \le 1 \right\},
 \label{eq:cpsep-ellipsoid}
\end{align}
the center and shape matrices are updated via:
\begin{align}
 \boldsymbol{\theta}_{j+1} &= \boldsymbol{\theta}_j - \frac{\mathbf{H}_j \mathbf{g}_j}{(q_s + 1)\sqrt{\mathbf{g}_j^\top \mathbf{H}_j \mathbf{g}_j}},
 \label{eq:cpsep-ellipsoid-center-update}\\
 \mathbf{H}_{j+1} &= \frac{q_s^2}{q_s^2 - 1}\left( \mathbf{H}_j - \frac{2\mathbf{H}_j \mathbf{g}_j \mathbf{g}_j^\top \mathbf{H}_j}{(q_s + 1)\mathbf{g}_j^\top \mathbf{H}_j \mathbf{g}_j} \right).
 \label{eq:cpsep-ellipsoid-shape-update}
\end{align}
On WEAK-FEASIBLE, the lower value bound is certified as:
\begin{align}
 \tau_j = \frac{s_s \|\mathbf{Z}_j\|_{\rm op}}{2^{b_j} - 1} + \delta_j,
 \qquad
 \underline{p}_j = d_{s,k}(\boldsymbol{\theta}_j) - T_s \tau_j.
 \label{eq:cpsep-ellipsoid-weak-value}
\end{align}

\begin{proposition}[Conditional Ellipsoid Oracle Complexity]
\label{prop:cpsep-ellipsoid-complexity}
Let $q_s = \dim(\mathbf{u}, \mathbf{h})$, and let the initial ellipsoid contain $\mathcal{S}_{s,k}(\ell)$ within a radius-$R_s$ ball, with non-empty level sets containing a radius-$r_s$ ball.  Suppose $\|\mathbf{Z}_{s,k}(\boldsymbol{\theta})\|_{\rm op} \le M_s$ on the localization domain.  Then achieving target accuracy $\eta > 0$ across bisection width $B_s$ requires:
\begin{align}
 \mathcal{O}\!\left(q_s^2 \log\frac{R_s}{r_s} \cdot \log\frac{B_s}{\eta}\right)
 \label{eq:cpsep-ellipsoid-oracle-complexity}
\end{align}
QUBO separation queries, with sufficient per-iteration parameter schedules:
\begin{align}
 \xi_j \le \frac{\eta}{2}, \qquad \delta_j \le \frac{\eta}{4T_s}, \qquad b_j \ge \left\lceil \log_2\left(1 + \frac{4T_s s_s M_s}{\eta}\right) \right\rceil,
 \label{eq:cpsep-ellipsoid-precision}
\end{align}
requiring $s_s b_j = \mathcal{O}(s_s \log(T_s s_s M_s / \eta))$ logical QUBO variables per query (proof in Supplementary Section~\ref{sec:supp-samplewise-algorithm-proofs}).
\end{proposition}

\subsubsection{End-to-end convergence inside the QPH framework}
For inner tolerance $\eta_{s,k} := \xi_{s,k} + T_s \tau_{s,k}$:
\begin{align}
 \eta_{s,k} := \xi_{s,k} + T_s \left( \frac{s_s \|\mathbf{Z}_{s,k}\|_{\rm op}}{2^{b_{s,k}} - 1} + \delta_{s,k} \right),
 \label{eq:cpsep-qph-local-error}
\end{align}
$\rho$-strong convexity of the proximal term in $\mathcal{R}_s(\mathbf{P})$ implies:
\begin{align}
 \|\mathcal{R}_s(\mathbf{P}_s^{k+1}) - \mathcal{R}_s(\widetilde{\mathbf{P}}_s^{k+1})\|_2 \le \sqrt{\frac{2\eta_{s,k}}{\rho}}.
 \label{eq:cpsep-proximal-statistic-error}
\end{align}

\begin{corollary}[Global QPH Convergence with QUBO Weak Separation]
\label{cor:cpsep-qph-convergence}
Under the assumptions of Proposition~\ref{prop:cpsep-sample-certificate}, suppose that each sample restricted master problem returns a locally feasible point and the precision schedule satisfies:
\begin{align}
 \sum_{k=0}^\infty \sqrt{\eta_{s,k}} < \infty \qquad (\forall s \in [N]).
 \label{eq:cpsep-qph-summability}
\end{align}
Then the non-anticipativity residual vanishes ($\lim_{k\to\infty} r_{\rm NA}^k = 0$), the objective values converge to the global optimum of the chosen convex DLBO formulation \eqref{eq:qph-convex-consensus}, and every cluster point is DLBO-optimal.
\end{corollary}

A sufficient schedule enforcing \eqref{eq:cpsep-qph-summability} sets $\eta_{s,k}^{\rm tar} = c_s(k+1)^{-2-2\kappa}$ ($\kappa > 0$) with:
\begin{align}
 \xi_{s,k} \le \frac{\eta_{s,k}^{\rm tar}}{3}, \quad T_s \delta_{s,k} \le \frac{\eta_{s,k}^{\rm tar}}{3}, \quad b_{s,k} \ge \left\lceil \log_2\left(1 + \frac{3T_s s_s M_s}{\eta_{s,k}^{\rm tar}}\right) \right\rceil, \quad \|\mathbf{Z}_{s,k}\|_{\rm op} \le M_s,
 \label{eq:cpsep-explicit-outer-schedule}
\end{align}
\section{Experimental validation}
\label{sec:experimental-validation}

To systematically assess the empirical efficacy, computational complexity, and representational robustness of our discrete optimization framework for neural network training, we conduct comprehensive experiments across multiple standard machine learning benchmarks, including Fashion-MNIST~\cite{xiao2017online}, Wine~\cite{wine_109}, and Optical Recognition of Handwritten Digits~\cite{optical_recognition_of_handwritten_digits_80}.

\subsection{Direct Ising Optimization for Binary Classification and Quantization Robustness}
\label{subsec:exp-direct-ising}

We first investigate direct discrete optimization for binary image classification on the Fashion-MNIST dataset~\cite{xiao2017online}, focusing on the standard coat versus sandal classification benchmark ($28\times 28$ grayscale images).

\begin{figure}[tbp]
    \centering
    \includegraphics[width=\textwidth]{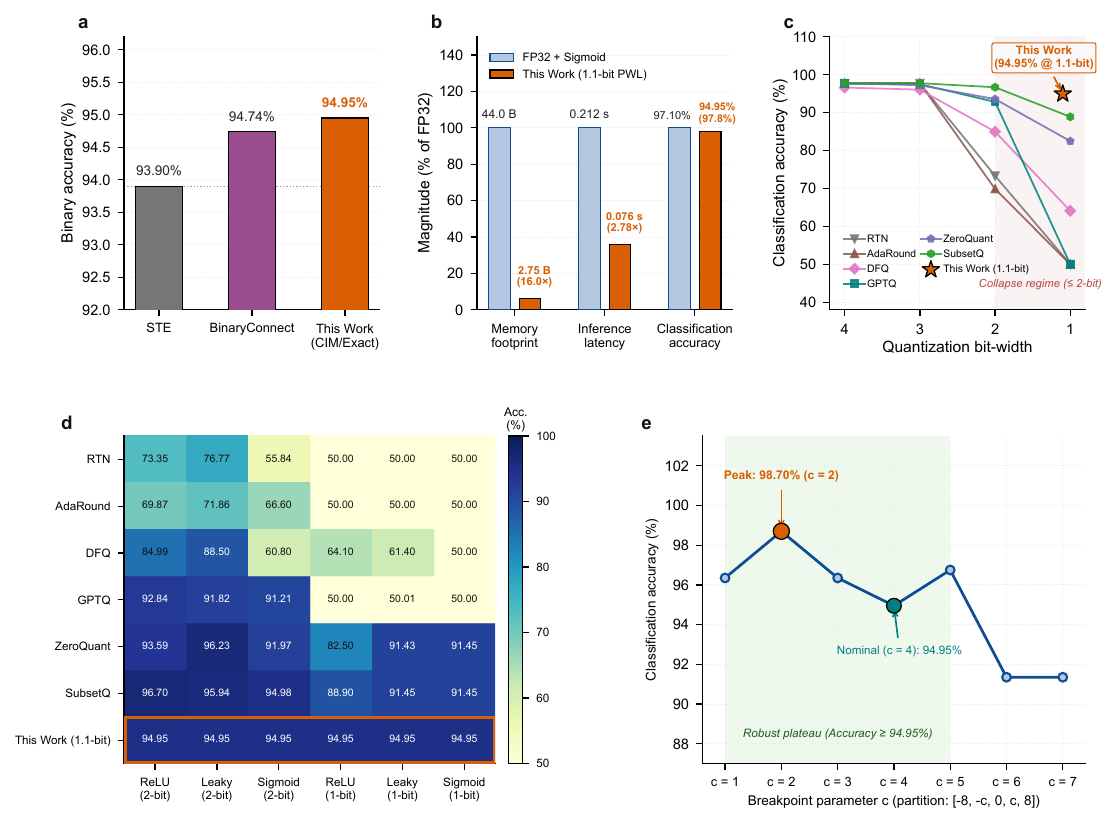}
    \caption{Direct Ising optimization benchmarks for binary classification, computational complexity trade-offs, and low-bit quantization robustness. 
    \textbf{a}, Classification accuracy comparison on Fashion-MNIST binary classification across discrete training paradigms: Straight-Through Estimator (STE, $93.90\%$)~\cite{bengio2013estimating}, BinaryConnect ($94.74\%$)~\cite{courbariaux2015binaryconnect}, and this work via direct discrete optimization ($94.95\%$). Dotted line indicates the STE baseline level.
    \textbf{b}, Multi-metric computational footprint and inference efficiency profile comparing a full-precision baseline network (FP32 continuous weights and smooth transcendental sigmoid activation) against our quantized neural network (1.1-bit discrete weights and 4-interval piecewise-linear activation). Quantization achieves a $16.0\times$ parameter memory reduction (from 44.0~B to 2.75~B) and a $2.78\times$ single-sample inference acceleration (from 0.2115~s to 0.0762~s) with a minor $2.15\%$ accuracy margin ($97.10\%$ versus $94.95\%$).
    \textbf{c}, Trajectories of classification accuracy across quantization precisions (4-bit down to 1-bit) for six post-training quantization (PTQ) baselines (RTN~\cite{kogan2025selective}, AdaRound~\cite{nagel2020up}, DFQ~\cite{nagel2019data}, GPTQ~\cite{frantargptq}, ZeroQuant~\cite{yao2022zeroquant}, and SubsetQ~\cite{oh2022non}) with ReLU activation, contrasted against our direct discrete optimization approach (starred point at 1.1-bit, $94.95\%$). Shaded area denotes the classical representational collapse regime ($\le 2$-bit).
    \textbf{d}, Cross-activation performance heatmap in the ultra-low precision regime (2-bit and 1-bit) across ReLU, LeakyReLU, and Sigmoid activation functions (complete 18-configuration numerical matrix in Supplementary Table~\ref{tab:supp_quantization_accuracy}). Framed row highlights the resilience of our model across all activation profiles.
    \textbf{e}, Sensitivity of classification accuracy to the symmetric breakpoint parameter $c \in \{1, \dots, 7\}$ under the 4-interval partition $[-8, -c, 0, c, 8]$ (Supplementary Table~\ref{tab:supp_accuracy_table}), demonstrating a broad operational plateau ($\ge 94.95\%$) across $c \in [1, 5]$ with peak accuracy at $c = 2$ ($98.70\%$).}
    \label{fig:direct_ising_benchmarks}
\end{figure}

\paragraph{Benchmark Classification Accuracy.}
Figure~\ref{fig:direct_ising_benchmarks}a presents the test classification accuracy across discrete training methodologies. By formulating the training objective directly as a binary quadratic program, our exact discrete formulation achieves a test accuracy of $94.95\%$. This performance surpasses standard discrete gradient heuristics, including the Straight-Through Estimator (STE, $93.90\%$)~\cite{bengio2013estimating} and BinaryConnect ($94.74\%$)~\cite{courbariaux2015binaryconnect}, confirming that global combinatorial formulation avoids suboptimal local traps induced by heuristic gradient approximations.

\paragraph{Memory Compression and Inference Acceleration.}
Beyond high classification accuracy, the proposed formulation yields dramatic reductions in hardware resource requirements and forward inference latency. As illustrated in Figure~\ref{fig:direct_ising_benchmarks}b, we benchmark a continuous neural network (32-bit floating-point weights, transcendental sigmoid activation) against our quantized discrete model (1.1-bit discrete weights, 4-interval piecewise-linear FIP activation). Our method compresses model parameter storage from 44.0~bytes to 2.75~bytes---a $16.0\times$ memory reduction---and decreases single-sample inference latency from 0.2115~s to 0.0762~s (a $2.78\times$ speedup), with only a minor accuracy margin of $2.15\%$ ($97.10\%$ vs. $94.95\%$).

The computational acceleration originates directly from the algebraic structure of piecewise-linear (PWL) evaluation. Evaluating a smooth transcendental activation (\emph{e.g.}, sigmoid or $\tanh$) on edge hardware requires expensive floating-point operations comprising exponentiation, division, and polynomial approximations:
\begin{align}
 T_{\mathrm{sigmoid}} = T_{\exp} + T_{\mathrm{add}} + T_{\mathrm{div}}.
\end{align}
In contrast, for a piecewise-linear activation parameterized by discrete interval selection variables $\beta_i \in \{0, 1\}$, forward evaluation reduces to a single interval indexing lookup followed by an affine scalar mapping:
\begin{align}
 T_{\mathrm{PWL}} = T_{\mathrm{lookup}} + T_{\mathrm{mul}} + T_{\mathrm{add}}.
\end{align}
Consequently, the FIP piecewise-linear formulation simultaneously facilitates an exact QCBO formulation during training and guarantees significant arithmetic acceleration during edge deployment.

\paragraph{Robustness Frontier in Ultra-Low Bit Regimes.}
To characterize resilience under aggressive weight quantization, Figure~\ref{fig:direct_ising_benchmarks}c and Supplementary Table~\ref{tab:supp_quantization_accuracy} compare our framework against six leading post-training quantization (PTQ) algorithms: RTN~\cite{kogan2025selective}, AdaRound~\cite{nagel2020up}, DFQ~\cite{nagel2019data}, GPTQ~\cite{frantargptq}, ZeroQuant~\cite{yao2022zeroquant}, and SubsetQ~\cite{oh2022non}.

While baseline PTQ techniques preserve accuracy at 4-bit and 3-bit precisions, they suffer catastrophic representational collapse when precision is reduced to $\le 2$-bit, with many baselines degrading to random guessing ($50.0\%$). Although SubsetQ~\cite{oh2022non} demonstrates resilience at 2-bit ($96.70\%$), its computational complexity scales exponentially with candidate quantization states ($\mathcal{O}(2^b)$), rendering higher precision or deeper networks prohibitive. In stark contrast, our direct global optimization operates natively in the discrete parameter domain at 1.1-bit effective precision, sustaining $94.95\%$ accuracy without relying on post-hoc rounding heuristics. As demonstrated in Figure~\ref{fig:direct_ising_benchmarks}d, this resilience is invariant across activation functions, achieving identical $94.95\%$ accuracy under ReLU, LeakyReLU, and Sigmoid profiles.

\paragraph{Piecewise-Linear Breakpoint Sensitivity.}
The resolution of activation segmentation governs the balance between discrete variable count and continuous approximation fidelity. In our nominal configuration, the pre-activation input domain $[-8, 8]$ is partitioned into four symmetric intervals using breakpoints $[-8, -c, 0, c, 8]$. Figure~\ref{fig:direct_ising_benchmarks}e and Supplementary Table~\ref{tab:supp_accuracy_table} report the classification accuracy across varying values of the scale parameter $c \in \{1, 2, \dots, 7\}$.

The empirical results demonstrate a broad and stable operational plateau across $c \in [1, 5]$, where classification accuracy consistently remains $\ge 94.95\%$, reaching a maximum of $98.70\%$ at $c = 2$. Performance degrades smoothly when $c \ge 6$ ($91.35\%$), attributable to coarsened spatial resolution within the central linear transition region around zero.

\subsection{DLBO Framework for Multi-Class Classification and Multi-Dataset Generalization}
\label{subsec:exp-dlbo-multiclass}

In the second phase of empirical validation, we evaluate the scalability, convergence dynamics, and generalization capability of the DLBO framework across multi-class classification tasks on three standard benchmark datasets: 3-class Fashion-MNIST~\cite{xiao2017online}, 3-class Wine~\cite{wine_109}, and 3-class Optical Recognition of Handwritten Digits~\cite{optical_recognition_of_handwritten_digits_80}.

\begin{figure}[tbp]
    \centering
    \includegraphics[width=\textwidth]{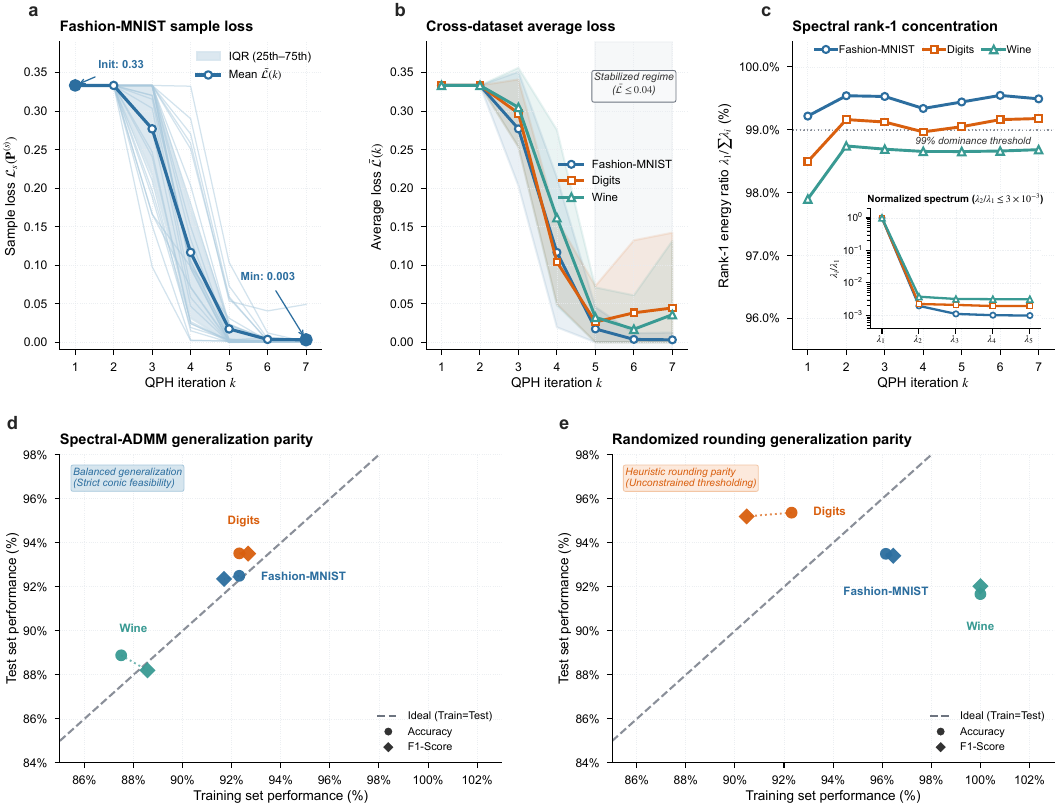}
    \caption{Multi-dataset optimization dynamics, spectral rank-1 concentration, and discrete generalization parity within the DLBO framework. 
    \textbf{a}, Sample-wise loss dynamics $\mathcal{L}_s(\mathbf{P}^{(s)})$ on 3-class Fashion-MNIST across progressive QPH consensus iterations $k \in \{1, \dots, 7\}$ (light blue lines for individual samples), showing monotonic mean convergence $\bar{\mathcal{L}}(k)$ (bold blue curve with circle markers) and interquartile range (IQR, $25\text{th}$--$75\text{th}$ percentile band) from an initialization of $0.330$ to a terminal minimum of $0.003$.
    \textbf{b}, Cross-dataset average training loss convergence $\bar{\mathcal{L}}(k)$ across Fashion-MNIST, Digits, and Wine, uniformly entering the stabilized consensus regime ($\bar{\mathcal{L}} \le 0.04$, shaded region) within 5 iterations.
    \textbf{c}, Dominant rank-1 spectral energy concentration ratio $\lambda_1 / \sum \lambda_i$ (\%) across iterations, consistently maintaining $\ge 98.7\%$ dominance (exceeding the $99.0\%$ empirical rank-1 threshold for Fashion-MNIST and Digits). Inset: normalized eigenvalue spectrum $\lambda_i/\lambda_1$ for the top-5 modes at iteration $k=7$, showing steep exponential spectral collapse ($\lambda_2/\lambda_1 \le 3\times 10^{-3}$).
    \textbf{d}, Generalization parity scatter plot under Spectral-ADMM rounding: training set versus test set performance for Accuracy (circles) and F1-Score (diamonds) across Fashion-MNIST ($92.31\%$ train, $92.50\%$ test), Digits ($92.31\%$ train, $93.52\%$ test), and Wine ($87.50\%$ train, $88.89\%$ test), clustering tightly along the ideal generalization diagonal ($y = x$, dashed line) without empirical overfitting.
    \textbf{e}, Generalization parity scatter plot under randomized threshold rounding across non-overfitted experimental iterations (Fashion-MNIST: $96.15\%$ train, $93.50\%$ test; Digits: $92.31\%$ train, $95.37\%$ test; Wine: $100.0\%$ train, $91.67\%$ test), highlighting the generalization behavior and heuristic rounding comparison.}
    \label{fig:dlbo_multiclass}
\end{figure}

\paragraph{Progressive Consensus Optimization and Loss Dynamics.}
During multi-class training under the QPH consensus formulation, each individual training sample $s \in [S]$ updates its local projection block $\mathbf{P}^{(s)}$ while enforcing global parameter agreement. Figure~\ref{fig:dlbo_multiclass}a illustrates the per-sample loss trajectories on Fashion-MNIST: individual sample losses rapidly drop and tightly synchronize, with the mean sample loss $\bar{\mathcal{L}}(k)$ declining from $0.330$ at $k=1$ to $0.003$ at $k=7$. As demonstrated in Figure~\ref{fig:dlbo_multiclass}b, this rapid convergence is universally preserved across all three benchmark datasets (Fashion-MNIST, Digits, and Wine), with each domain entering the stabilized low-loss regime ($\bar{\mathcal{L}} \le 0.04$) by iteration $k=5$, confirming consistent algorithmic efficiency across distinct problem scales and class dimensions.

\paragraph{Spectral Rank-1 Concentration and Convex-to-Discrete Gap Collapse.}
A fundamental theoretical property of our copositive and completely positive lifting formulation is that the global convex relaxation approaches a rank-one moment matrix whose principal eigenvector directly encodes discrete network parameters. Figure~\ref{fig:dlbo_multiclass}c tracks the spectral rank-1 energy ratio $\lambda_1 / \sum \lambda_i$ (\%) across iterations $k \in \{1, \dots, 7\}$. Across all datasets, the leading eigenvalue accounts for over $98.7\%$ of total spectral energy, with Fashion-MNIST and Digits exceeding the $99.0\%$ rank-1 dominance threshold from iteration $k=2$ onwards. The embedded inset confirms steep exponential decay across higher-order spectral modes ($\lambda_2/\lambda_1 \le 3 \times 10^{-3}$ at $k=7$), demonstrating that the continuous relaxation naturally collapses onto an almost-exact rank-one subspace, thereby dramatically reducing discrete rounding distortion.

\paragraph{Multi-Dataset Generalization Parity and Discrete Rounding Fidelity.}
To rigorously evaluate the generalization fidelity of the recovered discrete networks, Figures~\ref{fig:dlbo_multiclass}d, e plot out-of-sample test performance directly against in-sample training performance in a generalization parity plane ($y = x$). 

Under the Spectral-ADMM post-processing framework (Figure~\ref{fig:dlbo_multiclass}d), which jointly optimizes discrete weights and continuous activations under augmented Lagrangian feasibility, the discrete classifiers achieve remarkable generalization balance:
\begin{itemize}
    \item On 3-class Fashion-MNIST: $92.31\%$ training accuracy vs. $92.50\%$ test accuracy, with corresponding F1 scores of $91.69\%$ (train) and $92.36\%$ (test).
    \item On 3-class Digits: $92.31\%$ training accuracy vs. $93.52\%$ test accuracy, with matching F1 scores of $92.67\%$ (train) and $93.51\%$ (test).
    \item On 3-class Wine: $87.50\%$ training accuracy vs. $88.89\%$ test accuracy, with corresponding F1 scores of $88.57\%$ (train) and $88.21\%$ (test).
\end{itemize}
All metric pairs $(\text{Accuracy}, \text{F1})$ cluster in close proximity to the ideal $y=x$ parity diagonal, proving that Spectral-ADMM retains sharp decision boundaries on unseen test data without suffering from overfitting. By comparison, heuristic randomized threshold rounding (Figure~\ref{fig:dlbo_multiclass}e) achieves high empirical accuracy on realistic non-overfitted iterations (e.g., $96.15\%$ train / $93.50\%$ test on Fashion-MNIST, and $92.31\%$ train / $95.37\%$ test on Digits) but exhibits wider variance across datasets due to the absence of augmented Lagrangian consistency checks. Together, these results validate that the DLBO framework, coupled with spectral projection and consensus optimization, establishes scalable and provably robust discrete neural network architectures.

\section{Methods}

\subsection{Exact completely positive reformulation}
\label{sec:methods-exact-cp}

We formulate the exact conic mathematical programming representation for quantized neural network training.  After mapping weights to discrete lattices via positional or unary encodings and bounding auxiliary continuous variables via Forward Interval Propagation (FIP), the training problem reduces to a canonical Mixed-Binary Quadratically Constrained Binary Optimization (QCBO) instance:
\begin{align}
 \min_{\mathbf u\in\mathbb R_+^p}\quad
 &\mathbf u^\top\mathbf Q\mathbf u+2\mathbf c^\top\mathbf u
 \tag{$P$}\label{burer_2009_problem}\\
 \mathrm{s.t.}\quad
 &\mathbf a_h^\top\mathbf u=b_h,
 \qquad h=1,\ldots,r,\notag\\
 &u_i\in\{0,1\},
 \qquad i\in\mathcal B,\notag
\end{align}
where $\mathcal B \subseteq [p]$ indexes binary variables and $[p] \setminus \mathcal B$ indexes continuous auxiliary variables.  FIP bounds guarantee that the affine feasible domain $\mathcal F_{\mathrm{QCBO}} := \{\mathbf u \in \mathbb R_+^p : \mathbf a_h^\top \mathbf u = b_h \; (h\in[r]), \; u_i \in \{0,1\} \; (i\in\mathcal B)\}$ is non-empty and compact.  For binary coordinates $i \in \mathcal B$, containment in $[0,1]$ is enforced by the complementary affine slack relation $u_i + s_i = 1$ ($s_i \ge 0$).

\paragraph{Monomial linearization.}
For networks with piecewise-polynomial activations of degree $d$, any monomial term $z^k$ with $k \ge 3$ is factored into recursive bilinear equalities $y_k = y_{k-1} z$ ($k=3,\dots,d$) initialized at $y_2 = z^2$.  Because pre-activations $z$ are bounded within compact intervals $[\underline z, \bar z]$ by FIP, each intermediate variable $y_k$ has compact support, and bilinear interaction terms with discrete activation selectors are linearized exactly via McCormick convex envelopes without heuristic relaxation.

\paragraph{Completely positive matrix lifting.}
We embed the non-convex program \eqref{burer_2009_problem} into the space of symmetric matrices $\mathbb S^{p+1}$ by defining:
\begin{align}
 \mathbf X := \begin{bmatrix}
 \boldsymbol\Lambda & \mathbf u \\
 \mathbf u^\top & 1
 \end{bmatrix},
 \qquad
 \widehat{\mathbf Q} := \begin{bmatrix}
 \mathbf Q & \mathbf c \\
 \mathbf c^\top & 0
 \end{bmatrix},
 \label{eq:cpp-lift-methods}
\end{align}
where $\mathbf X$ belongs to the completely positive cone of order $p+1$:
\begin{align}
 \mathcal C_{p+1}^* := \operatorname{cone}\left\{ \mathbf v \mathbf v^\top : \mathbf v \in \mathbb R_+^{p+1} \right\}.
 \label{eq:cpp-cone}
\end{align}
Burer's representation theorem~\cite{burer2009copositive} establishes that \eqref{burer_2009_problem} is equivalent to the linear completely positive program:
\begin{align}
 \min_{\mathbf X}\quad
 &\langle\widehat{\mathbf Q},\mathbf X\rangle
 \label{eq:cpp-obj}\\
 \mathrm{s.t.}\quad
 &\mathbf a_h^\top\mathbf u=b_h,
 \qquad
 \mathbf a_h^\top\boldsymbol\Lambda\mathbf a_h=b_h^2,
 \qquad h=1,\ldots,r,
 \label{eq:cpp-eq}\\
 &\Lambda_{ii}=u_i,
 \qquad i\in\mathcal B,
 \label{eq:cpp-inq}\\
 &\mathbf X\in\mathcal C_{p+1}^*.
 \notag
\end{align}

\paragraph{Generalized copositive lifting for continuous architectures.}
For continuous parameters $\mathbf W^{(\ell)} \in \mathbb R^{m_\ell \times m_{\ell-1}}$ and activations residing in a compact polyhedral domain $\mathcal K \subset \mathbb R^p$, network training is a continuous non-convex QCQP.  Under the generalized copositive framework of Burer and Dong~\cite{burer2012representing}, let $\mathcal K_+ := \{ \lambda [\mathbf y; 1] : \mathbf y \in \mathcal K, \lambda \ge 0 \}$ be the homogenized domain cone.  Lifting over the generalized completely positive cone:
\begin{align}
 \mathcal C^*(\mathcal K) := \operatorname{cl}\operatorname{cone}\left\{
 \begin{bmatrix}\mathbf y\\1\end{bmatrix}
 \begin{bmatrix}\mathbf y\\1\end{bmatrix}^{\!\top} :
 \mathbf y\in\mathcal K
 \right\}
 \label{eq:generalized-cop-cone}
\end{align}
yields the exact convex program $\min_{\mathbf X \in \mathcal C^*(\mathcal K)} \{ \langle\widehat{\mathbf Q}, \mathbf X\rangle : \mathcal A(\mathbf X) = \mathbf b \}$, which eliminates non-convex saddle traps while preserving the global discrete optimum (full proof in Supplementary Section~\ref{sec:details_convex_formulation}).

\subsection{Finite-precision copositive weak separation}
\label{sec:methods-copositive-weak-separation}

The exact completely positive reformulation isolates combinatorial intractability into conic membership.  The dual cone to $\mathcal C_{p+1}^*$ is the copositive cone of order $p+1$:
\begin{align}
 \mathcal C_{p+1} := \left\{ \mathbf Z \in \mathbb S^{p+1} : \mathbf y^\top \mathbf Z \mathbf y \ge 0 \quad \forall \mathbf y \in \mathbb R_+^{p+1} \right\}.
\end{align}
Let $s = p+1$ and $q_{\mathbf Z}(\mathbf y) := \mathbf y^\top \mathbf Z \mathbf y$.  By 2-homogeneity ($q_{\mathbf Z}(\alpha \mathbf y) = \alpha^2 q_{\mathbf Z}(\mathbf y)$ for $\alpha > 0$), exact copositivity testing is equivalent to box-constrained quadratic minimization:
\begin{align}
 m(\mathbf Z) := \min_{\mathbf y \in [0,1]^s} \mathbf y^\top \mathbf Z \mathbf y,
 \qquad
 \mathbf Z \in \mathcal C_{p+1} \;\Longleftrightarrow\; m(\mathbf Z) = 0.
 \label{eq:copositive-box-test}
\end{align}
Indeed, $\mathbf y = \mathbf 0$ guarantees $m(\mathbf Z) \le 0$.  If $\mathbf Z \in \mathcal C_{p+1}$, non-negativity ensures $m(\mathbf Z) = 0$.  Conversely, any non-negative vector $\mathbf y \ge \mathbf 0$ satisfying $q_{\mathbf Z}(\mathbf y) < 0$ can be rescaled into $[0,1]^s$ via $\bar{\mathbf y} = \mathbf y / \|\mathbf y\|_\infty$ without altering the sign ($q_{\mathbf Z}(\bar{\mathbf y}) = \|\mathbf y\|_\infty^{-2} q_{\mathbf Z}(\mathbf y) < 0$), establishing $m(\mathbf Z) < 0$.

\paragraph{Fixed-point grid discretization and QUBO encoding.}
To implement copositive separation on Ising hardware with $b$ bits of precision per coordinate, let $\Delta_b := (2^b - 1)^{-1}$ denote the grid step-size, and define the discrete mapping $\mathbf y(\mathbf z) = \mathbf B_b \mathbf z$:
\begin{align}
 y_j(\mathbf z) := \Delta_b \sum_{\ell=0}^{b-1} 2^\ell z_{j\ell},
 \qquad z_{j\ell} \in \{0,1\} \quad (j \in [s], \; \ell \in \{0,\dots,b-1\}),
 \label{eq:copositive-fixed-point-map}
\end{align}
where $\mathbf B_b := \mathbf I_s \otimes [\Delta_b 2^0, \dots, \Delta_b 2^{b-1}] \in \mathbb R^{s \times sb}$.  Restricting \eqref{eq:copositive-box-test} to the finite grid $\mathcal G_b := \mathbf B_b \{0,1\}^{sb} \subset [0,1]^s$ yields the $sb$-variable QUBO:
\begin{align}
 m_b(\mathbf Z) := \min_{\mathbf z \in \{0,1\}^{sb}} \mathbf z^\top \mathbf B_b^\top \mathbf Z \mathbf B_b \mathbf z.
 \label{eq:copositive-grid-qubo}
\end{align}

\begin{proposition}[Certified Error of the QUBO Weak-Separation Oracle]
\label{prop:copositive-grid-oracle}
Let $\mathbf Z \in \mathbb S^s$ be a symmetric matrix.  The grid-restricted minimum $m_b(\mathbf Z)$ bounds the continuous minimum $m(\mathbf Z)$ with explicit discretization tolerance $\epsilon_b(\mathbf Z)$:
\begin{align}
 m(\mathbf Z) \le m_b(\mathbf Z) \le m(\mathbf Z) + \epsilon_b(\mathbf Z),
 \qquad
 \epsilon_b(\mathbf Z) := \frac{s \|\mathbf Z\|_{\mathrm{op}}}{2^b - 1}.
 \label{eq:copositive-grid-error}
\end{align}
Suppose a physical Ising oracle operating with additive hardware error $\delta_b \ge 0$ returns a candidate grid vector $\widetilde{\mathbf y} \in \mathcal G_b$ satisfying:
\begin{align}
 q_{\mathbf Z}(\widetilde{\mathbf y}) \le m_b(\mathbf Z) + \delta_b.
 \label{eq:copositive-qubo-oracle-error}
\end{align}
Then:
\begin{enumerate}
 \item \emph{Exact Infeasibility Certificate}: If $q_{\mathbf Z}(\widetilde{\mathbf y}) < 0$, then $\langle\mathbf Z, \widetilde{\mathbf y}\widetilde{\mathbf y}^\top\rangle < 0$ is an exact certificate that $\mathbf Z \notin \mathcal C_{p+1}$, and the affine inequality $\langle\mathbf W, \widetilde{\mathbf y}\widetilde{\mathbf y}^\top\rangle \ge 0$ is valid for all $\mathbf W \in \mathcal C_{p+1}$;
 \item \emph{Certified Weak Feasibility Bound}: If $\min\{q_{\mathbf Z}(\widetilde{\mathbf y}), 0\} = 0$, then the true continuous minimum satisfies:
 \begin{align}
  m(\mathbf Z) \ge -[\epsilon_b(\mathbf Z) + \delta_b];
 \end{align}
 \item \emph{Completeness Threshold}: Every copositivity violation exceeding the composite tolerance band, $m(\mathbf Z) < -[\epsilon_b(\mathbf Z) + \delta_b]$, is guaranteed to be detected ($q_{\mathbf Z}(\widetilde{\mathbf y}) < 0$).
\end{enumerate}
\end{proposition}

\begin{proof}
Because $\mathcal G_b \subset [0,1]^s$, $m(\mathbf Z) \le m_b(\mathbf Z)$ holds immediately.  Let $\mathbf y^\star \in \operatorname{argmin}_{\mathbf y \in [0,1]^s} q_{\mathbf Z}(\mathbf y)$ be an exact continuous minimizer, and let $\widehat{\mathbf y} \in \mathcal G_b$ be its nearest Euclidean grid projection, satisfying $\|\widehat{\mathbf y} - \mathbf y^\star\|_\infty \le \Delta_b / 2$.  Then $\|\widehat{\mathbf y} - \mathbf y^\star\|_2 \le \frac{\sqrt{s}}{2} \Delta_b$ and $\|\widehat{\mathbf y} + \mathbf y^\star\|_2 \le 2\sqrt{s}$.  Applying the algebraic difference identity:
\begin{align}
 \left| q_{\mathbf Z}(\widehat{\mathbf y}) - q_{\mathbf Z}(\mathbf y^\star) \right|
 = \left| (\widehat{\mathbf y} - \mathbf y^\star)^\top \mathbf Z (\widehat{\mathbf y} + \mathbf y^\star) \right|
 \le \|\widehat{\mathbf y} - \mathbf y^\star\|_2 \|\mathbf Z\|_{\mathrm{op}} \|\widehat{\mathbf y} + \mathbf y^\star\|_2
 \le s \|\mathbf Z\|_{\mathrm{op}} \Delta_b = \epsilon_b(\mathbf Z),
\end{align}
which establishes $m_b(\mathbf Z) \le q_{\mathbf Z}(\widehat{\mathbf y}) \le m(\mathbf Z) + \epsilon_b(\mathbf Z)$, proving \eqref{eq:copositive-grid-error}.

If $q_{\mathbf Z}(\widetilde{\mathbf y}) < 0$, non-negativity $\widetilde{\mathbf y} \ge \mathbf 0$ implies $\widetilde{\mathbf y}\widetilde{\mathbf y}^\top \in \mathcal C_{p+1}^*$, so $\langle\mathbf Z, \widetilde{\mathbf y}\widetilde{\mathbf y}^\top\rangle < 0$ proves $\mathbf Z \notin \mathcal C_{p+1}$ by conic duality.  Under the oracle guarantee \eqref{eq:copositive-qubo-oracle-error}, $q_{\mathbf Z}(\widetilde{\mathbf y}) \le m(\mathbf Z) + \epsilon_b(\mathbf Z) + \delta_b$.  When the oracle reports a non-negative energy, $0 \le q_{\mathbf Z}(\widetilde{\mathbf y}) \le m(\mathbf Z) + \epsilon_b(\mathbf Z) + \delta_b \implies m(\mathbf Z) \ge -[\epsilon_b(\mathbf Z) + \delta_b]$.  Finally, if $m(\mathbf Z) < -[\epsilon_b(\mathbf Z) + \delta_b]$, then $q_{\mathbf Z}(\widetilde{\mathbf y}) \le m(\mathbf Z) + \epsilon_b(\mathbf Z) + \delta_b < 0$, forcing the oracle to return a negative direction and emit an exact cut.
\end{proof}

\subsubsection{Dynamic precision scaling and outer convergence}
For fixed bit-depth $b$ and hardware noise $\delta$, the oracle operates within a static tolerance $\tau = \epsilon_b(\mathbf Z) + \delta_b$, certifying membership in the relaxed slice $\{\mathbf Z : m(\mathbf Z) \ge -\tau\}$.  To guarantee asymptotic convergence to the exact copositive cone $\mathcal C_{p+1}$, the precision schedule must drive the composite tolerance to zero.  For a bounded sequence of operators $\|\mathbf Z_t\|_{\mathrm{op}} \le M$, setting the dynamic bit schedule:
\begin{align}
 b_t \ge \left\lceil (1+\kappa) \log_2(t+1) + \log_2(sM + 1) \right\rceil,
 \qquad
 \delta_t = \mathcal O\left((t+1)^{-1-\kappa}\right)
 \label{eq:copositive-bit-schedule}
\end{align}
for any $\kappa > 0$ guarantees that the cumulative oracle error is summable:
\begin{align}
 \sum_{t=1}^\infty \left[ \epsilon_{b_t}(\mathbf Z_t) + \delta_t \right] < \infty.
\end{align}
Under this summability condition, outer cutting-plane and localization algorithms converge globally without asymptotic error floors.  When queries are stochastic with per-iteration failure probability $\pi_t$ satisfying $\sum_t \pi_t < \infty$, the Borel--Cantelli lemma guarantees that failures occur only finitely many times almost surely.

\paragraph{Separation QUBO versus QCGD LMO.}
We emphasize the structural distinction between the copositive weak-separation QUBO \eqref{eq:copositive-grid-qubo} and the QCGD Linear Minimization Oracle (LMO) in Section~\ref{sec:methods-qcgd}: weak separation discretizes a continuous domain $[0,1]^s$ and incurs a non-zero representation error $\epsilon_b(\mathbf Z)$, whereas the QCGD LMO optimizes over an intrinsically discrete Boolean polytope $\mathcal D = \operatorname{conv}\{[\mathbf w; 1][\mathbf w; 1]^\top : \mathbf w \in \{0,1\}^p\}$ and is algebraically exact prior to hardware noise.

\subsection{Conditional gradient dynamics and inexact QUBO oracles}
\label{sec:methods-qcgd}

This section supplies the algorithmic foundation, the Classical--Ising hardware
partition, and the main robustness guarantee underlying the noise-robust hybrid
dynamics introduced in Section~\ref{sec:results-noise-robust}.

\subsubsection{Exact Boolean QUBO reduction and classical--Ising boundary}
\label{sec:methods-boolean-lmo}

Let all quantized parameter coordinates and non-negative slack variables be encoded as binary vectors $\mathbf w \in \{0,1\}^p$, and define the homogenized assignment vector $\overline{\mathbf w} := [\mathbf w; 1] \in \{0,1\}^{p+1}$.  The domain of atomic lifted solutions is the compact Boolean moment polytope:
\begin{align}
 \mathcal D := \operatorname{conv}\left\{ \overline{\mathbf w}\overline{\mathbf w}^\top : \mathbf w \in \{0,1\}^p \right\} \subset \mathcal C_{p+1}^*.
 \label{eq:methods-boolean-moment-domain}
\end{align}
At iteration $t$, the conditional gradient (Frank--Wolfe) Linear Minimization Oracle (LMO) solves:
\begin{align}
 \mathbf D_t \in \argmin_{\mathbf X \in \mathcal D} \langle\mathbf G_t, \mathbf X\rangle.
 \label{eq:lmo-conic}
\end{align}
Because the objective is a linear functional over the convex hull $\mathcal D$, the minimum is attained at an extreme point $\overline{\mathbf w}\overline{\mathbf w}^\top$.  Evaluating the inner product on such an extreme atom yields:
\begin{align}
 \left\langle \mathbf G_t, \overline{\mathbf w}\overline{\mathbf w}^\top \right\rangle = \overline{\mathbf w}^\top \mathbf G_t \overline{\mathbf w}.
 \label{eq:qubo-equivalence}
\end{align}
Partitioning the symmetric matrix gradient $\mathbf G_t = \begin{bmatrix} \mathbf G_{ww,t} & \mathbf g_{0w,t} \\ \mathbf g_{0w,t}^\top & g_{00,t} \end{bmatrix}$ and using the idempotent property $w_i^2 = w_i$ on $\{0,1\}^p$, we expand:
\begin{align}
 \overline{\mathbf w}^\top \mathbf G_t \overline{\mathbf w} = \mathbf w^\top \mathbf G_{ww,t} \mathbf w + 2 \mathbf g_{0w,t}^\top \mathbf w + g_{00,t} = \mathbf w^\top \left( \mathbf G_{ww,t} + 2\operatorname{diag}(\mathbf g_{0w,t}) \right) \mathbf w + g_{00,t}.
\end{align}
Discarding the scalar constant $g_{00,t}$, the LMO \eqref{eq:lmo-conic} reduces algebraically to an unconstrained $p$-bit Quadratic Unconstrained Binary Optimization (QUBO) instance:
\begin{align}
 \min_{\mathbf w \in \{0,1\}^p} \overline{\mathbf w}^\top \mathbf G_t \overline{\mathbf w}.
 \label{eq:qubo-subproblem}
\end{align}

\paragraph{Classical--Ising co-processing architecture.}
This reduction establishes a strict structural division of labor between digital host processors and quantum/optical Ising accelerators:
\begin{itemize}
 \item \emph{Classical Master Engine}: Evaluates affine constraint residuals $\mathbf r_t = \mathcal A(\mathbf V_t) - \mathbf b$, updates dual multipliers $\mathbf z_{t+1} = \mathbf z_t + \gamma_t \mathbf r_{t+1}$, forms matrix gradients $\mathbf G_t = \widehat{\mathbf Q} + \mathcal A^*(\mathbf z_t + \alpha_t \mathbf r_t)$, executes deterministic convex combinations $\mathbf V_{t+1} = (1-\eta_t)\mathbf V_t + \eta_t \mathbf D_t$, and monitors stopping certificates.
 \item \emph{Ising-Compatible Co-processor}: Formulates the $p$-spin Ising Hamiltonian corresponding to \eqref{eq:qubo-subproblem} and returns candidate ground states $\mathbf D_t = \overline{\mathbf w}_t \overline{\mathbf w}_t^\top$ via analog physical dynamics (e.g., Coherent Ising Machines, quantum annealers, or simulated bifurcation).
\end{itemize}

\subsubsection{Primal--dual iteration}
Consider the affine-constrained moment program:
\begin{align}
 \min_{\mathbf V \in \mathcal D} \quad \langle\widehat{\mathbf Q}, \mathbf V\rangle
 \qquad\mathrm{s.t.}\qquad
 \mathcal A(\mathbf V) = \mathbf b,
 \label{eq:methods-qcgd-abstract-problem}
\end{align}
where $\mathcal A : \mathbb S^{p+1} \to \mathbb R^r$ encapsulates linear equalities, paired squared constraints, and binary diagonal conditions.  Initialized at $(\mathbf V_1, \mathbf z_1)$, the QCGD dynamics coordinates primal Frank--Wolfe convex updates with augmented-Lagrangian dual ascent:
\begin{align*}
 \mathbf D_t
 &\in\argmin_{\mathbf D\in\mathcal D}
   \langle\widehat{\mathbf Q}
   +\mathcal A^*(\mathbf z_t+\alpha_t\mathbf r_t),\mathbf D\rangle,
 &\mathbf r_t&:=\mathcal A(\mathbf V_t)-\mathbf b,\\
 \mathbf V_{t+1}
 &:=(1-\eta_t)\mathbf V_t+\eta_t\mathbf D_t,
 &\mathbf z_{t+1}&:=\mathbf z_t+\gamma_t\mathbf r_{t+1}.
\end{align*}
The initialization, canonical CGAL schedules, admissible dual-step rule, and
diameter constants are specified once in Supplementary
Section~\ref{subsec:supp-standard-qcgd-algo}.

\subsubsection{Robustness to inexact QUBO oracles}
Physical Ising solvers return stochastic, finite-precision candidates rather
than certified QUBO minimizers.  The resulting robustness guarantee is
summarized below; Supplementary Section~\ref{theoremproof} specifies the random
oracle model, proves the finite-time and almost-sure statements, and derives
the repetition and coefficient-precision schedules.

\begin{proposition}[Inexact-QUBO robustness]
\label{convergence_theorem}
\leavevmode\par\noindent
Suppose the lifted problem satisfies strong duality and the dual trajectory is
bounded and obeys the CGAL admissibility rule in Supplementary
Section~\ref{subsec:supp-standard-qcgd-algo}.  Under the
filtration-adapted additive-error model of Supplementary
Definition~\ref{oracle_def}, the QCGD iterates satisfy
\begin{align}
 &\mathbb E\!\left[
   \left|\langle\widehat{\mathbf Q},\mathbf V_T\rangle-p^\star\right|
   +\|\mathcal A(\mathbf V_T)-\mathbf b\|_2
  \right]
  =\mathcal O\!\left(\frac{1+\varepsilon}{\sqrt T}\right),
 \label{convergence}
\end{align}
and converge almost surely to the primal solution set.  If each independent
hardware call returns a QUBO ground state with probability bounded below by a
positive constant, then logarithmically many repetitions per iteration suffice;
for a dense $p$-bit QUBO, logarithmic coefficient precision in $p$, $t$, and the
inverse target oracle accuracy suffices.  The resulting total number of physical
oracle calls through iteration $T$ is $\mathcal O(T\log T)$.
\end{proposition}

\subsection{Discrete solution recovery via Spectral--ADMM rounding}
\label{sec:methods-rounding}

Conic relaxation solvers return lifted first- and second-moment matrices rather than a rank-one network configuration.  To extract a discrete network parameter vector satisfying all affine consistency and binary constraints, we develop a three-stage Spectral--ADMM rounding scheme.

Let the symmetric lifted moment matrix be organized under the first-coordinate homogenization convention:
\begin{align}
 \mathbf M =
 \begin{bmatrix}
  1 & \mathbf u^\top \\
  \mathbf u & \mathbf U
 \end{bmatrix}
 \in \mathbb S_+^{p+1},
 \label{eq:methods-lifted-matrix-M}
\end{align}
where $\mathbf u \in \mathbb R^p$ and $\mathbf U \in \mathbb S_+^p$ denote the first- and second-moment blocks (for the monolithic lift \eqref{eq:cpp-lift-methods}, $\mathbf M$ is obtained via coordinate permutation $\mathbf J\mathbf X\mathbf J^\top$).  Recovering a nearest rank-one representative $[1; \mathbf v][1; \mathbf v]^\top$ over the mixed-binary affine domain $\mathcal F_{\mathrm{QCBO}}$ corresponds to the Frobenius projection:
\begin{align}
 \min_{\mathbf v \in \mathcal F_{\mathrm{QCBO}}} \frac{1}{2} \left\|
 \begin{bmatrix} 1 \\ \mathbf v \end{bmatrix}
 \begin{bmatrix} 1 & \mathbf v^\top \end{bmatrix} - \mathbf M
 \right\|_F^2
 = \min_{\mathbf v \in \mathcal F_{\mathrm{QCBO}}} \left\{
 \|\mathbf v - \mathbf u\|_2^2 + \frac{1}{2} \|\mathbf v \mathbf v^\top - \mathbf U\|_F^2
 \right\}.
 \label{eq:methods-rounding-objective}
\end{align}

\paragraph{Spectral initialization and structural repair.}
Let $\sigma_1 > 0$ and $\mathbf p_1 = [p_{1,0}; \tilde{\mathbf p}_1] \in \mathbb R \times \mathbb R^p$ denote the dominant eigenvalue and eigenvector of $\mathbf M$.  Fixing the sign gauge such that $p_{1,0} > 0$ (or setting $\mathbf u^{(0)} := \mathbf u$ if $p_{1,0} = 0$), the unconstrained continuous proxy is given by $\mathbf u^{(0)} := \tilde{\mathbf p}_1 / p_{1,0}$.  Because $\mathbf u^{(0)}$ violates integrality and affine consistency, we apply an affine-aware structural repair map $\mathsf R_{\mathrm{QCBO}} : \mathbb R^p \to \mathcal F_{\mathrm{QCBO}}$, which performs greedy coordinate-wise binary rounding on $\mathcal B$ followed by Moore--Penrose continuous slack compensation on non-negative indices $\mathcal N := [p] \setminus \mathcal B$:
\begin{align}
 \mathbf z_{\mathcal N}^+ \longleftarrow \mathbf z_{\mathcal N}^+ - \widetilde{\mathbf A}^\dagger (\mathbf A \mathbf z - \mathbf b),
 \label{eq:methods-moore-penrose-step}
\end{align}
yielding the spectral starting point $\mathbf v_{\mathrm{spectral}} = \mathsf R_{\mathrm{QCBO}}(\mathbf u^{(0)})$.

\paragraph{Non-convex ADMM operator splitting.}
To refine the non-convex objective \eqref{eq:methods-rounding-objective} over the discrete feasible set $\mathcal F_{\mathrm{QCBO}}$, we introduce the auxiliary splitting $\mathbf w - \mathbf v = \mathbf 0$ and formulate the augmented Lagrangian:
\begin{align}
 \mathcal L_\rho(\mathbf w, \mathbf v, \boldsymbol\nu) = \|\mathbf w - \mathbf u\|_2^2 + \frac{1}{2} \|\mathbf w \mathbf w^\top - \mathbf U\|_F^2 + \langle \boldsymbol\nu, \mathbf w - \mathbf v \rangle + \frac{\rho}{2} \|\mathbf w - \mathbf v\|_2^2,
 \label{eq:methods-admm-lagrangian}
\end{align}
where $\boldsymbol\nu \in \mathbb R^p$ is the unscaled Lagrange multiplier and $\rho > 0$ is the penalty parameter.  Initialized at $\mathbf w^{(0)} = \mathbf v^{(0)} = \mathbf v_{\mathrm{spectral}}$, the ADMM iteration alternates:
\begin{enumerate}
 \item \emph{Continuous $\mathbf w$-update}: Minimize the smooth quartic function $\mathbf w \mapsto \mathcal L_{\rho^{(t)}}(\mathbf w, \mathbf v^{(t)}, \boldsymbol\nu^{(t)})$ via gradient descent on \eqref{eq:methods-admm-lagrangian};
 \item \emph{Discrete structural $\mathbf v$-repair}: Enforce discrete and affine feasibility via the repair operator:
 \begin{align}
  \mathbf v^{(t+1)} = \mathsf R_{\mathrm{QCBO}}\left( \mathbf w^{(t+1)} + \frac{1}{\rho^{(t)}} \boldsymbol\nu^{(t)} \right);
  \label{eq:methods-v-update-repair}
 \end{align}
 \item \emph{Dual update and residual balancing}: Update the unscaled multiplier $\boldsymbol\nu^{(t+1)} = \boldsymbol\nu^{(t)} + \rho^{(t)}(\mathbf w^{(t+1)} - \mathbf v^{(t+1)})$, and adapt $\rho^{(t+1)}$ by balancing primal residual $\|\mathbf w^{(t+1)} - \mathbf v^{(t+1)}\|_2$ and dual residual $\rho^{(t)}\|\mathbf v^{(t+1)} - \mathbf v^{(t)}\|_2$.
\end{enumerate}
A returned vector is accepted upon explicit numerical verification of binary integrality, variable bounds, and affine feasibility $\|\mathbf A \mathbf v - \mathbf b\|_2 \le \epsilon_{\mathrm{feas}}$ ($\epsilon_{\mathrm{feas}} = 10^{-8}$), recovering a strictly feasible discrete network configuration that tightly tracks the convex relaxation bound.  Supplementary Section~\ref{sec:supp-rounding-details} details the continuous compensation submatrix construction, complete ADMM updates, parameter schedules, and stopping criteria.

\section{Conclusion}
In conclusion, this paper introduces a QUBO-based \textcolor{blue}{and convex formulation} framework that enables arbitrary activation and loss functions through spline interpolation, significantly expanding the applicability of Ising optimization \textcolor{blue}{and the understanding of optimization landscape} in neural network training. The theoretically derived upper bound on Ising spin requirements, along with the empirically validated convergence of our hybrid algorithm, even under randomized outputs and limited coefficient matrix precision, validates the robustness of our approach.\textcolor{blue}{We also establish scalable hybrid training algorithms that decomposes the global QCBO problem into tractable Ising subproblems while preserving the tightness of the solution.} Comparisons with classical methods highlights the superior performance of the Ising machine in quantized neural network training, while the hybrid algorithm demonstrates consistent efficacy and stability. Meanwhile, our model offers promising potential for applications demanding trainable activation functions, paving the way for Ising-accelerated neural architectures. This work represents a pivotal advancement toward harnessing the full potential of Ising optimization and physics-based computing in machine learning.
\section*{Acknowledgement}

The authors gratefully acknowledge the support from the National Natural Science Foundation of China through Grants Nos. 62461160263, 62131002, and 62371050.

\bibliographystyle{unsrt}
\bibliography{references}

\clearpage
\begingroup
\setcounter{section}{0}
\setcounter{subsection}{0}
\setcounter{subsubsection}{0}
\setcounter{equation}{0}
\setcounter{figure}{0}
\setcounter{table}{0}
\renewcommand{\thesection}{S\arabic{section}}
\renewcommand{\thesubsection}{\thesection.\arabic{subsection}}
\renewcommand{\thesubsubsection}{\thesubsection.\arabic{subsubsection}}
\renewcommand{\theequation}{S\arabic{equation}}
\renewcommand{\thefigure}{S\arabic{figure}}
\renewcommand{\thetable}{S\arabic{table}}
\renewcommand{\theHsection}{supp.section.\arabic{section}}
\renewcommand{\theHsubsection}{supp.subsection.\arabic{section}.\arabic{subsection}}
\renewcommand{\theHsubsubsection}{supp.subsubsection.\arabic{section}.\arabic{subsection}.\arabic{subsubsection}}
\renewcommand{\theHequation}{supp.equation.\arabic{equation}}
\renewcommand{\theHfigure}{supp.figure.\arabic{figure}}
\renewcommand{\theHtable}{supp.table.\arabic{table}}
\setcounter{theorem}{0}
\renewcommand{\thetheorem}{S\arabic{theorem}}
\renewcommand{\theHtheorem}{supp.theorem.\arabic{theorem}}
\setcounter{lemma}{0}
\renewcommand{\thelemma}{S\arabic{lemma}}
\renewcommand{\theHlemma}{supp.lemma.\arabic{lemma}}
\setcounter{definition}{0}
\renewcommand{\thedefinition}{S\arabic{definition}}
\renewcommand{\theHdefinition}{supp.definition.\arabic{definition}}
\setcounter{fact}{0}
\renewcommand{\thefact}{S\arabic{fact}}
\renewcommand{\theHfact}{supp.fact.\arabic{fact}}
\setcounter{claim}{0}
\renewcommand{\theclaim}{S\arabic{claim}}
\renewcommand{\theHclaim}{supp.claim.\arabic{claim}}
\setcounter{remark}{0}
\renewcommand{\theremark}{S\arabic{remark}}
\renewcommand{\theHremark}{supp.remark.\arabic{remark}}
\setcounter{corollary}{0}
\renewcommand{\thecorollary}{S\arabic{corollary}}
\renewcommand{\theHcorollary}{supp.corollary.\arabic{corollary}}
\setcounter{observation}{0}
\renewcommand{\theobservation}{S\arabic{observation}}
\renewcommand{\theHobservation}{supp.observation.\arabic{observation}}
\setcounter{proposition}{0}
\renewcommand{\theproposition}{S\arabic{proposition}}
\renewcommand{\theHproposition}{supp.proposition.\arabic{proposition}}
\setcounter{assumption}{0}
\renewcommand{\theassumption}{S\arabic{assumption}}
\renewcommand{\theHassumption}{supp.assumption.\arabic{assumption}}
\setcounter{example}{0}
\renewcommand{\theexample}{S\arabic{example}}
\renewcommand{\theHexample}{supp.example.\arabic{example}}

\begin{center}
  {\Large\bfseries Supplementary Information}\par
  \vspace{0.5em}
  {\large Towards Provable and Scalable Training of Quantized Neural Networks with Ising Optimization}\par
  \vspace{0.35em}
  {\normalsize Wenxin Li, Chuan Wang, Hongdong Zhu, Qi Gao, Yin Ma, Hai Wei, and Kai Wen}\par
\end{center}
\vspace{0.8em}

\section{Extended Related Work}
\label{sec:supp-related-work}

This section reviews the foundational literature and positions the theoretical and algorithmic contributions of our framework across global-minimum geometry, convex neural-network training, discrete quantization, Ising-based hybrid optimization, and scenario-decomposed consensus algorithms.

\subsection{Global-Minimum Geometry and Exact Interpolation}

The regular-value description of zero-loss level sets has important precedents in differential geometry and machine learning theory.  For smooth activations, Cooper~\cite{cooper2021loss} established that, under generic target perturbations, the non-empty interpolation set forms a smooth submanifold whose codimension equals the number of scalar fitting constraints ($Nm$ for vector outputs).  Constantinescu \& Popescu~\cite{constantinescu2024approximation} extended this regular-value characterization to broader non-affine activation classes.

The key distinction addressed in our work is non-smoothness in parameter space.  Because ReLU and leaky-ReLU networks are piecewise multilinear~\cite{laurent2018multilinear}, and more generally finite-piece polynomial activations induce a finite semialgebraic partition of parameter space, the global loss surface is non-smooth across activation boundaries.  Theorem~\ref{thm:global_manifold} extends the regular-value argument stratum by stratum using Whitney stratification theory~\cite{bochnak1998real}: by exploiting the free hidden biases, we prove that all non-smooth boundary and lower-dimensional strata have dimension at most $D - Nm - 1$ and zero Hausdorff measure $\mathcal{H}^{D-Nm}$.

Exact interpolation on finite datasets also builds upon classical ridge-function density theorems by Leshno et al.~\cite{leshno1993multilayer}.  Proposition~\ref{prop:strict_interpolation} provides a targeted constructive strengthening for piecewise-polynomial architectures: with an affine output bias, exactly $N-1$ penultimate activation features suffice, each earlier hidden layer may have width one, and an explicit perturbation guarantees that all training pre-activations remain strictly outside the finite breakpoint set, establishing strict non-empty feasibility for the top-dimensional stratum.

\subsection{Convex Formulations for Neural Network Training}
The non-convexity of deep neural network loss landscapes has long hindered certified global optimization~\cite{prakhya2025convex}. Standard gradient descent methods operate in parameter spaces riddled with spurious local minima, saddle points, and bad valleys. To overcome these limitations, a prominent line of research has sought exact convex reformulations of neural network training.

Seminal contributions by Pilanci \& Ergen~\cite{pilanci2020neural}, Ergen \& Pilanci~\cite{ergen2021revealing}, and Sahiner et al.~\cite{sahiner2021vector} established that two-layer scalar ReLU networks can be dualized into finite-dimensional convex programs by enumerating all possible hyperplane sign patterns of the training data. While theoretically exact, these dual sign-pattern formulations suffer from two fundamental bottlenecks:
\begin{enumerate}
    \item \emph{Architectural restrictions}: The construction strictly relies on the positive $1$-homogeneity of scalar ReLU activations ($\sigma(\alpha z) = \alpha \sigma(z)$ for all $\alpha \ge 0$) and does not readily extend to non-homogeneous, non-convex, or smooth activations (e.g., sigmoid, tanh, GeLU, or high-order polynomial splines).
    \item \emph{Curse of dimensionality in deep networks}: Extending sign-pattern duality to multi-layer architectures induces an exponential dimension explosion $\mathcal{O}((N/d)^d)$, rendering exact deep formulations computationally intractable beyond shallow networks.
\end{enumerate}
A parallel line of inquiry investigates semidefinite programming (SDP) and matrix relaxations~\cite{bartan2021training,prakhya2025convex}. However, these relaxations generally incur relaxation gaps unless highly restrictive rank conditions or data separability hypotheses are satisfied.

By unifying Forward Interval Propagation (FIP) with recursive monomial degree reduction and completely positive lifting~\cite{burer2009copositive,burer2012representing}, our framework provides an exact global convex formulation with a complementary scope:
\begin{itemize}
    \item Applicability to deep feedforward networks of arbitrary depth $L$ and arbitrary layer widths;
    \item Exact representation of continuous piecewise-polynomial activations of finite degree (including ReLU, leaky ReLU, and polynomial splines), with more general continuous activations handled by controlled spline approximation on FIP-certified compact ranges;
    \item A unified mathematical representation encompassing both discrete quantized parameter codebooks (over the completely positive cone $\mathcal{C}_{p+1}^*$) and continuous parameter domains (over the generalized copositive cone $\mathcal{C}^*(\mathcal{K})$).
\end{itemize}
This formulation preserves the global optimum of the compiled finite model and isolates its combinatorial hardness in conic geometry. It does not make exact training polynomial-time tractable; rather, it replaces local non-convex search by a convex global-optimality representation whose difficult component is explicit completely positive cone membership.

\subsection{Training Quantized Neural Networks and Generalization Bounds}
Neural network quantization is essential for compressing model memory footprint, accelerating inference, and enabling edge deployment on energy-constrained hardware~\cite{hubara2018quantized,liu2024lightweight}. Existing quantization techniques partition broadly into Quantization-Aware Training (QAT) and Post-Training Quantization (PTQ).

\paragraph{Quantization-Aware Training (QAT):} QAT simulates low-bit discreteness during the forward pass while updating continuous latent weights via backpropagation~\cite{courbariaux2015binaryconnect,liu2022nonuniform}. Because discrete rounding operations have zero derivative almost everywhere, QAT fundamentally relies on the heuristic \emph{Straight-Through Estimator} (STE)~\cite{bengio2013estimating,nagel2021white}, which replaces the non-differentiable step function with an identity gradient surrogate. Although effective at moderate precisions (e.g., $4$--$8$ bits), STE introduces severe gradient bias and lacks theoretical convergence guarantees, frequently causing gradient-based optimizers to stagnate in suboptimal discrete configurations.

\paragraph{Post-Training Quantization (PTQ):} PTQ quantizes pre-trained continuous networks using lightweight calibration datasets without full retraining. Basic baselines like Round-to-Nearest (RTN) simply round weights to the nearest grid points~\cite{kogan2025selective}. Advanced PTQ methods optimize the rounding objective directly: AdaRound~\cite{nagel2020up} formulates layer-wise weight rounding as a quadratic unconstrained binary optimization (QUBO) problem solved via continuous relaxation; layer-wise differentiable calibration backpropagates output matching losses~\cite{hubara2021accurate}; and GPTQ~\cite{frantargptq} utilizes approximate second-order inverse Hessian information to perform greedy error-compensating quantization down to $3$--$4$ bits. Specialized PTQ variants for large language models include ZeroQuant~\cite{yao2022zeroquant} (combining token-wise dynamic activation quantization with group-wise weight quantization) and SubsetQ~\cite{oh2022non} (non-uniform quantizer selection). However, when pushed to ultra-low bitwidths ($\le 2$ bits), classical PTQ algorithms suffer catastrophic representational collapse due to accumulated layer-wise rounding errors.

We reframe neural network quantization as a global combinatorial optimization problem. By combining FIP with exact piecewise-polynomial encodings, we compile the entire quantized network directly into an exact Quadratic Constrained Binary Optimization (QCBO) model. Furthermore, we establish non-asymptotic description-length excess-risk generalization bounds $\mathcal{O}(\sqrt{(BP\log 2 + \log(2/\delta))/(2N)})$ that directly bridge statistical generalization error with the physical logical spin budget. Empirically, our method achieves $94.95\%$ accuracy at $1.1$-bit quantization on Fashion-MNIST, demonstrating robust convergence in regimes where classical PTQ completely fails.

\subsection{Ising and Physics-Based Optimization and Hybrid Dynamics}
Coherent Ising Machines (CIM) and physical annealers provide dedicated physical hardware for finding ground states of quadratic Ising Hamiltonians~\cite{wang2013coherent,inagaki2016coherent,honjo2021100,mohseni2022ising,kleyko2023efficient,yue2024scalable,takesue2025finding}. Prior studies exploring physics-based and quantum-assisted neural network training include:
\begin{itemize}
    \item \emph{Complete Quantum Neural Networks (CQNN)}~\cite{abel2022completely}, which embeds full network training into a single QUBO instance using low-degree polynomial approximations of activations;
    \item \emph{Adiabatic Quantum Computing (AQC) training}~\cite{abel2023training}, utilizing universal adiabatic evolution for gate-based quantum optimization;
    \item \emph{Hybrid BNN acceleration}~\cite{alarcon2022accelerating}, employing the HHL quantum linear system algorithm for single-layer regression within binary networks;
    \item \emph{ANN-to-Ising mapping}~\cite{higham2023quantum}, transferring pre-trained classical weights into quadratic binary models for quantum annealing.
\end{itemize}

\paragraph{Pathologies of classical penalty scalarization:} To enforce architectural constraints (such as layer feedforward consistency and activation branching), existing Ising and QUBO embedding approaches rely on penalty scalarization~\cite{lucas2014ising}, adding squared penalty terms $\rho (\mathbf{a}^\top \mathbf{u} - b)^2$ into the Hamiltonian. As highlighted in recent literature~\cite{mirkarimi2024quantum,pelofske2023comparing}, this introduces severe physical failure modes: (1) \emph{Energy compression}: large penalty coefficients force Hamiltonian normalization that compresses task loss energy differences below the hardware intrinsic noise threshold; (2) \emph{Graph fill-in}: squared affine constraints generate dense quadratic cross-terms, dramatically increasing graph degree and exceeding hardware minor-embedding connectivity.

Our completely positive formulation eliminates penalty scalarization entirely. All affine consistency constraints are maintained deterministically by classical primal--dual and augmented-Lagrangian updates, deploying the physical Ising machine strictly as an unconstrained Boolean Linear Minimization Oracle (LMO)~\cite{yurtsever2022q}. We prove the first certified $\mathcal{O}(T^{-1/2})$ convergence bounds for Quantum Conditional Gradient Descent (QCGD) under stochastic additive hardware noise $\xi_t$ and finite coefficient precision, establishing a noise-resilient bridge between Ising physics and deep learning.

\subsection{Stochastic Decomposition and Consensus Optimization}
Scaling exact conic lifts to large training sets requires decomposing monolithic problem formulations. In classical stochastic programming and operations research, Progressive Hedging (PH)~\cite{rockafellar1991scenarios} provides a foundational framework for decomposing multistage decision problems across scenarios via probability-weighted consensus projections and non-anticipativity multiplier updates. In modern conic optimization, sparse and structured copositive decompositions~\cite{bomze2022two,gabl2023sparse,brown2024copositive} have emerged to decouple large-scale block matrices.

\section{A Logarithmic Binary Encoding for Piecewise-Linear Functions}
\label{sec:log_pwl_supp}

Section~2.2.1 employs a one-hot representation for each of the $K$ segments for conceptual clarity.  This section establishes a self-contained, exact logarithmic alternative that compresses the number of binary segment selection variables from $K$ to $r := \lceil\log_2 K\rceil$, based on the disaggregated convex-combination formulation of the SOS2 graph~\cite{vielma2011modeling}.

\subsection{Logarithmic disaggregated formulation}

Let $z_0 < z_1 < \cdots < z_K$ be the strictly ordered activation breakpoints and let $f_k := \sigma(z_k)$ ($k=0,\dots,K$).  Assign each segment $h \in [K]$ a distinct binary codeword $\mathbf g^h = (g_1^h, \dots, g_r^h) \in \{0,1\}^r$ with $r = \lceil\log_2 K\rceil$ (e.g., reflected binary Gray codes where adjacent segments differ in exactly one bit).

For each segment $h \in [K]$, introduce continuous endpoint weights $\lambda_{h,0}, \lambda_{h,1} \ge 0$ and segment selector $w_h := \lambda_{h,0} + \lambda_{h,1}$.  The disaggregated convex-combination equations are:
\begin{align}
 z = \sum_{h=1}^K \left( z_{h-1}\lambda_{h,0} + z_h\lambda_{h,1} \right),
 \qquad
 a = \sum_{h=1}^K \left( f_{h-1}\lambda_{h,0} + f_h\lambda_{h,1} \right),
 \qquad
 \sum_{h=1}^K w_h = 1,
 \qquad
 \lambda_{h,0}, \lambda_{h,1} \ge 0.
 \label{eq:log-pwl-disaggregated}
\end{align}
Introducing binary branching variables $\mathbf b = (b_1, \dots, b_r) \in \{0,1\}^r$, we enforce the logarithmic linking inequalities for each bit position $j \in [r]$:
\begin{align}
 \sum_{h : g_j^h = 1} w_h \le b_j,
 \qquad
 \sum_{h : g_j^h = 0} w_h \le 1 - b_j.
 \label{eq:gray_branch_supp}
\end{align}

\begin{proposition}[Exactness of Logarithmic Piecewise-Linear Encoding]
\label{prop:log-pwl-exactness}
The projection of the linear-mixed integer system \eqref{eq:log-pwl-disaggregated}--\eqref{eq:gray_branch_supp} onto $(z,a)$ is identically the continuous piecewise-linear graph $\operatorname{graph}(\sigma) = \{(z, \sigma(z)) : z \in [z_0, z_K]\}$.
\end{proposition}

\begin{proof}
Fix any feasible binary assignment $\mathbf b \in \{0,1\}^r$.  If $w_h > 0$ for some segment $h \in [K]$, then \eqref{eq:gray_branch_supp} requires $b_j = 1$ whenever $g_j^h = 1$ and $b_j = 0$ whenever $g_j^h = 0$, implying $\mathbf g^h = \mathbf b$.  Because the codewords $\{\mathbf g^h\}_{h=1}^K$ are pairwise distinct, at most one segment can satisfy $w_h > 0$.  Together with the normalization $\sum_{h=1}^K w_h = 1$, exactly one unique segment $h^* \in [K]$ satisfies $w_{h^*} = 1$ (with $w_h = 0$ for all $h \neq h^*$).  Hence $\lambda_{h^*,0} + \lambda_{h^*,1} = 1$, and $(z,a)$ is an exact convex combination of $(z_{h^*-1}, f_{h^*-1})$ and $(z_{h^*}, f_{h^*})$.  Conversely, any point $(z, \sigma(z))$ lying in segment $h$ is generated by setting $\mathbf b = \mathbf g^h$, assigning $(\lambda_{h,0}, \lambda_{h,1})$ to the barycentric coordinates of $z$ within $[z_{h-1}, z_h]$, and setting all other weights to zero.
\end{proof}

\subsection{Loss interpolation and network coupling}

The identical logarithmic construction applies directly to sample loss functions by substituting $(z_k, f_k)$ with empirical loss-evaluation pairs $(P_k, \mathcal L(P_k, \mathbf y^{(s)}))$, requiring only $\lceil\log_2 q\rceil$ binary selectors for a $q$-piece discretized loss.

To propagate continuous activations $a \in [\underline a, \bar a]$ through binary-encoded quantized weights $W = d + \sum_{r'=1}^R c_{r'} \delta_{r'}$ ($\delta_{r'} \in \{0,1\}$), we define bilinear product variables $v_{r'} := \delta_{r'} a$ bounded by exact McCormick envelopes:
\begin{align}
 \underline a \, \delta_{r'} \le v_{r'} \le \bar a \, \delta_{r'},
 \qquad
 a - \bar a (1 - \delta_{r'}) \le v_{r'} \le a - \underline a (1 - \delta_{r'}).
 \label{eq:log-pwl-binary-product}
\end{align}
Because $\delta_{r'} \in \{0,1\}$, these envelopes are algebraically exact without relaxation gap.  Thus, the activation-weight product $Wa = da + \sum_{r'=1}^R c_{r'} v_{r'}$ is strictly affine, embedding the network into the canonical mixed-binary QCBO framework.

\section{Proof of the Description-Length Excess-Risk Bound}
\label{sec:supp-sample-complexity}

This section provides the complete statistical proofs for the non-asymptotic description-length excess-risk generalization bound (Theorem~\ref{thm:finite-class-sample-complexity}) and its induced logical-spin complexity (Corollary~\ref{cor:statistical-spin-complexity}).

\subsection{Proof of Theorem~\ref{thm:finite-class-sample-complexity}}

\begin{proofof}{Theorem~\ref{thm:finite-class-sample-complexity}}
Let $\mathcal{P}$ denote the data-generating distribution on $\mathcal{X} \times \mathcal{Y}$.  For any fixed parameter configuration $\boldsymbol{\theta} \in \Theta_B$, the loss random variable $C(f_{\boldsymbol{\theta}}(\mathbf{x}), \mathbf{y}) \in [0, \mathcal L_{\max}]$ is bounded almost surely.  By Hoeffding's inequality applied to $N$ independent identically distributed samples $\{(\mathbf{x}^{(s)}, \mathbf{y}^{(s)})\}_{s=1}^N$:
\begin{align}
 \mathbb P\!\left( \left| \widehat{\mathcal R}_N(\boldsymbol{\theta}) - \mathcal R(\boldsymbol{\theta}) \right| > \frac{\varepsilon_{\mathrm{stat}}}{2} \right)
 \le 2\exp\!\left( -\frac{N\varepsilon_{\mathrm{stat}}^2}{2\mathcal L_{\max}^2} \right).
 \label{eq:supp-hoeffding-single}
\end{align}
Because each of the $P$ scalar network parameters is restricted to a $B$-bit discrete codebook, the total hypothesis class has cardinality $|\Theta_B| \le 2^{BP}$.  Applying a union bound across the finite discrete parameter space $\Theta_B$ yields the uniform convergence guarantee:
\begin{align}
 \mathbb P\!\left( \sup_{\boldsymbol{\theta} \in \Theta_B} \left| \widehat{\mathcal R}_N(\boldsymbol{\theta}) - \mathcal R(\boldsymbol{\theta}) \right| > \frac{\varepsilon_{\mathrm{stat}}}{2} \right)
 \le 2|\Theta_B|\exp\!\left( -\frac{N\varepsilon_{\mathrm{stat}}^2}{2\mathcal L_{\max}^2} \right)
 \le 2\exp\!\left( BP\log 2 - \frac{N\varepsilon_{\mathrm{stat}}^2}{2\mathcal L_{\max}^2} \right).
 \label{eq:supp-uniform-finite-class}
\end{align}
Setting the right-hand side of \eqref{eq:supp-uniform-finite-class} to be at most $\delta \in (0,1)$ requires:
\begin{align}
 N \ge \frac{2\mathcal L_{\max}^2}{\varepsilon_{\mathrm{stat}}^2} \left( BP\log 2 + \log\frac{2}{\delta} \right),
 \label{eq:supp-finite-class-sample-bound}
\end{align}
which establishes the sample complexity condition \eqref{eq:finite-class-sample-bound}.

On the event where $\sup_{\boldsymbol{\theta} \in \Theta_B} |\widehat{\mathcal R}_N(\boldsymbol{\theta}) - \mathcal R(\boldsymbol{\theta})| \le \varepsilon_{\mathrm{stat}}/2$ (which holds with probability at least $1-\delta$), the empirical risk minimizer $\widehat{\boldsymbol{\theta}}_N$ with empirical suboptimality $\varepsilon_{\mathrm{opt}}$ satisfies:
\begin{align}
 \mathcal R(\widehat{\boldsymbol{\theta}}_N)
 &\le \widehat{\mathcal R}_N(\widehat{\boldsymbol{\theta}}_N) + \frac{\varepsilon_{\mathrm{stat}}}{2} \notag \\
 &\le \min_{\boldsymbol{\theta} \in \Theta_B} \widehat{\mathcal R}_N(\boldsymbol{\theta}) + \varepsilon_{\mathrm{opt}} + \frac{\varepsilon_{\mathrm{stat}}}{2} \notag \\
 &\le \widehat{\mathcal R}_N(\boldsymbol{\theta}_B^\star) + \varepsilon_{\mathrm{opt}} + \frac{\varepsilon_{\mathrm{stat}}}{2} \notag \\
 &\le \mathcal R(\boldsymbol{\theta}_B^\star) + \frac{\varepsilon_{\mathrm{stat}}}{2} + \varepsilon_{\mathrm{opt}} + \frac{\varepsilon_{\mathrm{stat}}}{2}
 = \mathcal R(\boldsymbol{\theta}_B^\star) + \varepsilon_{\mathrm{stat}} + \varepsilon_{\mathrm{opt}},
\end{align}
which proves the excess-risk bound \eqref{eq:stat-opt-decomposition}.
\end{proofof}

\subsection{Derivation of Corollary~\ref{cor:statistical-spin-complexity}}

\begin{proofof}{Corollary~\ref{cor:statistical-spin-complexity}}
By definition of the representation error $\varepsilon_{\mathrm{repr}}(B,n) := \mathcal R(\boldsymbol{\theta}_B^\star) - \mathcal R_{\mathrm{ref}}^\star$, we decompose the excess population risk relative to the continuous reference class $\mathcal F_{\mathrm{ref}}$ as:
\begin{align}
 \mathcal R(\widehat{\boldsymbol{\theta}}_N) - \mathcal R_{\mathrm{ref}}^\star
 = \big[ \mathcal R(\widehat{\boldsymbol{\theta}}_N) - \mathcal R(\boldsymbol{\theta}_B^\star) \big] + \big[ \mathcal R(\boldsymbol{\theta}_B^\star) - \mathcal R_{\mathrm{ref}}^\star \big].
\end{align}
Under condition \eqref{eq:supp-finite-class-sample-bound}, Theorem~\ref{thm:finite-class-sample-complexity} bounds the first bracket by $\varepsilon_{\mathrm{stat}} + \varepsilon_{\mathrm{opt}}$ with probability at least $1-\delta$, immediately establishing \eqref{eq:end-to-end-risk-decomposition}.

A sufficient integer sample size guaranteeing this excess-risk bound is:
\begin{align}
 N_{\mathrm{stat}} := \left\lceil \frac{2\mathcal L_{\max}^2}{\varepsilon_{\mathrm{stat}}^2} \left( BP\log 2 + \log\frac{2}{\delta} \right) \right\rceil.
\end{align}
Substituting $N_{\mathrm{stat}}$ into the monolithic logical-spin footprint $S_{\mathrm{logical}}(N) = \mathcal O(BP + N(nU + BP))$ (Table~\ref{tab:logical-spin-count}) yields:
\begin{align}
 S_{\mathrm{logical}}
 = \mathcal O\!\left( BP + (nU + BP)\left[ 1 + \frac{\mathcal L_{\max}^2}{\varepsilon_{\mathrm{stat}}^2}\left(BP + \log\frac{1}{\delta}\right) \right] \right),
\end{align}
establishing \eqref{eq:statistical-spin-composition}.
\end{proofof}

\section{Proof of the Manifold Structure}
\label{app:proof}

This section establishes the stratified manifold structure of the zero-loss level set $\mathcal{M}_0$ (Theorem~\ref{thm:global_manifold}) and proves constructive strict interpolation for nonpolynomial piecewise-polynomial networks (Proposition~\ref{prop:strict_interpolation}).

\subsection{Whitney semialgebraic stratification of parameter space}

Let $\mathcal B_\sigma = \{\beta_1, \dots, \beta_r\}$ denote the finite set of activation breakpoints.  For any training sample $s \in [N]$, layer $\ell \in [L-1]$, and neuron $j \in [m_\ell]$, the pre-activation $z_{s,j}^{(\ell)}(\boldsymbol{\theta})$ is a multivariate polynomial on each region where previous layer activations are smooth.

Let each activation interval and breakpoint define a polynomial equality or inequality.  Because the network architecture and activation pieces are polynomial, every sign configuration defines a semialgebraic subset of the parameter space $\mathbb{R}^D$.  By the finite semialgebraic stratification theorem~\cite{bochnak1998real}, $\mathbb{R}^D$ admits a finite disjoint partition into connected $C^\infty$ semialgebraic submanifolds (strata):
\begin{equation}
 \mathbb{R}^D = \bigsqcup_{\alpha \in A} S_\alpha,
\end{equation}
such that on each stratum $S_\alpha$, the network evaluation map $\Phi_\alpha := \Phi|_{S_\alpha} : S_\alpha \to \mathbb{R}^{Nm}$ is smooth ($C^\infty$).

We partition the index set $A = A_{\mathrm{int}} \sqcup A_{\mathrm{bd}}$ into interior strata $A_{\mathrm{int}}$ (where no pre-activation hits a breakpoint) and boundary strata $A_{\mathrm{bd}}$ (where at least one pre-activation hits a breakpoint).
\begin{itemize}
 \item \emph{Interior Strata ($A_{\mathrm{int}}$)}: The interior activation-pattern regions are open in $\mathbb{R}^D$, so each interior stratum $S_\alpha$ ($\alpha \in A_{\mathrm{int}}$) has maximal manifold dimension $\dim S_\alpha = D$.
 \item \emph{Boundary Strata ($A_{\mathrm{bd}}$)}: The boundary is contained in the finite union of hypersurfaces:
 \begin{equation}
  z_{s,j}^{(\ell)}(\boldsymbol{\theta}) - \beta = 0 \qquad (\beta \in \mathcal B_\sigma).
 \end{equation}
 Because the hidden bias $b_j^{(\ell)}$ enters $z_{s,j}^{(\ell)}(\boldsymbol{\theta})$ with coefficient one, the gradient with respect to parameter coordinates never vanishes identically.  Hence, each boundary stratum $S_\alpha$ ($\alpha \in A_{\mathrm{bd}}$) is a semialgebraic submanifold of dimension $\dim S_\alpha \le D - 1$.
\end{itemize}

\subsection{Proof of Theorem~\ref{thm:global_manifold}}

\begin{proofof}{Theorem~\ref{thm:global_manifold}}
Let $q := Nm$ denote the dimension of the label target space.  For each stratum $S_\alpha$, define the set of critical values:
\begin{equation}
 C_\alpha := \left\{ \Phi_\alpha(\boldsymbol{\theta}) \in \mathbb{R}^q : \boldsymbol{\theta} \in S_\alpha, \; \operatorname{rank} D\Phi_\alpha(\boldsymbol{\theta}) < q \right\}.
\end{equation}
\begin{enumerate}
 \item If $\dim S_\alpha < q$, every point in $S_\alpha$ is critical, and Sard's theorem on smooth semialgebraic manifolds~\cite{lee2013introduction} implies that $\Phi_\alpha(S_\alpha)$ has $q$-dimensional Lebesgue measure zero.
 \item If $\dim S_\alpha \ge q$, Sard's theorem implies that the critical value set $C_\alpha$ has $q$-dimensional Lebesgue measure zero.
\end{enumerate}
Because the index set $A$ is finite, the exceptional target set:
\begin{equation}
 E := \bigcup_{\alpha \in A} C_\alpha \subset \mathbb{R}^{Nm}
\end{equation}
has Lebesgue measure zero ($\operatorname{Leb}_{Nm}(E) = 0$).

Fix any generic label target $\mathbf{Y} \notin E$.
\begin{enumerate}
 \item \emph{Top-Dimensional Interior Components}: For any $\alpha \in A_{\mathrm{int}}$, $\mathbf{Y}$ is a regular value of $\Phi_\alpha$.  By the regular level-set theorem, if the fiber $\Phi_\alpha^{-1}(\mathbf{Y})$ is non-empty, it is a smooth $C^\infty$ embedded submanifold of dimension:
 \begin{equation}
  \dim \Phi_\alpha^{-1}(\mathbf{Y}) = \dim S_\alpha - q = D - Nm.
 \end{equation}
 \item \emph{Boundary and Lower-Dimensional Strata}: For any $\alpha \in A_{\mathrm{bd}}$, the fiber $\Phi_\alpha^{-1}(\mathbf{Y})$ has manifold dimension:
 \begin{equation}
  \dim \Phi_\alpha^{-1}(\mathbf{Y}) \le \dim S_\alpha - q \le (D - 1) - Nm = D - Nm - 1.
 \end{equation}
 A finite union of smooth submanifolds of dimension at most $D - Nm - 1$ has Hausdorff dimension at most $D - Nm - 1$, and consequently has $(D - Nm)$-dimensional Hausdorff measure zero:
 \begin{equation}
  \mathcal{H}^{D-Nm}\left( \Phi^{-1}(\mathbf{Y}) \cap \bigcup_{\alpha \in A_{\mathrm{bd}}} S_\alpha \right) = 0.
 \end{equation}
\end{enumerate}
This completes the proof of Theorem~\ref{thm:global_manifold}.
\end{proofof}

\subsection{Constructive strict interpolation for nonpolynomial piecewise-polynomial activations}
\label{app:strict_interpolation}

We prove that the zero-loss level set $\mathcal{M}_0(\mathbf{Y})$ is strictly non-empty under the conditions of Proposition~\ref{prop:strict_interpolation}.

\begin{lemma}[One-Dimensional Strict Embedding]
\label{lem:embedding}
Let $\mathbf{x}^{(1)}, \dots, \mathbf{x}^{(N)} \in \mathbb{R}^{m_0}$ be pairwise distinct training inputs.  Let $\sigma : \mathbb{R} \to \mathbb{R}$ be a continuous, piecewise-polynomial activation with finite breakpoint set $\mathcal B_\sigma$ that does not agree almost everywhere with a polynomial.  If $L \ge 3$ and $m_\ell \ge 1$ for all $1 \le \ell \le L-2$, there exist parameters in the first $L-2$ hidden layers such that the scalar outputs:
\begin{equation}
 h_{1,1}^{(L-2)}, \dots, h_{N,1}^{(L-2)} \quad \text{are pairwise distinct},
\end{equation}
and every training pre-activation in these layers lies strictly outside the breakpoint set $\mathcal B_\sigma$.
\end{lemma}

\begin{proofof}{Lemma~\ref{lem:embedding}}
For any pair $s \neq t$, the difference vector $\mathbf{x}^{(s)} - \mathbf{x}^{(t)} \neq \mathbf{0}$ defines a proper affine hyperplane $\{ \mathbf{w} : \mathbf{w}^\top (\mathbf{x}^{(s)} - \mathbf{x}^{(t)}) = 0 \}$.  Because a finite union of proper hyperplanes cannot cover $\mathbb{R}^{m_0}$, there exists $\mathbf{w} \in \mathbb{R}^{m_0}$ such that the scalar projections $v_s := \mathbf{w}^\top \mathbf{x}^{(s)}$ are pairwise distinct.

Because $\sigma$ is non-polynomial and continuous, there exists an open interval $I \subset \mathbb{R} \setminus \mathcal B_\sigma$ on which $\sigma$ is strictly monotone.  Choosing $\lambda_1 \neq 0$ sufficiently small and $\mu_1 \in \mathbb{R}$ such that $\mu_1 + \lambda_1 v_s \in I$ for all $s \in [N]$, the outputs $\sigma(\mu_1 + \lambda_1 v_s)$ remain strictly pairwise distinct.  Setting the weights of all unused neurons to zero and biases to $\rho \in \mathbb{R} \setminus \mathcal B_\sigma$, induction across layers $\ell = 2, \dots, L-2$ guarantees pairwise distinctness and strict breakpoint avoidance.
\end{proofof}

\begin{lemma}[Full-Rank Nonpolynomial Activation Features]
\label{lem:full-rank-features}
Let $u_1, \dots, u_N \in \mathbb{R}$ be pairwise distinct, and let $\sigma : \mathbb{R} \to \mathbb{R}$ be a continuous, nonpolynomial piecewise-polynomial activation.  There exist affine parameters $(a_j, b_j)_{j=1}^{N-1}$ such that $a_j u_s + b_j \notin \mathcal B_\sigma$ for all $s \in [N], j \in [N-1]$, and the feature matrix:
\begin{equation}
 \mathbf{A} := \left[ \mathbf{1}_N \;\middle|\; \left(\sigma(a_j u_s + b_j)\right)_{s \in [N], \, j \in [N-1]} \right] \in \mathbb{R}^{N \times N}
\end{equation}
is non-singular ($\det \mathbf{A} \neq 0$).
\end{lemma}

\begin{proofof}{Lemma~\ref{lem:full-rank-features}}
Let $K \subset \mathbb{R}$ be a compact interval containing $\{u_1, \dots, u_N\}$.  By the Leshno et al.~\cite{leshno1993multilayer} density theorem, $\operatorname{span}\{ u \mapsto \sigma(au+b) : a,b \in \mathbb{R} \}$ is dense in $C(K)$.  Because the evaluation map $T : C(K) \to \mathbb{R}^N$ given by $T(f) = (f(u_1), \dots, f(u_N))^\top$ is surjective, the finite-dimensional span $\operatorname{span}\{ (\sigma(au_1+b), \dots, \sigma(au_N+b))^\top : a,b \in \mathbb{R} \} = \mathbb{R}^N$ is the full space.  Thus, the constant vector $\mathbf{1}_N$ can be extended by $N-1$ ridge directions to form an invertible matrix $\mathbf{A}$.

Because $\det \mathbf{A}$ is a continuous function of the parameters $(a_j, b_j)$, it remains non-zero on an open neighborhood.  The forbidden breakpoint conditions $a_j u_s + b_j = \beta$ ($\beta \in \mathcal B_\sigma$) form a finite union of affine hyperplanes with dense complement.  Perturbing $(a_j, b_j)$ within the non-singular neighborhood yields non-singularity while avoiding all breakpoints.
\end{proofof}

\begin{proofof}{Proposition~\ref{prop:strict_interpolation}}
For $L \ge 3$, apply Lemma~\ref{lem:embedding} to obtain pairwise distinct penultimate scalar features $u_s = h_{s,1}^{(L-2)}$; for $L=2$, apply the separating projection directly.  Applying Lemma~\ref{lem:full-rank-features} produces $N-1$ hidden neurons whose outputs, combined with the output bias $\mathbf{1}_N$, form an augmented feature matrix $\widetilde{\mathbf{H}}^{(L-1)} \in \mathbb{R}^{N \times (m_{L-1}+1)}$ of full row rank $N$.

The output-layer equation $\widetilde{\mathbf{H}}^{(L-1)} \mathbf{C} = \mathbf{Y}$ has exact solution $\mathbf{C} = (\widetilde{\mathbf{H}}^{(L-1)})^\top (\widetilde{\mathbf{H}}^{(L-1)} (\widetilde{\mathbf{H}}^{(L-1)})^\top)^{-1} \mathbf{Y}$ for any target matrix $\mathbf{Y} \in \mathbb{R}^{N \times m}$.  Because all pre-activations strictly avoid $\mathcal B_\sigma$, the constructed parameter configuration lies in the interior stratum, proving $\mathcal{M}_0(\mathbf{Y}) \neq \emptyset$.
\end{proofof}

\subsection{Topological disconnectedness of the zero-loss set (Counterexample)}

\begin{proposition}[Disconnectedness of $\mathcal{M}_0$]
There exists an overparameterized shallow ReLU network and training dataset for which the zero-loss level set $\mathcal{M}_0$ is topologically disconnected.
\end{proposition}

\begin{proof}
Consider a two-layer ReLU network with 1D input/output and $m_1 = 2$ hidden neurons: $f(x; \boldsymbol{\theta}) = w_2 \sigma(w_1 x) + v_2 \sigma(v_1 x)$ with parameter vector $\boldsymbol{\theta} = (w_1, w_2, v_1, v_2) \in \mathbb{R}^4$ and $\sigma(z) = \max(0, z)$.  Let the training dataset be $\mathcal{D} = \{(1, 1), (-1, 1)\}$, which satisfies overparameterization ($D = 4 > Nm = 2$).  The zero-loss system is:
\begin{align}
 w_2 \sigma(w_1) + v_2 \sigma(v_1) &= 1, \label{eq:supp-counter-1} \\
 w_2 \sigma(-w_1) + v_2 \sigma(-v_1) &= 1. \label{eq:supp-counter-2}
\end{align}
If $w_1 = 0$, then \eqref{eq:supp-counter-1}--\eqref{eq:supp-counter-2} requires $v_2 \sigma(v_1) = v_2 \sigma(-v_1) = 1$, which is impossible since $\sigma(v_1)\sigma(-v_1) = 0$ for all $v_1 \in \mathbb{R}$.  Thus, $\mathcal{M}_0 \cap \{ \boldsymbol{\theta} : w_1 = 0 \} = \emptyset$.  By symmetry, $\mathcal{M}_0 \cap \{ \boldsymbol{\theta} : v_1 = 0 \} = \emptyset$.

Solving the system in the remaining open quadrants:
\begin{enumerate}
 \item If $w_1 > 0, v_1 > 0$ or $w_1 < 0, v_1 < 0$: on each sample, at least one neuron has zero activation, yielding $0 = 1$ (no solution).
 \item If $w_1 > 0, v_1 < 0$: the equations reduce to $w_2 w_1 = 1$ and $-v_2 v_1 = 1$, defining the smooth manifold component:
 \begin{equation}
  \mathcal{C}_1 = \left\{ \left(w_1, \frac{1}{w_1}, v_1, -\frac{1}{v_1}\right) \in \mathbb{R}^4 \;\middle|\; w_1 > 0, \, v_1 < 0 \right\}.
 \end{equation}
 \item If $w_1 < 0, v_1 > 0$: the equations reduce to $-w_2 w_1 = 1$ and $v_2 v_1 = 1$, defining the smooth manifold component:
 \begin{equation}
  \mathcal{C}_2 = \left\{ \left(w_1, -\frac{1}{w_1}, v_1, \frac{1}{v_1}\right) \in \mathbb{R}^4 \;\middle|\; w_1 < 0, \, v_1 > 0 \right\}.
 \end{equation}
\end{enumerate}
The global zero-loss set is $\mathcal{M}_0 = \mathcal{C}_1 \cup \mathcal{C}_2$.  Because $\mathcal{C}_1 \subset \{ \boldsymbol{\theta} : w_1 > 0 \}$ and $\mathcal{C}_2 \subset \{ \boldsymbol{\theta} : w_1 < 0 \}$ are separated by the empty-solution hyperplane $\{w_1 = 0\}$, $\mathcal{M}_0$ consists of two disjoint non-empty closed connected components, proving topological disconnectedness.
\end{proof}

\section{Details of Convex Formulation}
\label{sec:details_convex_formulation}

This section establishes the explicit, bounded mixed-binary quadratically constrained formulation underlying the exact completely positive reformulation in Theorem~\ref{thm:exact_cpp_equivalence}, and details its extension to arbitrary-degree piecewise-polynomial activations and continuous parameter spaces.

\subsection{Bounded mixed-binary quadratic normal form via FIP}
\label{subsec:supp-fip-formulation}

\paragraph{Problem setting and index conventions.}
Consider the empirical risk minimization of an $L$-layer feedforward neural network on training dataset $\{(\mathbf{x}^{(s)}, \mathbf{y}^{(s)})\}_{s=1}^N \subset \mathcal{X} \times \mathcal{Y}$.  Let $\ell \in [L]$ index layers, with widths $m_\ell$ ($m_0 = d_{\mathrm{in}}$, $m_L = m$).  For sample $s \in [N]$, let $\mathbf{a}^{(s,0)} = \mathbf{x}^{(s)}$, and let $\mathbf{z}^{(s,\ell)}, \mathbf{a}^{(s,\ell)} \in \mathbb{R}^{m_\ell}$ denote pre- and post-activation vectors.

Each quantized weight coordinate $W_{jk}^{(\ell)}$ ($j \in [m_\ell], k \in [m_{\ell-1}]$) and bias coordinate $b_j^{(\ell)}$ ($j \in [m_\ell]$) is parameterized via an affine Boolean codebook:
\begin{align}
 W_{jk}^{(\ell)} = d_{W,jk}^{(\ell)} + \sum_{r=1}^{R_\ell} c_{jk,r}^{(\ell)} \delta_{jk,r}^{(\ell)}, \qquad \delta_{jk,r}^{(\ell)} \in \{0,1\},
 \label{eq:supp-weight-code}
\end{align}
with base offsets $d_{W,jk}^{(\ell)} \in \mathbb{R}$, scale coefficients $c_{jk,r}^{(\ell)} \in \mathbb{R}$, and code bits $\delta_{jk,r}^{(\ell)}$.  Biases are parameterized analogously as $\mathbf{b}^{(\ell)} = \mathbf{d}_b^{(\ell)} + \mathbf{C}_b^{(\ell)} \boldsymbol{\delta}_b^{(\ell)}$ with $\boldsymbol{\delta}_b^{(\ell)} \in \{0,1\}^{m_\ell R_{b,\ell}}$.  Forward Interval Propagation (FIP) over the compact input domain $\mathcal{X}$ supplies deterministic, valid hyper-rectangular bounds for all activations:
\begin{align}
 \underline{a}_k^{(s,\ell-1)} \le a_k^{(s,\ell-1)} \le \overline{a}_k^{(s,\ell-1)}, \qquad \forall s \in [N], \, \ell \in [L], \, k \in [m_{\ell-1}].
 \label{eq:supp-fip-bounds}
\end{align}

\paragraph{Exact bilinear linearization via McCormick envelopes.}
The forward pass couples discrete bits $\delta_{jk,r}^{(\ell)}$ and continuous activations $a_k^{(s,\ell-1)}$.  Let $q_\ell := m_\ell m_{\ell-1} R_\ell$ denote the layer bit dimension, with bijective lexicographic mapping $\pi_\ell(j,k,r) := ((j-1)m_{\ell-1} + k - 1) R_\ell + r$.  The concatenated weight code $\boldsymbol{\delta}_W^{(\ell)} \in \{0,1\}^{q_\ell}$ and auxiliary product vector $\mathbf{v}^{(s,\ell)} \in \mathbb{R}^{q_\ell}$ follow this ordering, satisfying the bilinear identity $v_{jk,r}^{(s,\ell)} = \delta_{jk,r}^{(\ell)} a_k^{(s,\ell-1)}$.

Define the replication matrix $\mathbf{P}^{(\ell)} := \mathbf{1}_{m_\ell} \otimes (\mathbf{I}_{m_{\ell-1}} \otimes \mathbf{1}_{R_\ell}) \in \mathbb{R}^{q_\ell \times m_{\ell-1}}$ and diagonal bound matrices:
\begin{align}
 \underline{\mathbf{A}}^{(s,\ell-1)} := \mathbf{I}_{m_\ell} \otimes \operatorname{Diag}(\underline{\mathbf{a}}^{(s,\ell-1)}) \otimes \mathbf{I}_{R_\ell}, \qquad
 \overline{\mathbf{A}}^{(s,\ell-1)} := \mathbf{I}_{m_\ell} \otimes \operatorname{Diag}(\overline{\mathbf{a}}^{(s,\ell-1)}) \otimes \mathbf{I}_{R_\ell}.
\end{align}
The complete system of McCormick inequalities across all coordinates is expressed as:
\begin{align}
 \mathbf{N}_{\mathrm{Mc}}^{(s,\ell)}
 \begin{bmatrix} \boldsymbol{\delta}_W^{(\ell)} \\ \mathbf{a}^{(s,\ell-1)} \\ \mathbf{v}^{(s,\ell)} \end{bmatrix}
 \le \mathbf{n}_{\mathrm{Mc}}^{(s,\ell)},
 \qquad
 \mathbf{N}_{\mathrm{Mc}}^{(s,\ell)} := \begin{bmatrix}
  -\overline{\mathbf{A}}^{(s,\ell-1)} & \mathbf{0} & \mathbf{I} \\
   \underline{\mathbf{A}}^{(s,\ell-1)} & \mathbf{0} & -\mathbf{I} \\
  -\underline{\mathbf{A}}^{(s,\ell-1)} & -\mathbf{P}^{(\ell)} & \mathbf{I} \\
   \overline{\mathbf{A}}^{(s,\ell-1)}  &  \mathbf{P}^{(\ell)} & -\mathbf{I}
 \end{bmatrix},
 \qquad
 \mathbf{n}_{\mathrm{Mc}}^{(s,\ell)} := \begin{bmatrix}
  \mathbf{0} \\
  \mathbf{0} \\
  -\mathbf{P}^{(\ell)} \underline{\mathbf{a}}^{(s,\ell-1)} \\
   \mathbf{P}^{(\ell)} \overline{\mathbf{a}}^{(s,\ell-1)}
 \end{bmatrix}.
 \label{eq:supp-mcc-matrix}
\end{align}

\begin{lemma}[Exactness of Polyhedral McCormick Linearization]
\label{lem:supp-mcc-exact}
Under valid FIP bounds \eqref{eq:supp-fip-bounds} and binary integrality $\boldsymbol{\delta}_W^{(\ell)} \in \{0,1\}^{q_\ell}$, the polyhedral system \eqref{eq:supp-mcc-matrix} holds if and only if $v_{jk,r}^{(s,\ell)} = \delta_{jk,r}^{(\ell)} a_k^{(s,\ell-1)}$ holds for all $(j,k,r)$.
\end{lemma}

\begin{proof}
For any scalar coordinate with $\delta \in \{0,1\}$ and $a \in [\underline{a}, \overline{a}]$, system \eqref{eq:supp-mcc-matrix} expands to:
\begin{align}
 \underline{a} \delta \le v \le \overline{a} \delta, \qquad a - \overline{a}(1 - \delta) \le v \le a - \underline{a}(1 - \delta).
 \label{eq:scalar-mcc}
\end{align}
If $\delta = 0$, the first pair forces $0 \le v \le 0 \implies v = 0$, while the second pair becomes $a - \overline{a} \le 0 \le a - \underline{a}$, which holds identically by \eqref{eq:supp-fip-bounds}.  If $\delta = 1$, the second pair forces $a \le v \le a \implies v = a$, while the first pair reduces to $\underline{a} \le a \le \overline{a}$.  Conversely, substituting $v = \delta a$ satisfies all four inequalities.
\end{proof}

\paragraph{Affine forward propagation.}
Let $p_\ell := m_\ell m_{\ell-1}$ and $\pi'_\ell(j,k) := (j-1)m_{\ell-1} + k$.  Define the bit routing matrix $\mathbf{C}_{\mathrm{bit}}^{(\ell)} \in \mathbb{R}^{p_\ell \times q_\ell}$ by $[\mathbf{C}_{\mathrm{bit}}^{(\ell)}]_{\pi'_\ell(j,k), \, \pi_\ell(j',k',r)} := c_{jk,r}^{(\ell)}$ if $(j',k') = (j,k)$ and $0$ otherwise.  With offset vector $\mathbf{d}_W^{(\ell)} \in \mathbb{R}^{p_\ell}$, intermediate interaction vector $\mathbf{u}_{\mathrm{bil}}^{(s,\ell)} \in \mathbb{R}^{p_\ell}$, and summation operator $\mathbf{S}_{\mathrm{sum}}^{(\ell)} := \mathbf{I}_{m_\ell} \otimes \mathbf{1}_{m_{\ell-1}}^\top$, forward propagation is affine:
\begin{align}
 \mathbf{u}_{\mathrm{bil}}^{(s,\ell)} &= \mathbf{C}_{\mathrm{bit}}^{(\ell)} \mathbf{v}^{(s,\ell)} + (\mathbf{I}_{m_\ell} \otimes \operatorname{Diag}(\mathbf{d}_{W,j\cdot}^{(\ell)})) \mathbf{a}^{(s,\ell-1)}, \label{eq:reconstruct-bilinear} \\
 \mathbf{z}^{(s,\ell)} &= \mathbf{S}_{\mathrm{sum}}^{(\ell)} \mathbf{u}_{\mathrm{bil}}^{(s,\ell)} + \mathbf{C}_b^{(\ell)} \boldsymbol{\delta}_b^{(\ell)} + \mathbf{d}_b^{(\ell)}. \label{eq:reconstruct}
\end{align}

\paragraph{Piecewise-linear activations and empirical loss.}
Let $M_0 < M_1 < \dots < M_n$ partition the operating range of activation $\sigma$.  For sample $s$, layer $\ell$, and neuron $j$, introduce one-hot selector $\boldsymbol{\beta}_{z,j}^{(s,\ell)} \in \{0,1\}^n$ and local coordinates $\boldsymbol{\theta}_{z,j}^{(s,\ell)} \in \mathbb{R}_+^n$:
\begin{align}
 \mathbf{1}^\top \boldsymbol{\beta}_{z,j}^{(s,\ell)} = 1, \qquad \mathbf{0} \le \boldsymbol{\theta}_{z,j}^{(s,\ell)} \le \boldsymbol{\beta}_{z,j}^{(s,\ell)}, \\
 z_j^{(s,\ell)} = \mathbf{M}_{\mathrm{L}} \boldsymbol{\beta}_{z,j}^{(s,\ell)} + \Delta\mathbf{M} \boldsymbol{\theta}_{z,j}^{(s,\ell)}, \qquad
 a_j^{(s,\ell)} = \boldsymbol{\sigma}_{\mathrm{L}} \boldsymbol{\beta}_{z,j}^{(s,\ell)} + \Delta\boldsymbol{\sigma} \boldsymbol{\theta}_{z,j}^{(s,\ell)},
\end{align}
where $\mathbf{M}_{\mathrm{L}} := [M_0, \dots, M_{n-1}]$, $\Delta\mathbf{M} := [M_1-M_0, \dots, M_n-M_{n-1}]$, $\boldsymbol{\sigma}_{\mathrm{L}} := [\sigma(M_0), \dots, \sigma(M_{n-1})]$, and $\Delta\boldsymbol{\sigma} := [\sigma(M_1)-\sigma(M_0), \dots, \sigma(M_n)-\sigma(M_{n-1})]$.

The terminal loss is discretized analogously: over grid $P_0 < P_1 < \dots < P_q$, introducing $\boldsymbol{\beta}_a^{(s)} \in \{0,1\}^q$ and $\boldsymbol{\theta}_a^{(s)} \in \mathbb{R}_+^q$, the empirical loss becomes the linear objective:
\begin{align}
 \widehat{\mathcal L}_{\mathrm{emp}} = \sum_{s=1}^N \sum_{i=1}^q \left( L_{i,-}^{(s)} \beta_{a,i}^{(s)} + \Delta L_i^{(s)} \theta_{a,i}^{(s)} \right).
 \label{eq:supp-pwl-loss}
\end{align}

\paragraph{Canonical Burer mixed-binary normal form.}
Concatenating all decision variables into vector $\boldsymbol{\xi} \in \mathbb{R}^{d_{\mathrm{prim}}}$, shifting continuous coordinates to $\mathbb{R}_+$, adding non-negative upper-bound complements $\bar{u}$, and equalizing all linear inequalities with non-negative slacks $\mathbf{s} \ge \mathbf{0}$ yields the composite vector $\mathbf{u} := [\widetilde{\boldsymbol{\xi}}^\top, \bar{\mathbf{u}}^\top, \mathbf{s}^\top]^\top \in \mathbb{R}_+^p$.  The optimization problem is cast into Burer's canonical form:
\begin{align}
 (P): \quad \min_{\mathbf{u} \ge \mathbf{0}} \quad \mathbf{u}^\top \mathbf{Q} \mathbf{u} + 2\mathbf{c}^\top \mathbf{u} \qquad \text{s.t.} \quad \mathbf{A}\mathbf{u} = \mathbf{b}, \quad u_i \in \{0,1\} \; (i \in \mathcal{B}),
 \label{eq:supp-burer-normal-form}
\end{align}
with compact continuous relaxation $\mathcal{L} := \{\mathbf{u} \in \mathbb{R}_+^p : \mathbf{A}\mathbf{u} = \mathbf{b}\}$.

\subsection{Completely positive reformulation (Theorem~\ref{thm:exact_cpp_equivalence})}
\label{sec:exact_cpp_proof}

Let $\mathbf{a}_q^\top$ denote row $q$ of $\mathbf{A}$, with right-hand side $b_q$.  Define the lifted matrix and cost matrix:
\begin{align}
 \mathbf{X} := \begin{bmatrix} \boldsymbol{\Lambda} & \mathbf{u} \\ \mathbf{u}^\top & 1 \end{bmatrix}, \qquad
 \widehat{\mathbf{Q}} := \begin{bmatrix} \mathbf{Q} & \mathbf{c} \\ \mathbf{c}^\top & 0 \end{bmatrix}.
 \label{eq:supp-cp-lift}
\end{align}
The completely positive program (CPP) is:
\begin{align}
 \min_{\mathbf{X}} \quad \langle \widehat{\mathbf{Q}}, \mathbf{X} \rangle \qquad \text{s.t.} \quad \mathbf{a}_q^\top \mathbf{u} = b_q, \; \mathbf{a}_q^\top \boldsymbol{\Lambda} \mathbf{a}_q = b_q^2 \; (q \in [m]), \quad \Lambda_{ii} = u_i \; (i \in \mathcal{B}), \quad \mathbf{X} \in \mathcal{C}_{p+1}^*,
 \label{eq:supp-burer-cpp}
\end{align}
where $\mathcal{C}_{p+1}^* := \operatorname{cone}\{ \mathbf{w}\mathbf{w}^\top : \mathbf{w} \in \mathbb{R}_+^{p+1} \}$ is the completely positive cone.

Because $(P)$ satisfies non-emptiness, compactness of $\mathcal{L}$, and binary containment ($0 \le u_i \le 1$), Burer's Theorem~\cite{burer2009copositive} establishes:
\begin{equation}
 \mathcal{F}_{\mathrm{CPP}} = \operatorname{conv}\left\{ \begin{bmatrix} \mathbf{u} \\ 1 \end{bmatrix}\begin{bmatrix} \mathbf{u} \\ 1 \end{bmatrix}^\top : \mathbf{u} \in \mathcal{F}_{\mathrm{QCBO}} \right\}.
\end{equation}
Linearity of the objective over $\operatorname{conv}(\dots)$ guarantees $p^*_{\mathrm{CPP}} = p^*_{\mathrm{QCBO}}$, every extreme point of $\operatorname{Argmin}(\mathcal{P}_{\mathrm{CPP}})$ is an exact rank-one atomic lift $[\mathbf{u}^\star; 1][\mathbf{u}^\star; 1]^\top$, and projecting the CPP optimal face onto $\mathbf{u}$ exactly recovers $\operatorname{conv}(\operatorname{Argmin}(\mathcal{P}_{\mathrm{QCBO}}))$.

\subsection{Piecewise-polynomial activations and continuous parameter domains}
\label{subsec:supp-general-piecewise-polynomial}

\paragraph{Piecewise-polynomial activations of degree $d \ge 1$.}
Let activation $\sigma$ be piecewise-polynomial of degree $d$ on segments $z \in [M_{i-1}, M_i]$:
\begin{align}
 a = \sum_{i=1}^n \beta_i p_i(z) = \sum_{i=1}^n \beta_i \left( \sum_{r=0}^d \gamma_{i,r} z^r \right), \qquad \boldsymbol{\beta} \in \{0,1\}^n, \quad \sum_{i=1}^n \beta_i = 1.
\end{align}
We reduce high-degree monomials $z^r$ ($r \ge 3$) to quadratic equality constraints via auxiliary variables $y_2, \dots, y_d$:
\begin{align}
 y_2 = z^2, \qquad y_3 = y_2 z, \qquad \dots, \qquad y_d = y_{d-1} z.
\end{align}
FIP provides valid compact intervals $y_r \in [\underline{y}_r, \overline{y}_r]$.  Binary-continuous products $w_{i,z} := \beta_i z$ and $w_{i,r} := \beta_i y_r$ are linearized exactly via McCormick envelopes.  Coupled with the quadratic equalities $y_r - y_{r-1}z = 0$, the network is cast into a mixed-binary QCQP on a compact domain, which lifts to an exact linear program over $\mathcal{C}_{p+1}^*$.

\paragraph{Continuous parameter domains via generalized copositive cones.}
When parameters $\mathbf{W}^{(\ell)}$ and $\mathbf{b}^{(\ell)}$ are continuous (floating-point), FIP confines all variables to a compact polyhedral domain $\mathcal{K} \subset \mathbb{R}^p$.  The training problem is a continuous QCQP:
\begin{align}
 \min_{\mathbf{y} \in \mathcal{K}} \quad \mathbf{y}^\top \mathbf{Q}_0 \mathbf{y} + 2\mathbf{c}_0^\top \mathbf{y} \qquad \text{s.t.} \quad \mathbf{y}^\top \mathbf{A}_k \mathbf{y} + 2\mathbf{b}_k^\top \mathbf{y} = c_k \quad (k \in [m]).
\end{align}
By Burer \& Dong~\cite{burer2012representing}, this lifts to an exact linear program over the generalized copositive cone:
\begin{align}
 \mathcal{C}^*(\mathcal{K}) := \operatorname{cl}\operatorname{cone}\left\{ \begin{bmatrix} \mathbf{y} \\ 1 \end{bmatrix} \begin{bmatrix} \mathbf{y} \\ 1 \end{bmatrix}^\top : \mathbf{y} \in \mathcal{K} \right\},
\end{align}
with exact equivalence ($p^*_{\mathrm{GCPP}} = p^*_{\mathrm{QCQP}}$), establishing the universality of the conic representation across discrete and continuous regimes.

\section{QCGD with Random Inexact QUBO Oracles}
\label{theoremproof}

This section provides the complete oracle model, algorithmic specification,
and convergence proof for QCGD under stochastic hardware noise and finite
coefficient precision.  Theorem~\ref{prop:supp-random-oracle-full} gives the
full statement underlying Proposition~\ref{convergence_theorem} in the main
text.

\subsection{Standard Primal--Dual QCGD Iteration Algorithm}
\label{subsec:supp-standard-qcgd-algo}

For self-contained reference, we formalize the step-by-step conditional-gradient augmented-Lagrangian (CGAL) algorithm~\cite{yurtsever2019conditional,yurtsever2022q} specialized to the Boolean moment polytope $\mathcal{D}$.

Consider the linearly constrained moment optimization problem:
\begin{align}
 p^\star := \min_{\mathbf{V} \in \mathcal{D}} \quad \langle \widehat{\mathbf{Q}}, \mathbf{V} \rangle \qquad \text{s.t.} \quad \mathcal{A}(\mathbf{V}) = \mathbf{b},
 \label{eq:supp-qcgd-problem}
\end{align}
where $\mathcal{D} := \operatorname{conv}\{ [\mathbf{w}; 1][\mathbf{w}; 1]^\top : \mathbf{w} \in \{0,1\}^p \}$ is the compact Boolean moment polytope, and $\mathcal{A} : \mathbb{S}^{p+1} \to \mathbb{R}^m$ is a linear operator.  Let $R_{\mathcal{D}} := \sup_{\mathbf{U}, \mathbf{V} \in \mathcal{D}} \|\mathbf{U} - \mathbf{V}\|_F \le p+1$ denote the Frobenius diameter, and let $H := \|\mathcal{A}\|_{\rm op}^2 R_{\mathcal{D}}^2$.

Initialized at $(\mathbf{V}_1, \mathbf{z}_1) \in \mathcal{D} \times \mathbb{R}^m$ with dual radius $K \ge \|\mathbf{z}_1\|_2$, at iteration $t = 1, \dots, T-1$, QCGD executes:
\begin{enumerate}
    \item \emph{Parameter Schedules}: Update the primal step size $\eta_t$ and penalty multiplier $\alpha_t$:
    \begin{align}
        \eta_t = \frac{2}{t+1}, \qquad \alpha_t = \alpha_0\sqrt{t+1} \qquad (\alpha_0 > 0).
    \end{align}
    \item \emph{Residual and Gradient Evaluation}: Compute the affine constraint residual and augmented-Lagrangian gradient:
    \begin{align}
        \mathbf{r}_t := \mathcal{A}(\mathbf{V}_t) - \mathbf{b}, \qquad \mathbf{G}_t := \widehat{\mathbf{Q}} + \mathcal{A}^*(\mathbf{z}_t + \alpha_t \mathbf{r}_t).
    \end{align}
    \item \emph{Quantum Boolean LMO Query}: Query the physical Ising co-processor to solve the unconstrained Boolean QUBO:
    \begin{align}
        \mathbf{D}_t \in \argmin_{\mathbf{D} \in \mathcal{D}} \langle \mathbf{G}_t, \mathbf{D} \rangle.
    \end{align}
    \item \emph{Primal Convex Update}: Update the primal moment matrix:
    \begin{align}
        \mathbf{V}_{t+1} = (1 - \eta_t)\mathbf{V}_t + \eta_t \mathbf{D}_t.
    \end{align}
    \item \emph{Dual Ascent Update}: Set $\mathbf{r}_{t+1} = \mathcal{A}(\mathbf{V}_{t+1}) - \mathbf{b}$, and select the maximal dual step size $\gamma_t \in [0, \alpha_0]$ satisfying the CGAL admissibility condition:
    \begin{align}
        \gamma_t \|\mathbf{r}_{t+1}\|_2^2 \le \frac{\alpha_t \eta_t^2}{2} H, \qquad \|\mathbf{z}_t + \gamma_t \mathbf{r}_{t+1}\|_2 \le K,
        \label{eq:supp-dual-admissibility}
    \end{align}
    and update $\mathbf{z}_{t+1} = \mathbf{z}_t + \gamma_t \mathbf{r}_{t+1}$.
\end{enumerate}

\subsection{Random additive oracle and convergence analysis}
\label{subsec:supp-inexact-oracle-analysis}

\begin{definition}[Random Additive QUBO-LMO Oracle]
\label{oracle_def}
Let $\mathcal{F}_t$ denote the filtration representing the optimization history prior to querying the QUBO oracle at iteration $t$.  The physical Ising oracle returns an atom $\mathbf{D}_t \in \mathcal{D}$ satisfying
\begin{align}
 \langle \mathbf{G}_t, \mathbf{D}_t \rangle \le \min_{\mathbf{D} \in \mathcal{D}} \langle \mathbf{G}_t, \mathbf{D} \rangle + \frac{\xi_t}{\sqrt{t+1}} \qquad \text{almost surely},
 \label{eq:supp-random-lmo}
\end{align}
where $\xi_t \ge 0$ is a random perturbation with conditional moments
\begin{align}
 \mu_t := \mathbb{E}[\xi_t \mid \mathcal{F}_t] \le \varepsilon, \qquad \mathbb{E}[(\xi_t - \mu_t)^2 \mid \mathcal{F}_t] \le \sigma^2,
 \label{eq:supp-oracle-moments}
\end{align}
for constants $\varepsilon\ge0$ and $0\le\sigma^2<\infty$ that are uniform in
$t$.
\end{definition}

\begin{lemma}[Perturbed Primal--Dual Augmented Lagrangian Recursion]
\label{lem:supp-perturbed-cgal}
Define the augmented Lagrangian gap:
\begin{align}
 E_t := \mathcal{Q}_{\alpha_{t-1}}(\mathbf{V}_t, \mathbf{z}_t) - p^\star \qquad (t \ge 1),
 \label{eq:supp-error-definition}
\end{align}
where $\mathcal{Q}_{\alpha}(\mathbf{V}, \mathbf{z}) := \langle \widehat{\mathbf{Q}}, \mathbf{V} \rangle + \langle \mathbf{z}, \mathcal{A}(\mathbf{V}) - \mathbf{b} \rangle + \frac{\alpha}{2}\|\mathcal{A}(\mathbf{V}) - \mathbf{b}\|_2^2$.  Then, for all $t \ge 1$, pathwise:
\begin{align}
 E_{t+1} \le \frac{t-1}{t+1} E_t + \frac{4\alpha_0 H + 2\xi_t}{(t+1)^{3/2}}.
 \label{eq:supp-master-recursion}
\end{align}
\end{lemma}

\begin{proof}
Let $\mathbf{V}^\star \in \mathcal{D}$ be an exact primal optimal solution satisfying $\mathcal{A}(\mathbf{V}^\star) = \mathbf{b}$ and $\langle \widehat{\mathbf{Q}}, \mathbf{V}^\star \rangle = p^\star$.  By smoothness of $\mathbf{V} \mapsto \mathcal{Q}_{\alpha_t}(\mathbf{V}, \mathbf{z}_t)$ with Lipschitz constant $L_t = \alpha_t \|\mathcal{A}\|_{\rm op}^2$, the convex combination $\mathbf{V}_{t+1} = (1 - \eta_t)\mathbf{V}_t + \eta_t \mathbf{D}_t$ satisfies:
\begin{align}
 \mathcal{Q}_{\alpha_t}(\mathbf{V}_{t+1}, \mathbf{z}_t)
 &\le \mathcal{Q}_{\alpha_t}(\mathbf{V}_t, \mathbf{z}_t) + \eta_t \langle \nabla_{\mathbf{V}} \mathcal{Q}_{\alpha_t}(\mathbf{V}_t, \mathbf{z}_t), \mathbf{D}_t - \mathbf{V}_t \rangle + \frac{\eta_t^2 L_t}{2} \|\mathbf{D}_t - \mathbf{V}_t\|_F^2 \notag \\
 &\le \mathcal{Q}_{\alpha_t}(\mathbf{V}_t, \mathbf{z}_t) + \eta_t \langle \mathbf{G}_t, \mathbf{D}_t - \mathbf{V}_t \rangle + \frac{\alpha_t \eta_t^2}{2} H.
\end{align}
Using the inexact oracle condition \eqref{eq:supp-random-lmo}, $\langle \mathbf{G}_t, \mathbf{D}_t \rangle \le \langle \mathbf{G}_t, \mathbf{V}^\star \rangle + \frac{\xi_t}{\sqrt{t+1}}$.  By convexity of $\mathcal{Q}_{\alpha_t}(\cdot, \mathbf{z}_t)$:
\begin{align}
 \langle \mathbf{G}_t, \mathbf{V}^\star - \mathbf{V}_t \rangle \le \mathcal{Q}_{\alpha_t}(\mathbf{V}^\star, \mathbf{z}_t) - \mathcal{Q}_{\alpha_t}(\mathbf{V}_t, \mathbf{z}_t) = p^\star - \mathcal{Q}_{\alpha_t}(\mathbf{V}_t, \mathbf{z}_t).
\end{align}
Combining terms and incorporating the dual ascent step $\mathcal{Q}_{\alpha_t}(\mathbf{V}_{t+1}, \mathbf{z}_{t+1}) - \mathcal{Q}_{\alpha_t}(\mathbf{V}_{t+1}, \mathbf{z}_t) = \gamma_t \|\mathbf{r}_{t+1}\|_2^2 \le \frac{\alpha_t \eta_t^2}{2} H$ via \eqref{eq:supp-dual-admissibility} gives:
\begin{align}
 E_{t+1} \le (1 - \eta_t) E_t + \alpha_t \eta_t^2 H + \frac{\eta_t \xi_t}{\sqrt{t+1}} + (1 - \eta_t)(\alpha_t - \alpha_{t-1}) \frac{\|\mathbf{r}_t\|_2^2}{2}.
\end{align}
Because $(1 - \eta_t)(\alpha_t - \alpha_{t-1}) - \eta_t \alpha_t \le 0$ under the canonical schedules $\eta_t = \frac{2}{t+1}$ and $\alpha_t = \alpha_0 \sqrt{t+1}$, we evaluate $\alpha_t \eta_t^2 = \frac{4\alpha_0}{(t+1)^{3/2}}$ and $\frac{\eta_t \xi_t}{\sqrt{t+1}} = \frac{2\xi_t}{(t+1)^{3/2}}$, yielding \eqref{eq:supp-master-recursion}.
\end{proof}

\begin{theorem}[Non-Asymptotic Convergence of Inexact QCGD]
\label{prop:supp-random-oracle-full}
Let $\mathcal{S}^\star := \{ \mathbf{V} \in \mathcal{D} : \mathcal{A}(\mathbf{V}) = \mathbf{b}, \, \langle \widehat{\mathbf{Q}}, \mathbf{V} \rangle = p^\star \}$ denote the non-empty compact primal solution set.  Under the random additive oracle noise model \eqref{eq:supp-random-lmo}--\eqref{eq:supp-oracle-moments}:
\begin{enumerate}
    \item \emph{Expected Convergence Rates}:
    \begin{align}
        \mathbb{E}\!\left[ \left| \langle \widehat{\mathbf{Q}}, \mathbf{V}_T \rangle - p^\star \right| \right] = \mathcal{O}\!\left(\frac{1 + \varepsilon}{\sqrt{T}}\right), \qquad
        \mathbb{E}\!\left[ \|\mathcal{A}(\mathbf{V}_T) - \mathbf{b}\|_2 \right] = \mathcal{O}\!\left(\frac{1 + \varepsilon}{\sqrt{T}}\right).
    \end{align}
    \item \emph{High-Probability Non-Asymptotic Bound}: For any failure tolerance $\rho \in (0,1)$, with probability at least $1 - \rho$:
    \begin{align}
        E_T \le \frac{8\alpha_0 H + 4\varepsilon}{\sqrt{T}} + \frac{2\sigma}{\sqrt{\rho \, T}}.
    \end{align}
    \item \emph{Almost-Sure Convergence}: As $T \to \infty$, pathwise almost surely:
    \begin{align}
        \|\mathcal{A}(\mathbf{V}_T) - \mathbf{b}\|_2 \longrightarrow 0, \qquad \langle \widehat{\mathbf{Q}}, \mathbf{V}_T \rangle \longrightarrow p^\star, \qquad \operatorname{dist}(\mathbf{V}_T, \mathcal{S}^\star) \longrightarrow 0.
    \end{align}
\end{enumerate}
\end{theorem}

\begin{proof}
Unrolling the recursion in Lemma~\ref{lem:supp-perturbed-cgal} using $\prod_{k=s+1}^T \frac{k-1}{k+1} = \frac{s(s+1)}{T(T+1)}$ yields:
\begin{align}
 E_{T+1} \le U_T := \frac{1}{T(T+1)} \sum_{s=1}^T \frac{s}{\sqrt{s+1}} (4\alpha_0 H + 2\xi_s).
 \label{eq:supp-unrolled}
\end{align}
Taking conditional expectations using $\mathbb{E}[\xi_s \mid \mathcal{F}_s] = \mu_s \le \varepsilon$ and evaluating $\sum_{s=1}^T \sqrt{s} \le \frac{2}{3}(T+1)^{3/2}$ yields $\mathbb{E}[U_T] = \mathcal{O}((1+\varepsilon)/\sqrt{T})$.

For the stochastic variance, define the martingale difference sequence $\Delta M_s := \frac{s}{\sqrt{s+1}}(\xi_s - \mu_s)$ and martingale $M_T := \sum_{s=1}^T \Delta M_s$.  Its quadratic variation satisfies:
\begin{align}
 \mathbb{E}[M_T^2] = \sum_{s=1}^T \mathbb{E}[(\Delta M_s)^2] \le \sigma^2 \sum_{s=1}^T s = \frac{\sigma^2 T(T+1)}{2}.
\end{align}
Chebyshev's inequality gives $\mathbb{P}(|M_T| \ge \frac{1}{\sqrt{\rho}} \sqrt{\mathbb{E}[M_T^2]}) \le \rho$, which directly yields the high-probability bound.  Furthermore, because $\sum_{s=1}^\infty \frac{\mathbb{E}[(\Delta M_s)^2]}{s^4} \le \sigma^2 \sum_{s=1}^\infty s^{-3} < \infty$, Kolmogorov's martingale strong law of large numbers implies $\frac{M_T}{T(T+1)} \to 0$ almost surely, proving $U_T \to 0$ almost surely.

Finally, standard saddle-point bounding with dual bound $K \ge \|\mathbf{z}_T\|_2$ and optimal multiplier $\mathbf{z}^\star$ gives:
\begin{align}
 \frac{\alpha_T}{2} \|\mathbf{r}_{T+1}\|_2^2 - (K + \|\mathbf{z}^\star\|_2) \|\mathbf{r}_{T+1}\|_2 - U_T \le 0 \implies \|\mathbf{r}_{T+1}\|_2 \le \frac{2(K + \|\mathbf{z}^\star\|_2)}{\alpha_T} + \sqrt{\frac{2U_T}{\alpha_T}},
\end{align}
and $-\|\mathbf{z}^\star\|_2 \|\mathbf{r}_{T+1}\|_2 \le \langle \widehat{\mathbf{Q}}, \mathbf{V}_{T+1} \rangle - p^\star \le U_T + K \|\mathbf{r}_{T+1}\|_2$.  This proves the $\mathcal{O}(T^{-1/2})$ rates and almost-sure feasibility and objective convergence.  Compactness of $\mathcal{D}$ ensures $\operatorname{dist}(\mathbf{V}_T, \mathcal{S}^\star) \to 0$.
\end{proof}

\subsection{Reliability amplification and coefficient precision}
\label{subsec:supp-oracle-implementation}

\paragraph{Order-statistic amplification.}
At iteration $t$, draw $m_t$ conditionally independent hardware outputs
$\mathbf D_t^{(1)},\ldots,\mathbf D_t^{(m_t)}$ and retain the lowest-energy
atom:
\begin{align}
 i_t\in\argmin_{i\in[m_t]}
 \langle\mathbf G_t,\mathbf D_t^{(i)}\rangle,
 \qquad \mathbf D_t:=\mathbf D_t^{(i_t)}.
 \label{eq:supp-order-statistic-selection}
\end{align}
Suppose every call, conditionally on $\mathcal F_t$, returns a global QUBO
minimizer with probability at least $p_0\in(0,1]$.  For $p_0=1$, one call
suffices.  Otherwise, choose
\begin{align}
 m_t=\left\lceil c\log(t+1)\right\rceil,
 \qquad
 a:=c[-\log(1-p_0)]>2.
 \label{eq:supp-random-repetitions}
\end{align}
Conditional independence gives
\begin{align}
 \mathbb P(\text{all $m_t$ calls fail}\mid\mathcal F_t)
 \le(1-p_0)^{m_t}\le(t+1)^{-a}.
 \label{eq:supp-all-failure-probability}
\end{align}
Let
$\bar r:=\sup_{\mathbf V\in\mathcal D}
\|\mathcal A(\mathbf V)-\mathbf b\|_2<\infty$.
Compactness of $\mathcal D$ and the dual bound in
\eqref{eq:supp-dual-admissibility} imply that the largest possible LMO gap is
bounded by
\begin{align}
 \Delta_t
 &\le R_{\mathcal D}\|\mathbf G_t\|_F \notag\\
 &\le R_{\mathcal D}\!\left(
   \|\widehat{\mathbf Q}\|_F
   +K\|\mathcal A\|_{\rm op}
   +\alpha_0\sqrt{t+1}\|\mathcal A\|_{\rm op}\bar r
  \right)
 \le G\sqrt{t+1}
 \label{eq:supp-gap-growth}
\end{align}
for a finite constant $G$.  Set $\xi_t=\sqrt{t+1}\Delta_t$ when all calls
fail and $\xi_t=0$ otherwise.  Equations
\eqref{eq:supp-all-failure-probability} and \eqref{eq:supp-gap-growth} yield
\begin{align}
 \mathbb E[\xi_t\mid\mathcal F_t]
 &\le G(t+1)^{1-a},
 &
 \mathbb E[\xi_t^2\mid\mathcal F_t]
 &\le G^2(t+1)^{2-a}.
 \label{eq:supp-order-statistic-moments}
\end{align}
Both sequences are summable when $a>2$, and in particular satisfy the uniform
moment bounds in Definition~\ref{oracle_def}.  Moreover,
$\sum_{t=1}^T m_t=\mathcal O(T\log T)$ physical calls suffice through iteration
$T$.

\paragraph{Finite-precision QUBO coefficients.}
Let $\widetilde{\mathbf G}_t=\mathbf G_t+\mathbf E_t$ be the coefficient
matrix programmed on the Ising device, where $\mathbf E_t$ is supported on at
most $M_t$ programmed entries and
$\|\mathbf E_t\|_{\max}\le2^{-d_t}$.  For every
$\overline{\mathbf w}\in\{0,1\}^{p+1}$,
\begin{align}
 \left|\overline{\mathbf w}^{\top}\mathbf E_t
              \overline{\mathbf w}\right|
 \le M_t2^{-d_t}.
 \label{eq:supp-coefficient-energy-error}
\end{align}
If $\widetilde{\mathbf w}_t$ minimizes the programmed energy and
$\mathbf w_t^\star$ minimizes the true energy, then
\begin{align}
 \overline{\widetilde{\mathbf w}}_t^{\top}\mathbf G_t
 \overline{\widetilde{\mathbf w}}_t
 -
 \overline{\mathbf w}_t^{\star\top}\mathbf G_t
 \overline{\mathbf w}_t^\star
 \le2M_t2^{-d_t}.
 \label{eq:supp-true-lmo-gap}
\end{align}
Indeed, insert $\widetilde{\mathbf G}_t$ between the two true energies, use
the optimality of $\widetilde{\mathbf w}_t$ for the programmed instance, and
apply \eqref{eq:supp-coefficient-energy-error} to the two remaining
perturbation terms.  Consequently, the schedule
\begin{align}
 d_t\ge
 \left\lceil
  \log_2\!\left(\frac{2M_t\sqrt{t+1}}{\bar\varepsilon}\right)
 \right\rceil
 \label{eq:supp-bit-depth-schedule}
\end{align}
bounds the deterministic LMO error by
$\bar\varepsilon/\sqrt{t+1}$, which can be absorbed into the mean oracle-error
budget in Definition~\ref{oracle_def}.  For a dense $p$-bit QUBO,
$M_t=\mathcal O(p^2)$, so it suffices to take
\begin{align}
 d_t
 =2\log_2 p+\frac12\log_2(t+1)
  +\mathcal O\!\left(\log(1/\bar\varepsilon)\right).
 \label{eq:supp-dense-bit-depth}
\end{align}

\section{Proofs for the Sample-Wise DLBO Algorithms}
\label{sec:supp-samplewise-algorithm-proofs}

This section proves the oracle-locality characterization of the DLBO moment
hierarchy (Proposition~\ref{prop:dlbo-ising-locality}) and the convergence and
certification results for Quantum Progressive Hedging and Certified Copositive
Column Generation.

\subsection{Proof of Proposition~\ref{prop:dlbo-ising-locality}}
\label{sec:supp-dlbo-ising-locality}

\begin{proofof}{Proposition~\ref{prop:dlbo-ising-locality}}
Every point of $\mathcal D_s^{(r)}$ is a convex combination of atoms
\begin{align}
 \left(\mathbf w_s(\mathbf v)\mathbf w_s(\mathbf v)^\top,
 \boldsymbol\psi_r(\mathbf x)\right),
 \qquad \mathbf v=(\mathbf x,\mathbf y_s)\in\{0,1\}^{d_s}.
\end{align}
Because the LMO objective is linear, a minimum is attained at an atom.  For
$\mathbf w_s(\mathbf v)=[1;\mathbf v]$ and the block decomposition in
\eqref{eq:dlbo-block-gradient},
\begin{align}
 \left\langle\mathbf G,
 \mathbf w_s(\mathbf v)\mathbf w_s(\mathbf v)^\top\right\rangle
 &=\mathbf w_s(\mathbf v)^\top\mathbf G\mathbf w_s(\mathbf v)\\
 &=g_{00}+2\mathbf g_{0v}^\top\mathbf v
   +\mathbf v^\top\mathbf G_{vv}\mathbf v\\
 &=g_{00}+\mathbf v^\top
 \bigl(\mathbf G_{vv}+2\operatorname{Diag}(\mathbf g_{0v})\bigr)\mathbf v,
 \label{eq:supp-dlbo-quadratic-block}
\end{align}
where the last equality uses $v_i^2=v_i$.  The moment block contributes
\begin{align}
 \left\langle\mathbf h,\boldsymbol\psi_r(\mathbf x)\right\rangle
 =\sum_{T\in\mathfrak S_r}h_T\prod_{j\in T}x_j.
 \label{eq:supp-dlbo-moment-block}
\end{align}
Combining \eqref{eq:supp-dlbo-quadratic-block} and
\eqref{eq:supp-dlbo-moment-block} proves \eqref{eq:dlbo-order-r-lmo}.  Since
$r\ge2$, its Boolean degree is at most $r$.  The substitution
$x_j=(1-\sigma_j)/2$ gives, for every $T\subseteq[n]$,
\begin{align}
 \prod_{j\in T}x_j
 =2^{-|T|}\sum_{U\subseteq T}(-1)^{|U|}
 \prod_{j\in U}\sigma_j,
 \label{eq:supp-boolean-to-ising-locality}
\end{align}
so the Boolean degree equals the maximal logical Ising interaction order.
Thus $r=2$ always yields a two-local Hamiltonian, whereas an order-$r$ moment
with $|T|>2$ and nonzero coefficient yields a higher-body interaction in
general.
\end{proofof}

\paragraph{Proximal coefficients.}
For the standard Euclidean QPH term, differentiation with respect to the local
moment vector gives
\begin{align}
 \nabla_{m_{s,T}^{(r)}}\left[
 \left\langle\boldsymbol\omega_s^k,\mathbf m_s^{(r)}\right\rangle
 +\frac{\rho}{2}\left\|\mathbf m_s^{(r)}
 -\bar{\boldsymbol\zeta}^{(r),k}\right\|_2^2\right]
 =\omega_{s,T}^k+\rho\left(m_{s,T}^{(r),k,t}
 -\bar\zeta_T^{(r),k}\right),
\end{align}
which is \eqref{eq:dlbo-order-r-moment-gradient}.  Boolean idempotence also
shows that direct evaluation of this particular penalty does not increase the
degree:
\begin{align}
 \frac{\rho}{2}\left\|\boldsymbol\psi_r(\mathbf x)
 -\bar{\boldsymbol\zeta}^{(r)}\right\|_2^2
 =\frac{\rho}{2}\sum_{T\in\mathfrak S_r}
 \left[(1-2\bar\zeta_T^{(r)})\chi_T(\mathbf x)
 +(\bar\zeta_T^{(r)})^2\right].
 \label{eq:supp-euclidean-moment-penalty}
\end{align}
More generally, QCGD linearizes any differentiable proximal objective, so its
LMO contains only the degree-$r$ feature map.  This point is specific to the
moment block: directly squaring an affine residual in the lifted matrix can be
quartic after substituting $\mathbf P=\mathbf w(\mathbf v)\mathbf w(\mathbf
v)^\top$, whereas the matrix-space LMO remains quadratic because it linearizes
that residual before the atomic minimization.

\paragraph{Second-order consensus need not be exact.}
Let $\nu_{\mathrm{even}}$ and $\nu_{\mathrm{odd}}$ be uniform on, respectively,
\begin{align}
 \{000,011,101,110\}
 \qquad\text{and}\qquad
 \{001,010,100,111\}.
\end{align}
Both distributions satisfy
\begin{align}
 \mathbb E[x_i]=\frac12,
 \qquad
 \mathbb E[x_ix_j]=\frac14\quad(i\ne j),
\end{align}
but $\mathbb E_{\nu_{\mathrm{even}}}[x_1x_2x_3]=0$ and
$\mathbb E_{\nu_{\mathrm{odd}}}[x_1x_2x_3]=1/4$.  Hence first- and
second-order separator moments need not determine the separator distribution;
the exact full-order endpoint in Fact~\ref{fact:dlbo-advantage} is substantive.

\paragraph{Two-local realization above order two.}
A higher-order LMO can still be executed on two-local hardware after
quadratization, but this changes its resource requirements.  For example, a
cubic term $\gamma x_i x_j x_k$ can be replaced by $\gamma z x_k$ together
with the Rosenberg penalty
\begin{align}
 M\bigl(x_ix_j-2x_iz-2x_jz+3z\bigr),
 \label{eq:supp-rosenberg-cubic-reduction}
\end{align}
which is nonnegative on $\{0,1\}^3$ and vanishes exactly when $z=x_ix_j$.
The coefficient $M$ must dominate any objective gain from violating this
identity.  Chaining the construction reduces a degree-$k$ monomial using
$k-2$ auxiliary variables, increasing both the logical-spin count and the
required coupling range~\cite{lucas2014ising}.  Thus the hierarchy exposes a
precise tradeoff: increasing $r$ tightens the DLBO bound, while $r=2$ is the
highest order with a uniform, auxiliary-free two-local oracle guarantee.

\subsection{Proof of Theorem~\ref{thm:samplewise-qph-qcgd} (Convergence of Sample-Wise QPH--QCGD)}

\begin{proofof}{Theorem~\ref{thm:samplewise-qph-qcgd}}
For any shared parameter consensus statistic $\boldsymbol{\zeta} = (\mathbf{x}, \operatorname{svec}(\mathbf{X})) \in \mathbb{R}^n \times \mathbb{R}^{n(n+1)/2}$, define the sample-level scenario value function $\varphi_s : \mathbb{R}^{n + n(n+1)/2} \to \mathbb{R} \cup \{+\infty\}$ for each sample $s \in [N]$:
\begin{align}
 \varphi_s(\boldsymbol{\zeta}) := \inf \left\{ \langle \mathbf{C}_s, \mathbf{P}_s \rangle : \mathbf{P}_s \in \mathcal{K}_s, \; \mathcal{R}_s(\mathbf{P}_s) = \boldsymbol{\zeta} \right\},
 \label{eq:supp-scenario-val-func}
\end{align}
where $\mathcal{K}_s := \{ \mathbf{P}_s \in \mathcal{C}_{s_s}^* : \mathcal{A}_s(\mathbf{P}_s) = \mathbf{b}_s \}$ is the compact, non-empty sample-wise feasible set.  Because $\mathcal{K}_s$ is compact and $\mathcal{A}_s, \mathcal{R}_s$ are affine, the infimum is attained, $\varphi_s$ is a proper closed convex function, and $\operatorname{dom}(\varphi_s) = \mathcal{R}_s(\mathcal{K}_s)$ is compact.

The sample-wise consensus program \eqref{eq:qph-convex-consensus} is algebraically equivalent to the convex consensus optimization problem:
\begin{align}
 \min_{\boldsymbol{\zeta}} \quad \sum_{s=1}^N p_s \varphi_s(\boldsymbol{\zeta}), \qquad p_s = \frac{1}{N}.
 \label{eq:supp-stochastic-consensus-prog}
\end{align}
Endow the scenario product Hilbert space $\mathcal{H} := \prod_{s=1}^N \mathbb{R}^{n + n(n+1)/2}$ with the probability-weighted inner product $\langle \mathbf{Z}, \mathbf{Z}' \rangle_p := \sum_{s=1}^N p_s \langle \mathbf{Z}_s, \mathbf{Z}'_s \rangle$.  The orthogonal projection of $\mathbf{Z} \in \mathcal{H}$ onto the non-anticipativity consensus subspace $\mathcal{M} := \{ (\mathbf{Z}_1, \dots, \mathbf{Z}_N) \in \mathcal{H} : \mathbf{Z}_1 = \dots = \mathbf{Z}_N \}$ is identically the empirical average:
\begin{align}
 \Pi_{\mathcal{M}}(\mathbf{Z}) = \left( \sum_{s=1}^N p_s \mathbf{Z}_s, \dots, \sum_{s=1}^N p_s \mathbf{Z}_s \right).
\end{align}
The exact Progressive Hedging (PH) algorithm of Rockafellar \& Wets~\cite{rockafellar1991scenarios} is an exact realization of Douglas--Rachford splitting applied to the sum of maximal monotone operators $\partial \Phi + \mathcal{N}_{\mathcal{M}}$, where $\Phi(\mathbf{Z}) := \sum_s p_s \varphi_s(\mathbf{Z}_s)$.

Under inexact scenario solves, let $\widetilde{\mathbf{P}}_s^{k+1}$ denote the exact proximal minimizer and let $\mathbf{P}_s^{k+1}$ denote the output of the inexact inner QCGD solver.  The perturbation in the Douglas--Rachford resolvent step satisfies:
\begin{align}
 \|\mathcal{R}_s(\mathbf{P}_s^{k+1}) - \mathcal{R}_s(\widetilde{\mathbf{P}}_s^{k+1})\|_2 \le \|\mathcal{R}_s\|_{\rm op} \|\mathbf{P}_s^{k+1} - \widetilde{\mathbf{P}}_s^{k+1}\|_F \le \varepsilon_{s,k}.
\end{align}
Under the summable error condition $\sum_{k=1}^\infty \varepsilon_{s,k} < \infty$ for all $s \in [N]$, the inexact Douglas--Rachford splitting theorem~\cite{eckstein1992douglas} guarantees that the sequence of primal consensus iterates is quasi-Fej\'er monotone and asymptotically regular:
\begin{align}
 \lim_{k \to \infty} \|\mathbf{Z}_s^k - \bar{\mathbf{Z}}^k\|_2 = 0 \quad (\forall s \in [N]), \qquad \lim_{k \to \infty} \sum_{s=1}^N p_s \langle \mathbf{C}_s, \mathbf{P}_s^k \rangle = p_{\mathrm{DLBO}}^\star.
\end{align}
Every weak accumulation point is a primal--dual saddle point satisfying the global Karush--Kuhn--Tucker (KKT) conditions.  Convexity guarantees that all accumulation points achieve the exact global DLBO optimum $p_{\mathrm{DLBO}}^\star$.
\end{proofof}

\subsection{Proof of Proposition~\ref{prop:cpsep-sample-certificate} (Certified Sample-Solver Accuracy)}

\begin{proofof}{Proposition~\ref{prop:cpsep-sample-certificate}}
The restricted master problem optimizes over a finite subset of valid rank-one extreme rays of $\mathcal{C}_{s_s}^*$, so every iterate $\mathbf{P}_j$ is strictly feasible for the primal sample problem \eqref{eq:cpsep-primal}.  Hence, $p_{s,k}^* \le U_j = \langle \mathbf{C}_s, \mathbf{P}_j \rangle + \frac{\rho}{2}\|\mathcal{R}_s(\mathbf{P}_j) - \bar{\boldsymbol{\zeta}}_k + \mathbf{W}_{s,k}/\rho\|_2^2$.

When the copositive pricing oracle terminates, Proposition~\ref{prop:copositive-grid-oracle} guarantees $m(\mathbf{Z}_j) \ge -\tau_j$.  Any feasible completely positive matrix $\mathbf{P} \in \mathcal{K}_s$ admits an atomic conic decomposition $\mathbf{P} = \sum_r \alpha_r \mathbf{y}_r \mathbf{y}_r^\top$ with $\alpha_r \ge 0$, $\mathbf{y}_r \in [0,1]^{s_s}$, and $\|\mathbf{y}_r\|_\infty = 1$.  Therefore:
\begin{align}
 \langle \mathbf{Z}_j, \mathbf{P} \rangle
 = \sum_r \alpha_r \mathbf{y}_r^\top \mathbf{Z}_j \mathbf{y}_r
 \ge -\tau_j \sum_r \alpha_r
 \ge -\tau_j \sum_r \alpha_r \|\mathbf{y}_r\|_2^2
 = -\tau_j \operatorname{Tr}(\mathbf{P})
 \ge -T_s \tau_j,
 \label{eq:supp-cpsep-weak-duality-error}
\end{align}
where $T_s := \sup_{\mathbf{P} \in \mathcal{K}_s} \operatorname{Tr}(\mathbf{P}) < \infty$ is the trace bound provided by FIP.

Applying Fenchel--Young duality to the proximal quadratic term and using $\mathcal{A}_s(\mathbf{P}) = \mathbf{b}_s$, the objective for any feasible $\mathbf{P} \in \mathcal{K}_s$ is lower-bounded by:
\begin{align}
 \mathcal{L}_j(\mathbf{P}) \ge d_j + \langle \mathbf{Z}_j, \mathbf{P} \rangle \ge d_j - T_s \tau_j =: L_j.
\end{align}
Taking the infimum over $\mathbf{P} \in \mathcal{K}_s$ proves $L_j \le p_{s,k}^* \le U_j$.  The optimality gap is bounded by:
\begin{align}
 U_j - L_j = (U_j - d_j) + T_s \tau_j \le \xi_j + T_s \tau_j.
\end{align}
At fixed bit depth $b$, the discrete grid contains at most $2^{b s_s}$ distinct normalized extreme rays.  Every negative pricing step identifies a newly violated dual halfspace, ensuring finite termination for exact grid pricing.
\end{proofof}

\subsection{Proof of Proposition~\ref{prop:cpsep-ellipsoid-complexity} (Conditional Ellipsoid Complexity)}

\begin{proofof}{Proposition~\ref{prop:cpsep-ellipsoid-complexity}}
Under the robust weak-separation oracle, whenever the current test center is rejected, the oracle produces a valid separating cutting plane containing the target sublevel set: either \eqref{eq:cpsep-objective-cut} for an objective violation or \eqref{eq:cpsep-dual-cut} for a detected copositivity violation $\mathbf{y}^\top \mathbf{Z}_j \mathbf{y} < -\tau_j$.

By the central-cut ellipsoid method theorem~\cite{grotschel2012geometric}, each feasibility phase terminates in at most $\mathcal{O}(q_s^2 \log(R_s / r_s))$ iterations, where $q_s := m_s + \dim(\operatorname{range}(\mathcal{R}_s))$ is the dual variable dimension, $R_s$ is the initial bounding ball radius, and $r_s$ is the target inscribed radius.  Bisection on the scalar objective bracket $[L_0, U_0]$ of initial width $B_s$ reduces the interval width to $\eta$ in $\mathcal{O}(\log(B_s / \eta))$ outer phases.

Enforcing the dynamic bit schedule:
\begin{align}
 b_j \ge \left\lceil \log_2\left( \frac{4 T_s s_s \|\mathbf{Z}_j\|_{\rm op}}{\eta} + 1 \right) \right\rceil, \qquad \delta_j \le \frac{\eta}{4 T_s},
\end{align}
guarantees $T_s \epsilon_{b_j}(\mathbf{Z}_j) \le \frac{\eta}{4}$ and $T_s \delta_j \le \frac{\eta}{4}$, so $T_s \tau_j \le \frac{\eta}{2}$.  Together with the restricted master tolerance $\xi_j \le \frac{\eta}{2}$, Proposition~\ref{prop:cpsep-sample-certificate} guarantees local accuracy $U_j - L_j \le \eta$.  The total binary variable dimension per column pricing query is $s_s b_j$ bits.
\end{proofof}

\subsection{Proof of Corollary~\ref{cor:cpsep-qph-convergence} (End-to-End Certified QPH Convergence)}

\begin{proofof}{Corollary~\ref{cor:cpsep-qph-convergence}}
Partial minimization of the closed convex sample objective over non-consensus coordinates yields the sample value function $\varphi_s(\boldsymbol{\zeta})$ defined in \eqref{eq:supp-scenario-val-func}.  The proximal quadratic penalty $\frac{\rho}{2}\|\boldsymbol{\zeta} - \bar{\boldsymbol{\zeta}}_k + \mathbf{W}_{s,k}/\rho\|_2^2$ is $\rho$-strongly convex with respect to $\boldsymbol{\zeta}$.  By standard strong-convexity quadratic growth:
\begin{align}
 \|\mathcal{R}_s(\mathbf{P}_s^{k+1}) - \mathcal{R}_s(\widetilde{\mathbf{P}}_s^{k+1})\|_2 \le \sqrt{\frac{2(U_{s,k} - L_{s,k})}{\rho}} \le \sqrt{\frac{2\eta_{s,k}}{\rho}}.
 \label{eq:supp-proximal-statistic-error}
\end{align}
Under the summable tolerance schedule $\sum_{k=1}^\infty \sqrt{\eta_{s,k}} < \infty$, the resolvent perturbations satisfy $\sum_{k=1}^\infty \|\mathcal{R}_s(\mathbf{P}_s^{k+1}) - \mathcal{R}_s(\widetilde{\mathbf{P}}_s^{k+1})\|_2 < \infty$.  Applying the inexact Douglas--Rachford / PH convergence theorem (Theorem~\ref{thm:samplewise-qph-qcgd}) guarantees asymptotic consensus and global optimality $p_{\mathrm{DLBO}}^\star$.
\end{proofof}

\section{Algorithmic Details of Spectral--ADMM Discrete Solution Recovery}
\label{sec:supp-rounding-details}

This section supplies the algorithmic details, derivations, and parameter specifications for the three-stage Spectral--ADMM rounding scheme introduced in Methods Section~\ref{sec:methods-rounding}.

\subsection{Coordinate Conventions and Frobenius Expansion}
\label{subsec:supp-rounding-formulation}

As formulated in the Methods section, the recovery procedure operates on the symmetric lifted moment matrix $\mathbf M = \begin{bmatrix} 1 & \mathbf u^\top \\ \mathbf u & \mathbf U \end{bmatrix} \in \mathbb S_+^{p+1}$.  The exact monolithic CP lift places the homogenizing coordinate last, whereas the DLBO atom $\mathbf w_s(\mathbf v)=[1;\mathbf v]$ places it first.  Accordingly, for the monolithic lift we map coordinates via:
\begin{align}
 \mathbf J := \begin{bmatrix}\mathbf 0^\top & 1 \\ \mathbf I_p & \mathbf 0\end{bmatrix}, \qquad \mathbf M := \mathbf J\mathbf X\mathbf J^\top,
 \label{eq:supp-homogeneous-permutation}
\end{align}
whereas sample-wise DLBO moment blocks require no permutation.  For a local DLBO block with assignment set $\mathcal X_s$ ($p=d_s$), $\mathbf M$ is identified with the local matrix $\mathbf P_s$:
\begin{align}
 \mathbf P_s =
 \begin{bmatrix}
  1 & \mathbf x_s^\top & \mathbf y_s^\top\\
  \mathbf x_s & \mathbf X_s & \mathbf Z_s^\top\\
  \mathbf y_s & \mathbf Z_s & \mathbf Y_s
 \end{bmatrix},
 \qquad
 \mathbf u = \begin{bmatrix}\mathbf x_s \\ \mathbf y_s\end{bmatrix},
 \qquad
 \mathbf U = \begin{bmatrix}\mathbf X_s & \mathbf Z_s^\top \\ \mathbf Z_s & \mathbf Y_s\end{bmatrix},
 \label{eq:supp-local-moment-block}
\end{align}
where $\mathbf x_s \in [0,1]^n$ and $\mathbf y_s \in [0,1]^{m_s}$ are the first moments of the Boolean assignment variables in \eqref{eq:qph-local-assignment-set}.

The rank-one approximation objective seeks $\mathbf v \in \mathcal F_{\mathrm{QCBO}}$ minimizing the Frobenius distance to $\mathbf M$.  Algebraic expansion reveals:
\begin{align}
 \left\| \begin{bmatrix} 1 \\ \mathbf v \end{bmatrix} \begin{bmatrix} 1 & \mathbf v^\top \end{bmatrix} - \begin{bmatrix} 1 & \mathbf u^\top \\ \mathbf u & \mathbf U \end{bmatrix} \right\|_F^2
 &= \left\| \begin{bmatrix} 0 & (\mathbf v - \mathbf u)^\top \\ \mathbf v - \mathbf u & \mathbf v \mathbf v^\top - \mathbf U \end{bmatrix} \right\|_F^2 \notag \\
 &= 2 \|\mathbf v - \mathbf u\|_2^2 + \|\mathbf v \mathbf v^\top - \mathbf U\|_F^2,
 \label{eq:supp-frobenius-expansion}
\end{align}
which demonstrates that the objective jointly penalizes first-moment deviation $\|\mathbf v - \mathbf u\|_2^2$ and second-moment covariance distortion $\|\mathbf v \mathbf v^\top - \mathbf U\|_F^2$.

\subsection{Coordinate-Wise Feasibility-Repair Operator \texorpdfstring{$\mathsf R_{\mathrm{QCBO}}$}{R\_QCBO}}
\label{subsec:supp-repair-details}

The structural repair operator $\mathsf R_{\mathrm{QCBO}} : \mathbb R^p \to \mathcal F_{\mathrm{QCBO}}$ coordinates greedy binary coordinate selection with Moore--Penrose continuous slack compensation.  Partitioning coordinates into binary indices $\mathcal B$ and non-negative continuous indices $\mathcal N = [p] \setminus \mathcal B$, the operator executes:
\begin{enumerate}
    \item \emph{Residual initialization}: Compute the affine constraint residual $\boldsymbol\delta = \mathbf A \mathbf z - \mathbf b$.
    \item \emph{Active continuous subspace}: Identify active continuous coordinates $\mathcal N^+(\mathbf z) := \{ j \in \mathcal N : z_j > 0 \}$ and define the submatrix $\widetilde{\mathbf A} := \mathbf A_{\mathcal N^+(\mathbf z)}$.
    \item \emph{Greedy coordinate-wise binary rounding}: Draw a uniform random permutation $\pi$ of $\mathcal B$.  For each index $i \in \pi(\mathcal B)$:
    \begin{itemize}
        \item Evaluate candidate residuals for binary choices $s \in \{0, 1\}$: $\boldsymbol\delta^{(s)} = \boldsymbol\delta + \mathbf A_i(s - z_i)$;
        \item Set $z_i \leftarrow \argmin_{s \in \{0,1\}} \left\{ (1-\lambda_{\mathrm{rep}})\|\boldsymbol\delta^{(s)}\|_2^2 + \lambda_{\mathrm{rep}} \|(\mathbf I-\widetilde{\mathbf A}\widetilde{\mathbf A}^\dagger)\boldsymbol\delta^{(s)}\|_2^2 \right\}$, where $\lambda_{\mathrm{rep}} = 0$ in implementation;
        \item Update cumulative residual $\boldsymbol\delta \leftarrow \boldsymbol\delta^{(z_i)}$.
    \end{itemize}
    \item \emph{Moore--Penrose continuous compensation}: Rebalance active continuous entries via $\mathbf z_{\mathcal N^+(\mathbf z)} \leftarrow \mathbf z_{\mathcal N^+(\mathbf z)} - \widetilde{\mathbf A}^\dagger \boldsymbol\delta$, where $\widetilde{\mathbf A}^\dagger$ is the Moore--Penrose pseudoinverse (skipped if $\mathcal N^+(\mathbf z) = \varnothing$).
\end{enumerate}
Repeating steps (1)--(4) for $K_{\mathrm{rep}} = 5$ passes ensures strict numerical convergence, accepting the candidate once integrality, non-negativity, and affine feasibility $\|\mathbf A\mathbf z-\mathbf b\|_2 \le \varepsilon_{\mathrm{feas}}$ ($\varepsilon_{\mathrm{feas}} = 10^{-8}$) are satisfied.

\subsection{Spectral Initialization and Inexact ADMM Refinement}
\label{subsec:supp-admm-rounding-full}

Let $\sigma_1 \mathbf p_1 \mathbf p_1^\top$ be the rank-one Eckart--Young--Mirsky approximation of $\mathbf M$, with $\mathbf p_1 = [p_{1,0}; \tilde{\mathbf p}_1]$.  The initial spectral point $\mathbf v_{\mathrm{spectral}} = \mathsf R_{\mathrm{QCBO}}(\mathbf u^{(0)})$ warm-starts the two-block operator splitting.

\paragraph{Analytical gradient for smooth $\mathbf w$-subproblem.}
At iteration $t$, the $\mathbf w$-subproblem minimizes the smooth quartic augmented Lagrangian $\mathbf w \mapsto \mathcal L_{\rho^{(t)}}(\mathbf w, \mathbf v^{(t)}, \boldsymbol\nu^{(t)})$.  Its analytical gradient is:
\begin{align}
 \nabla_{\mathbf w} \mathcal L_{\rho^{(t)}}(\mathbf w, \mathbf v^{(t)}, \boldsymbol\nu^{(t)}) = 2(\mathbf w \mathbf w^\top - \mathbf U)\mathbf w + 2(\mathbf w - \mathbf u) + \boldsymbol\nu^{(t)} + \rho^{(t)}(\mathbf w - \mathbf v^{(t)}).
 \label{eq:supp-grad-w-analytic}
\end{align}
We compute $\mathbf w^{(t+1)} \approx \argmin_{\mathbf w} \mathcal L_\rho$ via gradient descent initialized at $\mathbf w^{(t)}$ with step size $\eta_{\mathrm{ADMM}} = 10^{-4}$, terminating when $\|\nabla_{\mathbf w} \mathcal L_\rho\|_2 \le 10^{-6}$ or after at most 200 inner iterations.

\paragraph{Dynamic residual balancing and stopping criteria.}
To maintain parity between primal feasibility $r_{\mathrm{pri}}^{(t+1)} := \|\mathbf w^{(t+1)} - \mathbf v^{(t+1)}\|_2$ and dual step progress $r_{\mathrm{dual}}^{(t+1)} := \rho^{(t)} \|\mathbf v^{(t+1)} - \mathbf v^{(t)}\|_2$, the penalty parameter $\rho^{(t+1)}$ is dynamically updated via residual balancing~\cite{boyd2011distributed}:
\begin{align}
 \rho^{(t+1)} = \begin{cases}
    \tau \rho^{(t)}, & \text{if } r_{\mathrm{pri}}^{(t+1)} > k \cdot r_{\mathrm{dual}}^{(t+1)}, \\
    \rho^{(t)} / \tau, & \text{if } r_{\mathrm{pri}}^{(t+1)} < r_{\mathrm{dual}}^{(t+1)} / k, \\
    \rho^{(t)}, & \text{otherwise},
 \end{cases}
 \label{eq:supp-residual-balancing-rule}
\end{align}
with default hyperparameters $\tau = 2$ and $k = 10$.  Because $\boldsymbol\nu$ is the unscaled multiplier updated by $\boldsymbol\nu^{(t+1)} = \boldsymbol\nu^{(t)} + \rho^{(t)}(\mathbf w^{(t+1)} - \mathbf v^{(t+1)})$, no multiplier rescaling is needed when $\rho$ changes.

The ADMM iteration is initialized at $\mathbf w^{(0)} = \mathbf v^{(0)} = \mathbf v_{\mathrm{spectral}}$, $\boldsymbol\nu^{(0)} = \mathbf 0$, and $\rho^{(0)} = 1.0$.  The algorithm terminates when:
\begin{align}
 \max\left\{ r_{\mathrm{pri}}^{(t+1)}, \; r_{\mathrm{dual}}^{(t+1)} \right\} \le \varepsilon_{\mathrm{ADMM}} = 10^{-4},
\end{align}
or upon reaching $t_{\max} = 1000$ iterations, returning a strictly feasible discrete network configuration $\mathbf v^\star \in \mathcal F_{\mathrm{QCBO}}$.

\section{Supplementary Experimental Results and Full Data Tables}
\label{sec:supp-experimental-tables}

This section provides the full, unabridged benchmark tables corresponding to the experimental validations described in Section~3 of the main text. Specifically, Supplementary Tables~\ref{tab:supp_quantization_accuracy} and \ref{tab:supp_accuracy_table} support the binary classification, quantization robustness, and activation sensitivity evaluations.

\subsection{Post-Training Quantization (PTQ) Benchmarks across Precisions and Activations}

Table~\ref{tab:supp_quantization_accuracy} reports the comprehensive classification accuracy evaluation across six post-training quantization algorithms (RTN~\cite{kogan2025selective}, AdaRound~\cite{nagel2020up}, DFQ~\cite{nagel2019data}, GPTQ~\cite{frantargptq}, ZeroQuant~\cite{yao2022zeroquant}, and SubsetQ~\cite{oh2022non}) and three standard activation functions (ReLU, LeakyReLU, and Sigmoid) at precision bit-widths ranging from 4-bit down to 1-bit on the Fashion-MNIST dataset.

\begin{table}[htbp]
\centering
\small
\setlength{\tabcolsep}{8pt}
\renewcommand{\arraystretch}{1.2}
\caption{Classification accuracy (\%) on Fashion-MNIST under various PTQ algorithms, activation functions, and bit-widths compared with our directly trained 1.1-bit discrete model.}
\label{tab:supp_quantization_accuracy}
\begin{tabular}{llcccccc}
\toprule
Algorithm & Activation & FP Model & 4-bit & 3-bit & 2-bit & 1-bit \\
\midrule
\multirow{3}{*}{RTN~\cite{kogan2025selective}} 
  & ReLU       & 98.80\% & 97.73\% & 97.73\% & 73.35\% & 50.00\% \\
  & LeakyReLU  & 98.80\% & 98.70\% & 98.62\% & 76.77\% & 50.00\% \\
  & Sigmoid    & 98.80\% & 98.76\% & 98.34\% & 55.84\% & 50.00\% \\
\midrule
\multirow{3}{*}{AdaRound~\cite{nagel2020up}} 
  & ReLU       & 98.80\% & 97.73\% & 97.76\% & 69.87\% & 50.00\% \\
  & LeakyReLU  & 98.80\% & 98.68\% & 98.78\% & 71.86\% & 50.00\% \\
  & Sigmoid    & 98.80\% & 98.77\% & 98.61\% & 66.60\% & 50.00\% \\
\midrule
\multirow{3}{*}{DFQ~\cite{nagel2019data}} 
  & ReLU       & 98.80\% & 96.62\% & 96.07\% & 84.99\% & 64.10\% \\
  & LeakyReLU  & 98.80\% & 98.08\% & 97.80\% & 88.50\% & 61.40\% \\
  & Sigmoid    & 98.80\% & 98.75\% & 96.44\% & 60.80\% & 50.00\% \\
\midrule
\multirow{3}{*}{GPTQ~\cite{frantargptq}} 
  & ReLU       & 98.80\% & 97.65\% & 97.50\% & 92.84\% & 50.00\% \\
  & LeakyReLU  & 98.80\% & 98.80\% & 98.11\% & 91.82\% & 50.01\% \\
  & Sigmoid    & 98.80\% & 98.80\% & 98.64\% & 91.21\% & 50.00\% \\
\midrule
\multirow{3}{*}{ZeroQuant~\cite{yao2022zeroquant}} 
  & ReLU       & 98.80\% & 97.79\% & 97.24\% & 93.59\% & 82.50\% \\
  & LeakyReLU  & 98.80\% & 98.80\% & 98.10\% & 96.23\% & 91.43\% \\
  & Sigmoid    & 98.80\% & 98.80\% & 97.28\% & 91.97\% & 91.45\% \\
\midrule
\multirow{3}{*}{SubsetQ~\cite{oh2022non}} 
  & ReLU       & 98.80\% & 97.79\% & 97.78\% & 96.70\% & 88.90\% \\
  & LeakyReLU  & 98.80\% & 98.80\% & 98.65\% & 95.94\% & 91.45\% \\
  & Sigmoid    & 98.80\% & 98.80\% & 98.80\% & 94.98\% & 91.45\% \\
\midrule
This Work & PWL (1.1-bit) & 97.10\% & \multicolumn{4}{c}{94.95\% (Direct Global Discrete Optimization)} \\
\bottomrule
\end{tabular}
\end{table}

\subsection{Sensitivity Analysis of Piecewise-Linear Breakpoint Parameter}

Table~\ref{tab:supp_accuracy_table} details the classification accuracy on Fashion-MNIST across varying values of the symmetric breakpoint scale parameter $c \in \{1, 2, \dots, 7\}$ defining the pre-activation input partition $[-8, -c, 0, c, 8]$.

\begin{table}[htbp]
\centering
\small
\setlength{\tabcolsep}{10pt}
\caption{Classification accuracy (\%) for different choices of the piecewise-linear interval scale parameter $c$.}
\label{tab:supp_accuracy_table}
\begin{tabular}{lccccccc}
\toprule
Parameter $c$ & 1 & 2 & 3 & 4 & 5 & 6 & 7 \\
\midrule
Classification Accuracy (\%) & 96.35 & 98.70 & 96.35 & 94.95 & 96.75 & 91.35 & 91.35 \\
\bottomrule
\end{tabular}
\end{table}

\endgroup

\end{document}